%% file: main.tex
\pdfoutput=1
\documentclass{article}
\usepackage{colm2024_arxiv}

\usepackage{microtype}
\usepackage{hyperref}
\usepackage{url}
\usepackage{booktabs}
\definecolor{darkblue}{rgb}{0, 0, 0.5}
\hypersetup{colorlinks=true, citecolor=darkblue, linkcolor=darkblue, urlcolor=darkblue}

\usepackage{multirow}
\usepackage{CJKutf8}
\usepackage{tabularx}
\usepackage{listings}
\usepackage{xcolor} 
\usepackage{times}
\usepackage{latexsym}
\usepackage{authblk}
\usepackage[T1]{fontenc}
\usepackage{amssymb}
\usepackage{graphicx}
\usepackage[many]{tcolorbox}
\usepackage{bm}
\usepackage{CJK}
\usepackage{arabtex}
\usepackage{utf8}
\setcode{utf8}
\usepackage{fancyvrb}
\usepackage{blindtext}
\usepackage{multicol}
\usepackage{makecell}

\usepackage{subcaption}

\definecolor{verylightgray}{gray}{0.95} % The closer to 1, the lighter the gray

\usepackage{fourier} 
\usepackage{array}
\usepackage[export]{adjustbox}
\usepackage{import}
\usepackage{float}
\usepackage{amsmath}
\usepackage{siunitx}
\usepackage{framed}
\usepackage{enumitem}
\usepackage{caption}
\usepackage{footnote}
\usepackage{pgfplots}
\usepackage{wasysym}
\DeclareUnicodeCharacter{03A3}{\ensuremath{\Sigma}}

\usepackage{comment}
\usepackage{pifont}
\usepackage{tocloft}

\newcommand{\xmark}{\ding{55}}%

\setlist[2]{noitemsep} % sets the itemsep and parsep for all level two lists to 0
\setitemize{noitemsep} % sets no itemsep for enumerate lists only
\setenumerate{noitemsep} % sets no itemsep for enumerate lists only

\DeclareMathOperator*{\layernorm}{LayerNorm}

\makeatletter
\renewcommand\AB@affilsepx{, \protect\Affilfont}
\makeatother

\title{Eagle and Finch: RWKV with Matrix-Valued States \\ and Dynamic Recurrence}

\author[1,2,\thanks{Equal first authorship. Others listed alphabetically.}]{\textbf{Bo Peng}}
\author[2,3,*]{\textbf{Daniel Goldstein}}
\author[2,4,23,*]{\textbf{Quentin Anthony}}
\author[2,5,6]{\authorcr \textbf{Alon Albalak}}
\author[2,7,8]{\textbf{Eric Alcaide}}
\author[2]{\textbf{Stella Biderman}}
\author[1,2,3]{\textbf{Eugene Cheah}}
\author[1]{\textbf{Xingjian Du}}
\author[9]{\authorcr \textbf{Teddy Ferdinan}}
\author[10]{ \textbf{Haowen Hou}}
\author[9]{\textbf{Przemysław Kazienko}}
\author[2,11]{\textbf{Kranthi Kiran GV}}
\author[9]{\authorcr\textbf{Jan Kocoń}}
\author[9]{\textbf{Bartłomiej Koptyra}}
\author[12]{ \textbf{Satyapriya Krishna}}
\author[2,13]{\textbf{Ronald McClelland Jr.}}
\author[24]{\textbf{Jiaju Lin}}
\author[14]{\authorcr \textbf{Niklas Muennighoff}}
\author[2]{\textbf{Fares Obeid}}
\author[2,15]{\textbf{Atsushi Saito}}
\author[2,25]{\textbf{Guangyu Song}}
\author[16,17]{\textbf{Haoqin Tu}}
\author[18]{\authorcr \textbf{Cahya Wirawan}}
\author[9]{\textbf{Stanisław Woźniak}}
\author[19]{\textbf{Ruichong Zhang}}
\author[20]{ \textbf{Bingchen Zhao}}
\author[21]{\authorcr \textbf{Qihang Zhao}}
\author[21]{\textbf{Peng Zhou}}
\author[22]{ \textbf{Jian Zhu}}
\author[17]{\textbf{Rui-Jie Zhu}}

\affil[1]{RWKV Project (under Linux Foundation AI \& Data)}
\affil[2]{EleutherAI}
\affil[3]{Recursal AI}
\affil[4]{Ohio State University}
\affil[5]{University of California, Santa Barbara}
\affil[6]{SynthLabs}
\affil[7]{Charm Therapeutics}
\affil[8]{Dalle Molle Institute for Artificial Intelligence Research}
\affil[9]{Wroclaw Tech}
\affil[10]{Guangdong Laboratory of Artificial Intelligence and Digital Economy (SZ)}
\affil[11]{New York University}
\affil[12]{Harvard University}
\affil[13]{Ronsor Labs}
\affil[14]{Contextual AI}
\affil[15]{Nextremer Co. Ltd.}
\affil[16]{University of Chinese Academy of Sciences}
\affil[17]{University of California, Santa Cruz}
\affil[18]{AI-Research.id}
\affil[19]{Tsinghua University}
\affil[20]{University of Edinburgh}
\affil[21]{LuxiTech Co. Ltd.}
\affil[22]{University of British Columbia}
\affil[23]{Zyphra}
\affil[24]{Pennsylvania State University}
\affil[25]{Tano Labs}

\begin{document}

\begin{center}
    \maketitle
\end{center}
%We introduce two novel language models, Eagle (RWKV-5) and Finch (RWKV-6), containing improvements to the RWKV \citep{peng2023rwkv} architecture. They incorporate multiheaded matrix-valued states and a dynamic recurrence mechanism while maintaining the inference efficiency characteristic of RNNs. Finch outperforms Mamba and BasedConv in multi query associative recall task. We introduce a new trillion-token multilingual corpus on which we train these models and demonstrate that they are competitive with state-of-the-art language models across a variety of benchmarks. Additionally, we introduce a new tokenizer based on greedy matching with byte-level granularity, enhancing efficiency on the multilingual corpus. We provide four Eagle models, scaling from 0.46 to 7.5 billion parameters, and two Finch models, with 1.6 and 3.1 billion parameters. Released under the Apache 2.0 license, these models are publicly accessible on HuggingFace and free for commercial use.
\begin{abstract}
We present Eagle (RWKV-5) and Finch (RWKV-6), sequence models improving upon the RWKV (RWKV-4) \citep{peng2023rwkv} architecture. Our architectural design advancements include multi-headed matrix-valued states and a dynamic recurrence mechanism that improve expressivity while maintaining the inference efficiency characteristics of RNNs. We introduce a new multilingual corpus with 1.12 trillion tokens and a fast tokenizer based on greedy matching for enhanced multilinguality. We trained four Eagle models, ranging from 0.46 to 7.5 billion parameters, and two Finch models with 1.6 and 3.1 billion parameters and find that they achieve competitive performance across a wide variety of benchmarks.
% Uncomment below only after paper is deanonymized
We release all our models on HuggingFace under the Apache 2.0 license.\footnote{Models at: \url{https://huggingface.co/RWKV}\\Training code at: \url{https://github.com/RWKV/RWKV-LM}\\
Inference code at: \url{https://github.com/RWKV/ChatRWKV}\\
Time-parallel training code at: \url{https://github.com/RWKV/RWKV-infctx-trainer}}
\end{abstract}
 
% While Finch and Mamba adopt similar elements of their architectures, Finch surpasses Mamba \citep{gu2023mamba} in multi query associative recall which is hard to learn for sub-quadratic sequence models.
% \stella{Do these models have the same num params? Same pretraining compute? Same anything?} \resp{Song's runs based on zoology repo https://github.com/HazyResearch/zoology . Num-params are small but equally set.  No pretraining is done.}
%
% 
\pagebreak
\tableofcontents
 
\section{Introduction} \label{sec:introduction}
% See Original introduction: https://www.overleaf.com/project/643b153d31b2c3bf6f4aff1c

%%%% Introduction outline (Alon)
% Paragraph 1: Introduction on transformers and their inefficiences
Advancements in Large Language Models (LLMs) have significantly impacted Natural Language Processing (NLP) tasks. The field has traditionally been dominated by the transfomer architecture \citep{vaswani2023attention}. However, the expressive attention mechanism of transformers leads them to suffer from quadratic time complexity with respect to input sequence length. Various methods have been proposed to achieve sub-quadratic time complexity without significantly changing the core attention mechanism, typically relying on some form of sparsity techniques~\citep{child2019generatingsparse, beltagy2020longformer, zaheer2020bigbird}.

Recent works have achieved sub-quadratic time complexity without significantly sacrificing performance by introducing new mechanisms to replace attention at the core of the Transformer architecture. These models include gated recurrences ~\citep{fu2023hungry,gu2023mamba,gu2021s4,sun2023retentive,katsch2023gateloop, qin2023hierarchically, smith2023s5}, gated convolutions~\citep{poli2023hyena, peng2023rwkv}, data-dependent linear attention~\citep{yang2023gated, linearTrans2020inputDepend}, sparse attentions~\citep{tay2020sparse,child2019sparsetransformer,zaheer2020bigbird,blockwindow2019sparsetransformer} and their combinations \citep{de2024griffin, qin2024transnormerllm, qin2022deviltransnormer}. We build off RWKV-4 introduced in \cite{peng2023rwkv}, which provides efficient inference and training along with a parallelizable implementation compared to competing architectures as shown in Table \ref{tab:costs}.
%IN PROGRESS...
% we propose: (1) more expressive matrix valued states, (2) 
% However, prior approaches to linear attention have proposed relatively rigid communication mechanisms, where limited information-mixing occurs between channels, tokens, and layers, leading to the desire for more flexible and expressive communication mechanisms.

\begin{table}[h]
\centering
\small
\begin{tabular}{lccccc}
\toprule
\textbf{Architecture} & \multicolumn{2}{c}{\textbf{Inference}} & & \multicolumn{2}{c}{\textbf{Training}} \\
& \textbf{Time} & \textbf{Memory} &  \textbf{Parallel} & \textbf{Time} & \textbf{Memory}\\
\midrule
LSTM/LMU                &$O(1)$ & $O(1)$   & \xmark & $O(N)$     & $O(N)$   \\
% LMU                 & $O(1)$\!& $O(1)$   & \xmark & $O(N)$     & $O(N)$   \\
%HiPPO               & $O(1)$\!& $O(1)$   & \xmark & $O(N)$     & $O(N)$   \\
%Vanilla Transformer & $O(N)$\!& $O(N)$   & \checkmark & $O(N^2)$   & $O(N^2)$   \\
Transformer             & $O(N)$    & $O(N)^a$  & \checkmark & $O(N^2)$   & $O(N)^b$     \\
Linear  Transformer  & $O(1)$    & $O(1)$   & \checkmark & $O(N)$     & $O(N)$   \\
H3/S4               & $O(1)$    & $O(1)$ & \checkmark & $O(N \log N)$ & $O(N)$   \\
Hyena               & $O(N) $    & $O(N)$  & \checkmark & $O(N \log N)$ & $O(N)$   \\
RWKV/Mamba/RetNet        & $O(1)$    & $O(1)$ & \checkmark & $O(N)$ & $O(N)$    \\
% RetNet              & $O(1)$    & $O(1)$ & \checkmark & $O(N)$ & $O(N)$          \\
%RWKV-4              & $O(1)$    & $O(1)$ & \checkmark & $O(N)$ & $O(N)$          \\
% RWKV-4/Eagle/Finch         & $O(1)$    & $O(1)$ & \checkmark & $O(N)$ & $O(N)$          \\
\bottomrule
\end{tabular}
\vspace{5pt} % Adjust the value as needed to control the space
\caption{Comparative analysis of RWKV-4/5/6 and other LLM architectures regarding time and memory complexity for both inference per token and training per sequence, and training parallelizability across the sequence dimension. The context/sequence length is denoted by $N$.\\\footnotesize{$^a$$O(1)$ without KV cache $^b$ With Flash Attention}
% \footnotesize{$^b$ O(N) is worst case complexity. Hyena has time-space trade-offs depending on its order of hyena-operators.}
}
\label{tab:costs}
\end{table}

% RWKV \citep{peng2023rwkv}, influenced by the Attention Free Transformer (AFT) \citep{zhai2021attention}. RWKV demonstrates that RNN-based models can scale to over ten billion parameters while remaining competitive in performance to similarly sized transformers in NLP benchmarks. RWKV has been used in a range of applications, including streaming Automatic Speech Recognition (ASR)  \citep{an2023exploring} and language translation \citep{10.1007/978-981-99-8132-8_25}.

% Paragraph 3:
% How does this work improve on the previous methods (data-dependent decay, hardware-aware improvements [improved parallelization over non-time dimensions with a custom CUDA kernel], other architectural changes)
In this paper, we introduce two new architectures: \textbf{Eagle} (RWKV-5) and \textbf{Finch} (RWKV-6). First, Eagle improves upon the architecture and learned decay schedule from RWKV-4~\citep{peng2023rwkv} through the use of expressive multi-headed matrix-valued states (as opposed to vector-valued states), a reformulated receptance, and an additional gating mechanism. 
% Building upon the architectural improvements from RWKV-4 to Eagle, 
Finch further improves the expressivity and flexibility of the architecture by introducing new data-dependent functions for both the time-mixing and token-shift modules, consisting of parameterized linear interpolations. Additionally, Finch proposes a novel use of the Low Rank Adaptation~\citep{hu2022lora} function to allow for trainable weight matrices to efficiently augment the learned data decay vectors in a context-dependent manner.
% New tokenizer and dataset
Finally, we introduce a new tokenizer, the RWKV World Tokenizer, and a new dataset, RWKV World v2 (1.12 trillion tokens), specially designed to improve performance on multilingual and code data.

%Despite Mamba \citep{gu2023mamba} being developed concurrently with Finch and containing some similar architectural elements, 
%While certain related architectural changes are proposed in \citep{gu2023mamba}, Finch outperforms Mamba and competes closely with quadratic attentions in MQAR task, while maintaining time complexity that is linear for training and constant per token for inference.(See Table \ref{tab:costs} for a comparison.)  

% Concurrent works including GateLoop \citep{katsch2023gateloop}, Mamba \citep{gu2023mamba}, GLA \citep{yang2023gated} and Griffin \citep{de2024griffin} employ variations of some of these architectural changes, including matrix-valued states, forms of data-dependence, and/or learned decay schedules. 
%We introduce a tokenizer and a dataset to facilitate the scalability of our models for tackling diverse tasks as fllows: \textbf{(1) RWKV World Tokenizer} designed for multilinguality and source codes and \textbf{(2) RWKV World v2 Dataset} which is a new  multilingual training corpus with 1.12 trillion tokens.

% \stella{This is a sudden and abrupt change that needs to be smoothed out. In particular, its the first mention of languages}

% Paragraph 4:
% How do we demonstrate that the above improves on the prior works
%% ALON: we can probably wait until we have more updated results to write this part
% We show that...

Through extensive experimentation, we show that the Eagle and Finch models perform competitively, or improve upon existing models under a wide variety of sequence modeling domains and tasks. Specifically, we evaluate our trained models on commonly used English-only and multilingual text benchmarks, associative recall, music modeling, and vision-language benchmarks. Our experiments demonstrate that the advancements in Eagle and Finch provide significant progress towards developing more efficient AI models

In summary, our main contributions are:
\begin{itemize}
    \item The Eagle (RWKV-5) and Finch (RWKV-6) RWKV architectures, which significantly improve over RWKV-4 on benchmarks for LLMs.
    \item The RWKV World Tokenizer which contains underrepresented languages' vocabulary and which performs fast tokenization with Trie-based greedy matching.
    % which increases the average information density per token for non-English languages to better support multilingual capabilities
    \item The RWKV World v2 public dataset, comprised of 1.12 trillion tokens of publicly available multilingual data.
    \item Public release of four pre-trained Eagle models, scaling from 0.46 to 7.5 billion parameters, and two Finch models, with 1.6 and 3.1 billion parameters. Demonstrating that these novel architectures are competitive to transformers when trained using enough FLOPs to make meaningful scaling conclusions.
    \item A completely open training pipeline to enable interpretability and reproducibility of alternative-architecture LLMs (See Table \ref{tab:compare}).
    % copied from abstract but I'm not sure about those numbers ^
\end{itemize}

\begin{table}[h]
\centering
\small
\begin{tabular}{lcccccccc}
\toprule
\textbf{Model} & \textbf{Context} & \textbf{Training} & \textbf{Open} & \multicolumn{2}{c}{\textbf{Open Code}} & \textbf{Open} \\
& \textbf{Length} & \textbf{Tokens} & \textbf{Weights} & \textbf{Inference} & \textbf{Training} & \textbf{Dataset} \\
\midrule
GPT-4            & 128k$^a$ & Undisclosed        & \Circle      & \Circle       & \Circle      & \Circle \\
LLaMA2 7B        & 4k      & \num{2.0e12}           & \LEFTcircle      & \CIRCLE   & \Circle      & \Circle \\
Mistral 7B v0.1  & 32k$^b$ & Undisclosed         & \CIRCLE  & \CIRCLE   & \Circle      & \Circle \\
Gemma 7B         & 8k      & \num{6.0e12}    & \LEFTcircle & \CIRCLE & \CIRCLE & \Circle \\
StableLM 7B v2   & 4k      & \num{1.1e12}          & \CIRCLE  & \CIRCLE   & \CIRCLE  & \CIRCLE \\
Pythia 6.9B      & 2k      & \num{3.3e11}           & \CIRCLE  & \CIRCLE   & \CIRCLE  & \CIRCLE \\
Eagle 7B         & Indefinite$^c$ & \num{1.1e12}   & \CIRCLE  & \CIRCLE   & \CIRCLE  & \CIRCLE \\
\bottomrule
\end{tabular}
\vspace{5pt}
\caption{{Comparison of the openness and accessibility of public foundational LLMs with 7B+ parameters regarding model weights, official inference/training code, and dataset. Widely available but not under an open source license is indicated by $\LEFTcircle$.\\
\footnotesize{$^a$OpenAI's gpt-4-0125-preview model} \footnotesize{$^b$With sliding window attention} \footnotesize{$^c$Pretrained with context length 4096, but no fundamental context length limitation or relationship to speed, see ~\ref{subsec:long_context_experiments} for extrapolation details}}}
\label{tab:compare}
\end{table}

\section{Background} \label{sec:background}

%\stella{This has a terrible reading flow. It feels more like a list of definitions than a meaningful backgrounder. A background section should be narrative.}

Eagle and Finch are RNNs based on a multi-headed hybridization of the RWKV-4 architecture and linear attention. We discuss related work and the evolution of these two architectures below, with a more detailed review given in Appendix ~\ref{app_related_work}.

%\paragraph{Recurrent Neural Networks (RNNs)}

Recurrent Neural Networks (RNNs) are well suited to provide inexpensive inference on sequence modelling tasks, typically operating in O(1) time complexity per step with respect to sequence length. They model sequences with time dependencies by generating a hidden state $h_t$ at each time step, which is fed back in at the next time step as a secondary input. Classic RNNs (e.g. LSTM~\citep{hochreiter1997long} and GRU~\citep{cho2014learning}) became widely used for sequence modelling, but are difficult to parallelize across the time dimension for training.

%\paragraph{Transformers}
%
The Transformer architecture has enjoyed remarkable success in generative sequence modelling, and language modelling in particular~\citep{vaswani2023attention, radford2018improving}, providing SOTA performance across many tasks. However, the use of multi-headed dot-product self-attention (MHA) leads to a quadratic time complexity with respect to sequence length.  The deficiencies of classic RNNs and Transformers led to many attempts to develop architectures incorporating the best features of both in a single model, namely O(1) per token time complexity and fast highly parallelizable training.

%\paragraph{Linear Attention}

Linear Attention~\citep{schmidhuber1992unnormedlinattn, katharopoulos2020lineartransformrers} replaces the numerator of MHA's $\mathrm{softmax}(QK^T)V$ with $\phi(Q)\phi(K)^\mathrm{T}V$, allowing a reordering of operations via associativity to $\phi(Q)(\phi(K)^\mathrm{T}V)$, where $\phi$ represents a non-negative feature-map function.  It can be computed as an RNN in $O(1)$ time per step by adding $\phi(K_i^T)V_i$ to a recurrent state at each time step $i$, or trained in parallel much like MHA. This accomplishes the main goals outlined above, but naive linear attention suffers from significantly reduced performance compared to MHA-based transformers. 

%\paragraph{Attention Free Transformer (AFT)}

A modified form of linear attention, the Attention Free Transformer (AFT) \citep{zhai2021attention}, paved the way for the RWKV architecture, by using a number of attention heads equal to the size of the feature dimension and incorporating a set of learned pairwise positional biases, denoted as $w$.  

%The Attention Free Transformer (AFT) \citep{zhai2021attention} introduces a modified form of linear attention, where the number of attention heads is equal to the size of the feature dimension. It also incorporates a set of learned pairwise positional biases, denoted as $w$. This innovation paved the way for the RWKV architecture. %The AFT can be conceptualized as calculating a per-channel weighted average of values. The weight for a specific location is determined by the sum of the key at that location and the corresponding learned positional bias.

\begin{align}
\mathrm{AFTAttn}_t &= \sigma_q(q_t) \odot \frac{\sum_{i=1}^{t} \exp(k_i + w_{i,t}) \odot v_i}{\sum_{i=1}^{t} \exp(k_i + w_{i,t})}
\end{align}
%In the equation, the term $\sigma_q(q_t)$ represents the activation function applied to the query vector at the same time step. The symbol $k_i$ is used to represent the key at the location $i$, while $v_i$ stands for the value at the same location. The learned position bias for location $i$ at time step $t$ is represented by $w_{i,t}$.

%\paragraph{RWKV-4}

RWKV-4 reformulates the AFT equation by replacing the pair-wise positional biases with a channel-wise vector of additive weight decay rates $w$. It also adds a bonus term $u$ to offset the weight of only the current input specially.

\begin{align}
    \mathrm{wkv}_t &= \frac{ \sum_{i=1}^{t-1} \exp(-(t-1-i)w+k_i) \odot v_i + \exp(u+k_t) \odot v_t}{\sum_{i=1}^{t-1} \exp(-(t-1-i)w+k_i) + \exp(u+k_t)}.
\end{align}

RWKV-4 also adds token-shift and gating to both attention and feed-forward sub-blocks of transformer, and small embedding initialization and normalization to quickly arrive at well-distributed token embeddings. Combining all of these architectural changes led RWKV-4 to become the first RNN to rival the performance of Transformers, while maintaining fast parallelizable training and $O(1)$ time complexity per token.

There has been a recent revival of RNNs in NLP research \citep{tiezzi2024resurgence}. HGRN\citep{qin2023hierarchically} is a recent time-parallelizable data-dependent RNN that employs input and forget gates. TransNormer\citep{qin2022deviltransnormer} applies RMSNorm to linear attention to bound its output. Other new time-parallelizable data-dependent RNNs have also been invented concurrently with our work including GLA \citep{yang2023gated} and Griffin \citep{de2024griffin}. 

State Space Models (SSMs) employ a hidden state of basis function weights to model an approximation of the input function \citep{gu2020hippo}, updating that hidden state via a differential equation. Earlier SSMs \citep{gu2022efficiently} were historically computed using long convolutions in $O(N\log N)$ time per sequence, but could also be formulated as a recurrent network. Recently, it has been shown that SSMs can be parallelized across the time dimension via techniques including associative scan \citep{smith2023s5}.  A new class of SSMs has also emerged concurrently with our work \citep{katsch2023gateloop,gu2023mamba} that feature data-dependent $A$ and $B$ terms, which function similarly to the data-dependent dynamic recurrence used in Finch.

\section{Eagle/Finch Architecture}\label{sec:rwkv_model}
We refine the RWKV architecture in two steps, and observe significant modeling improvements with each. Compared to the baseline RWKV-4, Eagle adds matrix-valued attention states, LayerNorm over the attention heads, SiLU attention gating, and improved initialization. It also removes the Sigmoid activation of receptance. Finch further applies data-dependence to the decay schedule and token-shift.

\begin{figure}[hbt]
    \centering
    \includegraphics[width=\linewidth]{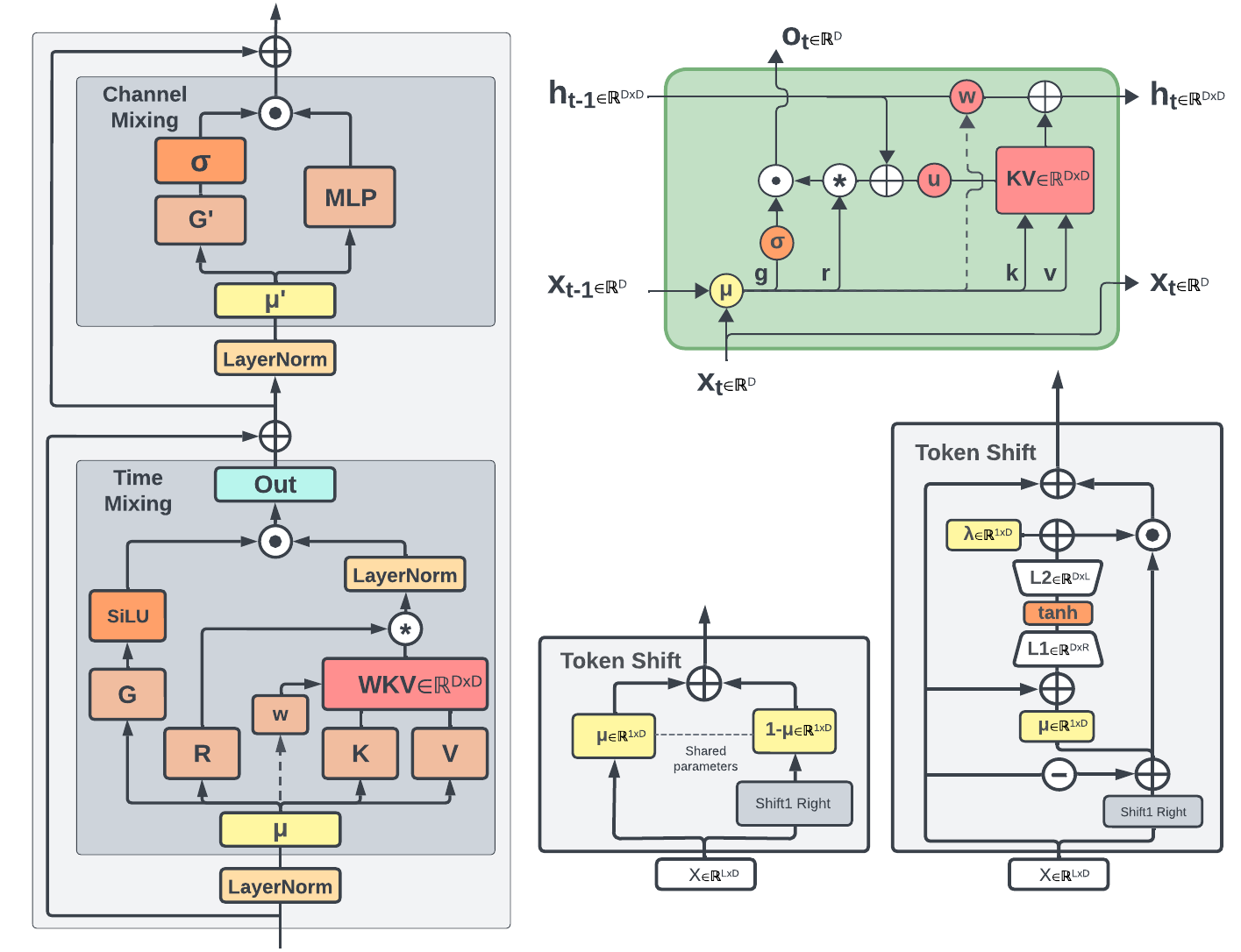}
    \caption{RWKV architecture overview. \textbf{Left:} time-mixing and channel-mixing blocks; \textbf{top-right:} RWKV time-mixing block as RNN cell; \textbf{center-bottom:} token-shift module in FeedForward module and Eagle time-mixing; \textbf{bottom-right:} token-shift module in Finch time-mixing. All shape annotations assume a single head for simplicity. Dashed arrows (left, top-right) indicate a connection in Finch, but not in Eagle.}
    \label{fig:blocks}
\end{figure}

The core architecture remains similar to that of RWKV-4, consisting of a series of stacked residual blocks shaped like a traditional Transformer. Following notation from \citep{NEURIPS2021mixer}, each block contains one Pre-LayerNorm Time-Mixing sub-layer followed by one Pre-LayerNorm Channel-Mixing sub-layer, as depicted in \hyperref[fig:blocks]{Figure \ref*{fig:blocks}}, left. These correspond to the traditional Attention and Feed Forward Network sub-layers of the Transformer. See Appendix \ref{app_architecture} for more details on our training implementation and the differences from RWKV-4, and Section \ref{sec:Speed and Memory Benchmarks} for speed and memory benchmarks.

\section{Method}\label{sec:method}
In this section, we use $D$ to denote the model dimension, and unless explicitly stated, all vectors appearing in this section are dimension $D/h$, where $h$ denotes the number of heads, belonging to $\mathbb{R}^{(D/h)}$. For compactness and simplicity we show calculations per-head, eliding the head index. We use the convention that all vectors are row vectors unless explicitly transposed, so all matrices operate on the right side. We use the square subscript to denote a variable.

\subsection{Eagle}

\subsubsection{Eagle Token Shift}

We adopt the Token Shift technique from the previous RWKV, similar to a 1D causal convolution of size = 2, as can be seen in \hyperref[fig:blocks]{Figure \ref*{fig:blocks}}, center-bottom. To better introduce the Token Shift technique, we define some notation. The \textbf{l}inear int\textbf{erp}olation (lerp) between $x_t$ and $x_{t-1}$ used in RWKV-4 and Eagle Token Shift is defined as:
\begin{align}
\label{eq:lerp5}
\mathrm{lerp}_\square (a, b) &= a + (b-a) \odot \mu_\square
\end{align}
where each $\mu_\square\in\mathbb{R}^{D}$ is a learnable vector.

Token Shift allows the model to learn how much new versus old information should be allocated per time step to each channel of receptance, key, value, and gate vectors ($r$, $k$, $v$, and $g$ respectively) independently and uniquely for each head. This makes it possible to form induction heads \citep{elhage2021mathematical} within a single layer since even a single head can directly accumulate both past and current token data into separate subspaces within these vectors.

\subsubsection{Eagle Time Mixing}

The formula of Eagle Time Mixing can be written as follows:

% \begin{align}
% \label{eq:token-shift5}
% r_t &= \mathrm{lerp}_r(x_t, x_{t-1})\bm{W}_r\\
% k_t &= \mathrm{lerp}_k(x_t, x_{t-1})\bm{W}_k\\
% v_t &= \mathrm{lerp}_v(x_t, x_{t-1})\bm{W}_v\\
% g_t &= \mathrm{lerp}_g(x_t, x_{t-1})\bm{W}_g\\
% w &= \exp(-\exp(\omega))
% \end{align}

% Alon's attempt at abbreviating the equations:
\begin{align}\label{eq:token-shift5}
\square_{t} &= \mathrm{lerp}_{\square}(x_t, x_{t-1})\bm{W}_{\square}, \quad \square\in \{r,k,v,g\}\\
w &= \exp(-\exp(\omega))
\end{align}
\begin{align}
\label{eq:time-mix5}
\bm{wkv}_{t} &= \mathrm{diag}(u)\cdot k_{t}^\mathrm{T} \cdot v_{t} + \sum_{i=1}^{t-1} \mathrm{diag}(w)^{t-1-i} \cdot  k_{i}^\mathrm{T} \cdot v_{i}  \in \mathbb{R}^{(D/h) \times (D/h)}  \\
  \label{eq:time-mixer5out}
o_t &= \mathrm{concat}\left(\mathrm{SiLU}(g_t) \odot \layernorm(r_{t}  \cdot \bm{wkv}_{t})\right) \bm{W}_o \in \mathbb{R}^{D}
\end{align}
Where LayerNorm operates on each of $h$ heads separately, which is also equivalent to the GroupNorm (\cite{wu2018group}) operation on $h$ groups. It is also worth noting that $w$ is obtained from $w = \exp(-\exp(\omega))$, where $\omega \in \mathbb{R}^{D/h}$ are the actual headwise trainable parameters. This ensures that $w$ falls within the interval $(0,1)$, guaranteeing that $\mathrm{diag}(w)$ is a contraction matrix. 

The $\bm{wkv}_{t}$ attention calculation can alternatively be written in a recurrent form:
\begin{align}
\label{eq:time-mixer5}
\bm{wkv}' &= \bm{s} + \mathrm{diag}(u) \cdot k^\mathrm{T} \cdot v \\
\label{eq:time-mixer5state}
\bm{s}' &= \mathrm{diag}(w) \cdot \bm{s} + k^\mathrm{T} \cdot v
\end{align}

RWKV's $\bm{wkv}$ term can be considered a decay-based equivalent to the normalised $k^\mathrm{T}v$ term in Linear Attention. It is instructive to note how for a given head $j$ the recurrent state $s$ is a sum of $k^Tv$ where each channel of $s$ individually decays by the corresponding channel of $w$ at each time step. Prior to the application of the receptance vector, gating, and output weights, a per-channel learned boost $u$ is multiplied with the current token's $k^\mathrm{T}v$ and summed with the state, as can be seen in \hyperref[fig:blocks]{Figure \ref*{fig:blocks}}, top-right. This gives the current token special treatment relative to the sum of past tokens contained within the decaying state history. The receptance is multiplied by this sum, acting like the query term in Linear Attention.

\subsubsection{Channel Mixing}
In both Eagle and Finch, the Channel Mixing module is identical to the previous RWKV-4 architecture, except for a slightly reduced hidden dimension from $4D$ to $3.5D$. This reduction accounts for new gating weights in Eagle Time Mixing to ensure an equi-parameter relation with the prior model at the same number of layers and embedding dimension. We do not further reduce the hidden dimension in Finch despite adding a small number of new parameters for LoRA weights. The formulas for Channel Mixing are the same as RWKV-4, but we restate them here to ensure notational consistency, using linear interpolation from Equation~\ref{eq:lerp5}:
\begin{align}
\label{eq:channel-mix5}
r'_t &= \mathrm{lerp}_{r'} (x'_t, x'_{t-1}) \bm{W}_{r'}\in \mathbb{R}^{D}\\
k'_t &= \mathrm{lerp}_{k'} (x'_t, x'_{t-1}) \bm{W}_{k'}\in \mathbb{R}^{3.5D}\\
v'_t &= \mathrm{ReLU}(k'_t)^2 \bm{W}_{v'}\in \mathbb{R}^{D}\\ 
o'_t &= \sigma(r'_t) \odot v'_t\in \mathbb{R}^{D}
\end{align}

\subsection{Finch}

\subsubsection{Finch Token Shift}

The \textbf{d}ata-\textbf{d}ependent \textbf{l}inear int\textbf{erp}olation (ddlerp) between $x_t$ and $x_{t-1}$ used in Finch Token Shift is defined as:
\begin{align}
\mathrm{lora}_\square(x) &= \lambda_\square + \tanh(x\bm{A}_\square)\bm{B}_\square \\
\mathrm{ddlerp}_\square (a, b) &= a + (b-a) \odot \mathrm{lora}_\square(a + (b-a) \odot \mu_x)
\end{align}
where $\mu_x$ and each $\lambda_\square$ introduce a trainable vector of dimension $D$ and each $\bm{A}_\square \in \mathbb{R}^{D \times 32}$, $\bm{B}_\square \in \mathbb{R}^{32 \times D}$ introduce new trainable weight matrices. For the special case of $\mathrm{LoRA}_\omega$ seen below we introduce double-sized trainable weight matrices $\bm{A}_\omega \in \mathbb{R}^{D \times 64}$, $\bm{B}_\omega \in \mathbb{R}^{64 \times D}$. A schematic representation can be found in \hyperref[fig:blocks]{Figure \ref*{fig:blocks}}, bottom-right. Please note that future 7B and larger Finch models are expected to further increase the size of these weight matrices by double or more.

This new form of Token Shift enhanced with data-dependence is intended to expand the abilities of the model beyond the RWKV-4/Eagle style of Token Shift so that the amount of new and old data allocated per channel now depends on the input at both current and prior time steps. %Intuitively, important information can effectively flag itself for inclusion using this mechanism, and less important information can flag itself to partially or fully avoid entering the data stream, leaving room for more important pre-existing data to remain. Viewed from the perspective of induction heads, this could allow for potential misleading matches to be pre-filtered out up front if they are not deemed useful for a given task.

\subsubsection{Finch Time Mixing}

% \begin{align}
% r_t &= \mathrm{ddlerp}_r ( x_t, x_{t-1} ) \bm{W}_r\\
% k_t &= \mathrm{ddlerp}_k ( x_t, x_{t-1} ) \bm{W}_k\\
% v_t &= \mathrm{ddlerp}_v ( x_t, x_{t-1} ) \bm{W}_v\\
% g_t &= \mathrm{ddlerp}_g ( x_t, x_{t-1} ) \bm{W}_g\\
% d_t &= \mathrm{lora_\omega}( \mathrm{ddlerp}_d ( x_t, x_{t-1} ) )\\
% w_t &= \exp(-\exp(d_t))
% \end{align}
% \begin{align}
% \bm{wkv}_{t} &=  \mathrm{diag}(u)\cdot k_{t}^\mathrm{T} \cdot v_{t} + \sum_{i=1}^{t-1}  \mathrm{diag}\left(\bigodot_{j=1}^{i-1}w_{j}\right) \cdot  k_{i}^\mathrm{T} \cdot v_{i}  \in \mathbb{R}^{(D/h) \times (D/h)} \\
% o_t &= \mathrm{concat}\left(\mathrm{SiLU}(g_t) \odot \layernorm(r_{t} \cdot \bm{wkv}_{t})\right)\bm{W}_o \in \mathbb{R}^{D} 
% \end{align}

% Alon's attempt at abbreviating the equations:
\begin{align}\label{eq:token-shift6}
\square_{t} &= \mathrm{ddlerp}_{\square}(x_t, x_{t-1})\bm{W}_{\square}, \quad \square\in \{r,k,v,g\}\\
d_t &= \mathrm{lora}_d( \mathrm{ddlerp}_d ( x_t, x_{t-1} ) )\\
w_t &= \exp(-\exp(d_t))
\end{align}
\begin{align}
\bm{wkv}_{t} &=  \mathrm{diag}(u)\cdot k_{t}^\mathrm{T} \cdot v_{t} + \sum_{i=1}^{t-1}  \mathrm{diag}\left(\bigodot_{j=i+1}^{t-1}w_{j}\right) \cdot  k_{i}^\mathrm{T} \cdot v_{i}  \in \mathbb{R}^{(D/h) \times (D/h)} \\
o_t &= \mathrm{concat}\left(\mathrm{SiLU}(g_t) \odot \layernorm(r_{t} \cdot \bm{wkv}_{t})\right)\bm{W}_o \in \mathbb{R}^{D} 
\end{align}

The $\bm{wkv}_{t}$ attention calculation can alternatively be written in a recurrent manner:
\begin{align}
\label{eq:time-mixer6}
\bm{wkv}' &= \bm{s} + \mathrm{diag}(u) \cdot k^\mathrm{T} \cdot v \\
\bm{s}' &= \mathrm{diag}(w) \cdot \bm{s} + k^\mathrm{T} \cdot v
\end{align}
Unlike in Eagle, $w_t$ here is not static across the sequence (dashed arrows in \hyperref[fig:blocks]{Figure \ref*{fig:blocks}}, left and top-right.). This is the core change to decay in Finch, as each channel of $w_t$ can now vary independently over time, in a data-dependent manner, whereas previously it was a fixed learned vector.

The new LoRA mechanisms above are used to take learned vectors, as seen in Eagle, and inexpensively augment them with additional offsets determined by the incoming input. Note that the LoRA process itself uses an Eagle style Token-Shifted value as its input, not just the latest token. The new time-varying decay $w_t$ goes one step further, applying LoRA again afterward. Intuitively, this is a second-order variant of Token-Shifting, allowing each channel of $w_t$ to vary based on a mix of the current and prior tokens, with the mix itself determined by aspects of both tokens.

% \subsubsection{Finch Channel Mixing}

% In Finch, the Channel Mixing module changes slightly to use our new data-dependent token-shift mechanism. This adds a small number of parameters.

% \begin{align}
% \label{eq:channel-mix6}
% r'_t &= \mathrm{ddlerp}_{r'} (x'_t, x'_{t-1}) \bm{W}_{r'}\\
% k'_t &= \mathrm{ddlerp}_{k'} (x'_t, x'_{t-1}) \bm{W}_{k'}\in \mathbb{R}^{3.5D}\\
% v'_t &= \mathrm{ReLU}(k'_t)^2 \cdot \bm{W}_{v'} \\
% o'_t &= \sigma(r'_t) \odot v'_t
% \end{align}

\section{RWKV World Tokenizer}\label{sec:tokenizer}
Tokenization is important in language modelling as it conditions the learning relationships between tokens and the generation of new text based on those patterns. The numbers of tokens to build a single semantic chunk are, however, often very unequally distributed against non-European and other underrepresented languages. Byte-pair-encoding (BPE) based tokenizers which are trained with this inequality result in not only lower performances against underrepresented languages but also undue economic costs such as inference~\cite{ahia2023languageCost} and continual pre-training  with extended vocabulary~\cite{mala500lin2024, elyzallama2023}. To address these problems, we manually select tokens from multiple vocabulary files such that non-European languages are well represented.

% \section{New Tokenizer Details} \label{app_tokenizer}
To construct the tokenizer's vocabulary, we merge the vocabularies of the following tokenizers and then manually select the tokens for non-European languages.
\begin{itemize}
    \item \textbf{GPT-NeoX-20B \citep{black-etal-2022-gpt}:} \url{https://huggingface.co/EleutherAI/gpt-neox-20b}
    \item \textbf{GPT2 \citep{radford2019language}:} \url{https://huggingface.co/openai-community/gpt2}
    \item \textbf{cl100k\_base of tiktoken:} \url{https://github.com/openai/tiktoken}
    \item \textbf{Llama2 \citep{touvron2023llama}:} \url{https://huggingface.co/meta-llama/Llama-2-7b-hf}
    \item \textbf{Bloom \citep{workshop2023bloom}:} \url{https://huggingface.co/bigscience/bloom}
\end{itemize}

This tokenizer has a vocabulary size of $V=65536$, numbered from 0 through 65535, where tokens are arranged by their lengths in bytes. Below is a brief overview:
\begin{itemize}
    \item \textbf{Token 0:} Represents the boundary between text documents, known as \texttt{<EOS>} or \texttt{<SOS>}. This token doesn't encode any specific content and is only used for document separation.
    \item \textbf{Tokens 1-256:} Consist of byte encodings (Token $k$ encodes byte $k-1$), wherein tokens 1-128 correspond to standard ASCII characters.
    \item \textbf{Tokens 257-65529:} Tokens with a minimum length of 2 bytes in UTF-8, including words, prefixes and suffixes, accented letters, Chinese characters, Hangul, Hiragana, Katakana and emojis. For example, Chinese characters are allocated from token 10250 to 18493.
    \item \textbf{Token 65530-65535:} Reserved tokens for future use.
\end{itemize}
These designations are intended to enhance the tokenizer's efficiency on the multilingual corpus, as well as on source code of programming languages.

This tokenizer is implemented via a Trie (Prefix Tree) to boost speed while maintaining simplicity. Encoding is performed as matching the longest element in vocabulary with an input string from left to right. We note that our tokenizer's vocabulary construction is to mitigate \textit{undue} burden, which naive BPE and related methods cause, on minor languages.

%  \textcolor{red}{TODO: Explain why we want to avoid BPE}

\section{RWKV World v2 Dataset}\label{sec:dataset}

We train our models on the new \textbf{RWKV World v2 Dataset}, a new multilingual 1.12 trillion token dataset drawn from a wide variety of hand selected publicly available data sources. This dataset is designed to go beyond the English-heavy focus of many datasets widely used to train LLMs today. We do this to support usage by the majority of the worldwide population who are not native English speakers, to improve representation within model responses, and also to enable transfer learning so that our models can apply knowledge across cultures and locales. We put a strong emphasis on factual knowledge and code, but also on cultural works including stories, books, subtitles, and conversations. The source data is approximately 70\% English, 15\% multilingual, and 15\% code. We describe the components of our dataset in detail in
\hyperref[sec:training_dataset_details]{Appendix \ref*{sec:training_dataset_details}}.

\section{Pre-Trained Models}
\label{sec:pretrained_models}

We have pre-trained and publicly released the six Apache 2.0 licensed Eagle and Finch models: \textbf{Eagle 0.4B}, \textbf{Eagle 1.5B}, \textbf{Eagle 3B}, \textbf{Eagle 7B}, \textbf{Finch 1.6B}, and \textbf{Finch 3B}. All of the models were trained on the 1.12 trillion token RWKV World v2 multilingual corpus. See \hyperref[flop_count]{Appendix \ref*{flop_count}} for detailed parameter counts and FLOPs calculations.

%Further details on hyperparameters are in Appendix~\ref{appendix_hyperparam}. 
% Learning rate: 1x, 2x, 3x for different parameter sets.
% Diverting from standard practice for transformers, we apply exponential decay to our learning rate. 
% We also incorporate the auxiliary loss introduced by PaLM \cite{PaLM}, supplementing the standard cross-entropy loss function. 
% This auxiliary loss encourages the softmax normalizer to approximate zero closely. As for the learning rate schedule, it remains constant for the initial iterations, and subsequently decays exponentially. 

\section{Language Modeling Experiments}\label{evaluations}

%To assess the performance of Eagle and Finch models, we evaluate on a series of common multi-lingual and English-focused benchmarks using lm\_evaluation\_harness. These benchmarks were designed to evaluate various aspects of language understanding such as zero-shot learning, reasoning and long range dependencies \citep{bisk2020piqa, sakaguchi2021winogrande, liu2023alignbench}. The following subsection introduces the benchmarks used and the corresponding results.

\subsection{LM Evaluation Harness Benchmarks}\label{subsec:evals}
% \subsection{Memory \& Long Range Dependency Benchmarks}

%The following are a set of experiments run using lm\_evaluation\_harness, split into common multi-lingual and English-focused benchmarks (see Table~\ref{tab:multilang_bench} and~\ref{tab:eng_bench}). 

To assess the performance of Eagle and Finch models, we evaluate on a series of common multi-lingual and English-focused benchmarks using lm\_evaluation\_harness \citep{gao10256836framework} as shown in Tables~\ref{tab:multilang_bench} and~\ref{tab:eng_bench}. We find that Eagle and Finch demonstrate exceptionally high capabilities on multi-lingual benchmarks, with nearly all results significantly outperforming the other similarly sized models we tested.

\begin{figure}[htbp]
\centering
\includegraphics[width=.9\textwidth]{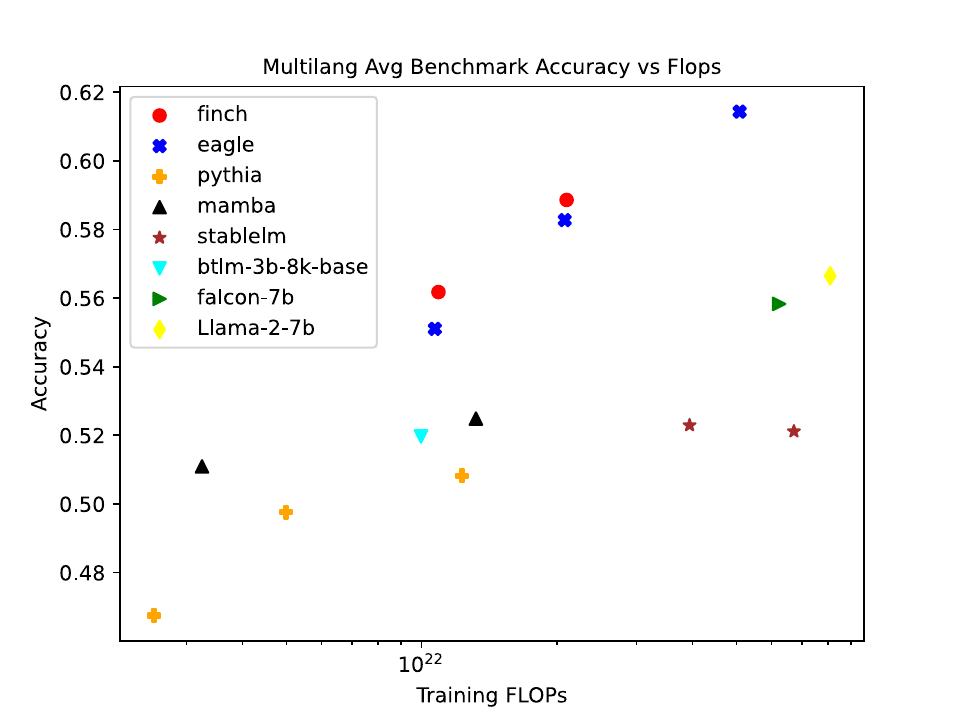}
\caption{\centering{Multilingual average benchmark accuracy versus training FLOPs. Average of LAMBADA Multilingual, xStoryCloze, xWinoGrande, and xCOPA}}
\label{fig:multilang-flops}
\end{figure}

\begin{figure}[htbp]
\centering
\includegraphics[width=.9\textwidth]{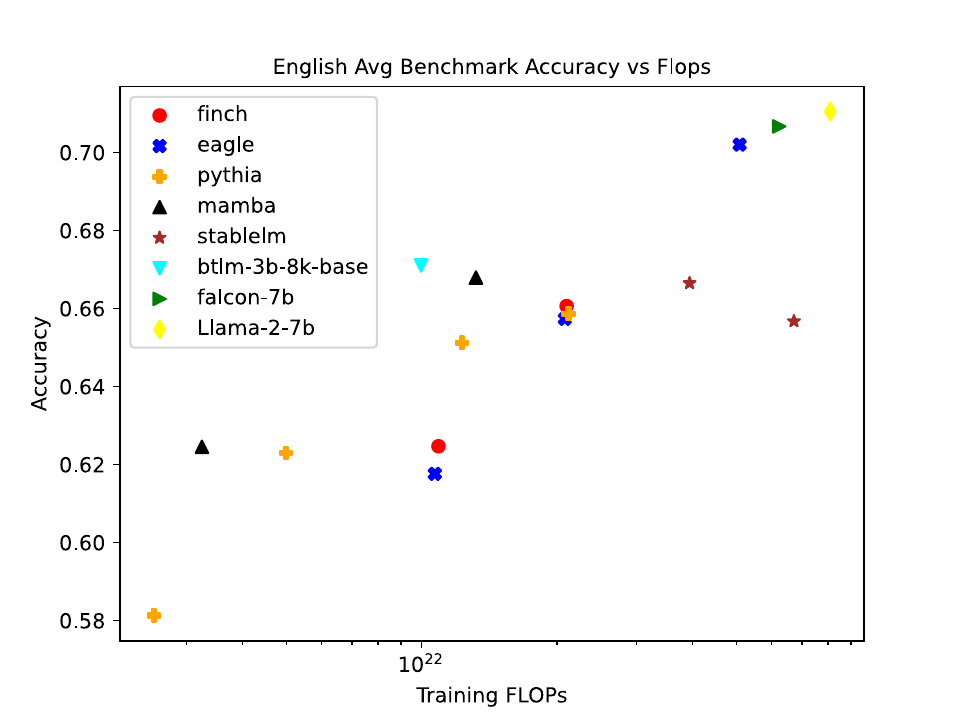}
\caption{\centering{English average benchmark accuracy versus training FLOPs. Average of LAMBADA (OpenAI), PIQA, StoryCloze16, HellaSwag, WinoGrande, Arc (challenge), Arc (easy), HeadQA (English), OpenBookQA, SciQ, ReCoRD and COPA}}
\label{fig:eng-flops}
\end{figure}

%In appendix \ref{sec:benchmarks-versus-flops}, we plot the accuracy versus FLOPs used to train various open models across a similar set of common benchmarks, and find that Eagle and Finch set a new SOTO Pareto frontier on these multilingual benchmarks. The two models additionally obtain a Pareto improvement across these English benchmarks. See figures \ref{fig:multilang-flops} and \ref{fig:eng-flops}.

In figures \ref{fig:multilang-flops} and \ref{fig:eng-flops} we plot the accuracy versus FLOPs used to train various open models across a similar set of common benchmarks. For multilingual benchmarks, Eagle and Finch represent a substantial improvement to the Pareto frontier, achieving far higher scores than other models trained for a similar number of FLOPs. The two models additionally obtain competitive performance across these English benchmarks.

\begin{table}[h]
    \setlength\extrarowheight{-5pt}
    \centering
    \footnotesize
    % \begin{adjustbox}{max width=0.8\linewidth}
    % \resizebox{0.8\textwidth}{!}{
    \begin{tabular}{lrrrrrrrrrr}
\toprule
    \bf Model & \bf lmb.m & \bf lmb.m & \bf pawsx & \bf xcopa & \bf xnli & \bf xsClz & \bf xwin & \bf avg \\
          & ppl $\downarrow$ & acc $\uparrow$ & acc $\uparrow$ & acc $\uparrow$ & acc $\uparrow$ & acc $\uparrow$ & acc $\uparrow$ & acc $\uparrow$ \\
\midrule
    Pythia-1.4b & 115.9 & 35.5  & 50.9  & 52.7  & 38.9  & 51.8  & 68.3  & 49.7  \\
    Mamba-1.4b & 73.1  & 40.4  & 48.0  & 54.4  & \textbf{41.6} & 54.2  & 72.4  & 51.8  \\
    RWKV-4-1.5b & 72.5  & 38.5  & \textbf{53.7} & 55.4  & 39.3  & 56.0  & 67.7  & 51.8  \\
    Eagle-1.5b & 43.2  & 44.8  & 51.9  & 57.9  & 40.4  & \textbf{57.9} & 73.0  & 54.3  \\
    \bf Finch-1.6b & \textbf{37.5} & \textbf{46.9} & 50.9  & \textbf{58.0} & 41.4  & \textbf{57.9} & \textbf{74.9} & \textbf{55.0} \\
\midrule
    Pythia-2.8b & 81.3  & 38.8  & 49.4  & 53.7  & 40.0  & 53.5  & 71.5  & 51.1  \\
    Mamba-2.8b & 53.7  & 43.5  & 43.6  & 55.3  & 42.1  & 56.3  & 75.6  & 52.7  \\
    RWKV-4-3b & 48.1  & 43.4  & 50.9  & 57.5  & 40.9  & 58.1  & 72.3  & 53.9  \\
    Eagle-3b & 30.8  & 49.1  & \textbf{51.6} & 59.0  & 42.3  & 59.8  & 76.9  & 56.5  \\
    \bf Finch-3b & \textbf{28.1} & \textbf{50.5} & 49.7  & \textbf{59.5} & \textbf{44.2} & \textbf{60.7} & \textbf{77.8} & \textbf{57.1} \\
\midrule
    Pythia-6.9b & 85.6  & 36.7  & 48.4  & 54.1  & 40.0  & 54.2  & 70.9  & 50.7  \\
    MPT-7b & 49.8  & 44.4  & 43.5  & 53.6  & 39.8  & 56.3  & 76.9  & 52.4  \\
    Llama-2-7b & 30.4  & 50.8  & 41.2  & 56.7  & 39.9  & 57.5  & 79.5  & 54.3  \\
    Falcon-7b & 28.7  & 51.3  & 48.2  & 56.0  & 39.0  & 56.0  & 77.7  & 54.7  \\
    Mistral-7B-v0.1 & 27.1  & 51.9  & 41.5  & 55.9  & 43.1  & 59.2  & \textbf{81.2} & 55.5  \\
    RWKV-4-7b & 33.1  & 47.4  & \textbf{52.1} & 60.1  & 41.2  & 60.9  & 76.5  & 56.4  \\
    \bf Eagle-7B & \textbf{21.0} & \textbf{53.7} & 45.6  & \textbf{62.2} & \textbf{44.0} & \textbf{63.3} & 80.4  & \textbf{58.2} \\
\bottomrule    
    \end{tabular}
    % }
    % \end{adjustbox}
  \caption{\centering{Multilingual Benchmarks, including LAMBADA Multilingual (\textbf{lmb.m}) \citep{gao10256836framework}, XCOPA \citep{ponti-etal-2020-xcopa}, XNLI \citep{conneau2018xnli}, PAWS-X \citep{yang2019paws}, XStoryCloze (\textbf{xsClz}) \citep{lin2022few}, xWinogrande (\textbf{xwin}) \citep{tikhonov2021s}.}}
  \label{tab:multilang_bench}%
\end{table}%
\begin{table}[h]
    \setlength\extrarowheight{-5pt}
    \setlength\tabcolsep{5pt}
    \footnotesize
    \centering
    % \scriptsize
    % \begin{adjustbox}{max width=0.8\linewidth}
    % \resizebox{0.8\textwidth}{!}{
    \begin{tabular}{lrrrrrrrrrr}
\toprule
    \bf Model & \bf lmb.o & \bf hella & \bf piqa & \bf arcE & \bf arcC & \bf glue & \bf winG & \bf sciq & \bf copa & \bf avg \\
          & acc $\uparrow$ & acc\_n $\uparrow$ & acc $\uparrow$ & acc $\uparrow$ & acc $\uparrow$ & acc $\uparrow$ & acc $\uparrow$ & acc $\uparrow$ & acc $\uparrow$ & acc $\uparrow$ \\
\midrule
    Pythia-1.4b & 61.0  & 52.0  & 70.8  & 61.4 & 26.2 & 47.1  & 57.3  & 86.5  & 71.0  & 59.2  \\
    RWKV-4-1.5b & 60.1  & 51.6  & 71.5  & 58.4 & 27.1  & 46.1  & 55.2  & 84.7  & 78.0  & 59.2 \\
    Eagle-1.5b & 65.7  & 55.0  & 71.1  & 62.2 & 28.7 & \textbf{54.1} & 59.1  & \textbf{89.7} & 76.0  & 62.4 \\
    \textbf{Finch-1.6b} & \textbf{66.8} & 57.3  & 72.6  & 62.7 & 29.8 & 49.8  & 59.4  & \textbf{89.6}  & 78.0  & \textbf{62.9} \\
    \textbf{Mamba-1.4b} & 64.5  & \textbf{59.0} & \textbf{74.2} & \textbf{65.0} & \textbf{30.1} & 47.0  & \textbf{61.3} & 87.1  & \textbf{80.0} & \textbf{63.1} \\
\midrule
    Pythia-2.8b & 63.8  & 59.1  & 73.9  & 63.8 & 29.0 & 47.3  & 58.2  & 88.6  & 79.0  & 62.5  \\
    RWKV-4-3b & 65.7  & 58.8  & 72.4  & 62.9 & 32.4 & 53.6  & 57.5  & 87.6  & \textbf{86.0} & 64.1  \\
    Eagle-3b & 68.7  & 62.6  & 74.3  & 68.6 & 33.8 & 46.3  & 62.0  & \textbf{92.6} & 85.0  & 66.0  \\
    Mamba-2.8b & 68.1  & \textbf{65.9} & \textbf{75.2} & \textbf{69.7} & 33.8 & 46.3  & 63.0  & 90.2  & 84.0  & 66.2  \\
    \textbf{Finch-3b} & \textbf{70.8} & 64.8  & 74.2  & 66.5 & \textbf{34.6} & \textbf{58.2} & \textbf{63.6} & \textbf{92.5} & 82.0  & \textbf{67.5} \\
\midrule

    Pythia-6.9b & 60.9  & 63.2  & 74.8  & 66.5 & 32.0 & 47.7  & 61.5  & 88.9  & 79.0  & 63.8  \\
    RWKV-4-7b & 69.8  & 65.3  & 75.0  & 67.4 & 34.0 & 56.4  & 62.4  & 90.8  & 85.0  & 67.3  \\
    MPT-7b & 68.7  & 76.3  & 79.3  & 74.9 & 39.7 & 48.7  & 68.1  & 93.9  & 88.0  & 70.9  \\
    Llama-2-7b & 73.5  & 76.0  & 78.1  & 76.4 & 43.1 & 42.9  & 69.1  & 93.9  & 87.0  & 71.1  \\
    Falcon-7b & 74.6  & 76.4  & 79.5  & 74.8 & 40.3 & 45.8  & 67.1  & 94.4  & 88.0  & 71.2  \\
    Eagle-7B & 74.2  & 70.9  & 77.0  & 73.8 & 39.5 & \textbf{57.5} & 67.4  & 95.5  & 88.0  & 71.5  \\
    \textbf{Mistral-7B-v0.1} & \textbf{75.5} & \textbf{81.0} & \textbf{80.5} & \textbf{80.8} & \textbf{50.1} & 51.5  & \textbf{73.6} & \textbf{95.9} & \textbf{93.0} & \textbf{75.8} \\
\bottomrule    
    \end{tabular}
    % }
    % \end{adjustbox}
  \caption{\centering{English Focused Benchmarks, including LAMBADA (\textbf{lmb.o}) \citep{paperno2016lambada}, Hellswag (\textbf{hella}) \citep{hampel1974influence}, PIQA \citep{bisk2020piqa}, AI2 ARC (\textbf{arcE}, \textbf{arcC}) \citep{bhakthavatsalam2021think}, GLUE \citep{wang2018glue}, Winogrande (\textbf{winG}) \citep{sakaguchi2021winogrande}, SciQ \citep{welbl2017crowdsourcing}, COPA \citep{roemmele2011choice}.}}
  \label{tab:eng_bench}%

\end{table}%
\newpage

\subsection{Associative Recall}
\label{subsec:associative_recall}

Associative recall (AR) is a synthetic task designed to mimic the way that humans associate and retrieve information. It measures a model's proficiency in recalling information that was previously mentioned in context. Prior research suggests that a model's ability to perform AR is indicative of its effectiveness in in-context learning \citep{elhage2021mathematical, olsson2022incontext}. As a result, AR has been adopted as a benchmark in developing new language model architectural designs. \citep{fu2023hungry, poli2023hyena, lutati2023focus}. \citet{arora2023zoology} benchmarked a range of models for multi-query associative recall (MQAR) and identified a performance gap between various linear transformer architectures and the transformer with attention. In MQAR tasks, prior RWKV models demonstrated a correlation between model dimension and sequence length. To compare architectures, we trained models using RWKV-4, Eagle and Finch on MQAR, using identical criteria with various model dimensions and sequence lengths. Our findings reveal significant improvements in MQAR with Eagle and Finch. Notably, Finch achieves extremely high accuracy in MQAR in our tests, and outperforms all well-known non-transformer architectures previously used to train large language models. Our experiments reveal performance disparities between Mamba~\citep{gu2023mamba} and Finch, despite their shared architectural features such as matrix-valued state and data-dependent memory modification, suggesting different combinations of these elements result in superior performance.
\begin{figure}[ht]
\centering
\includegraphics[width=\textwidth]%{images/RWKV_456_AR.png}
{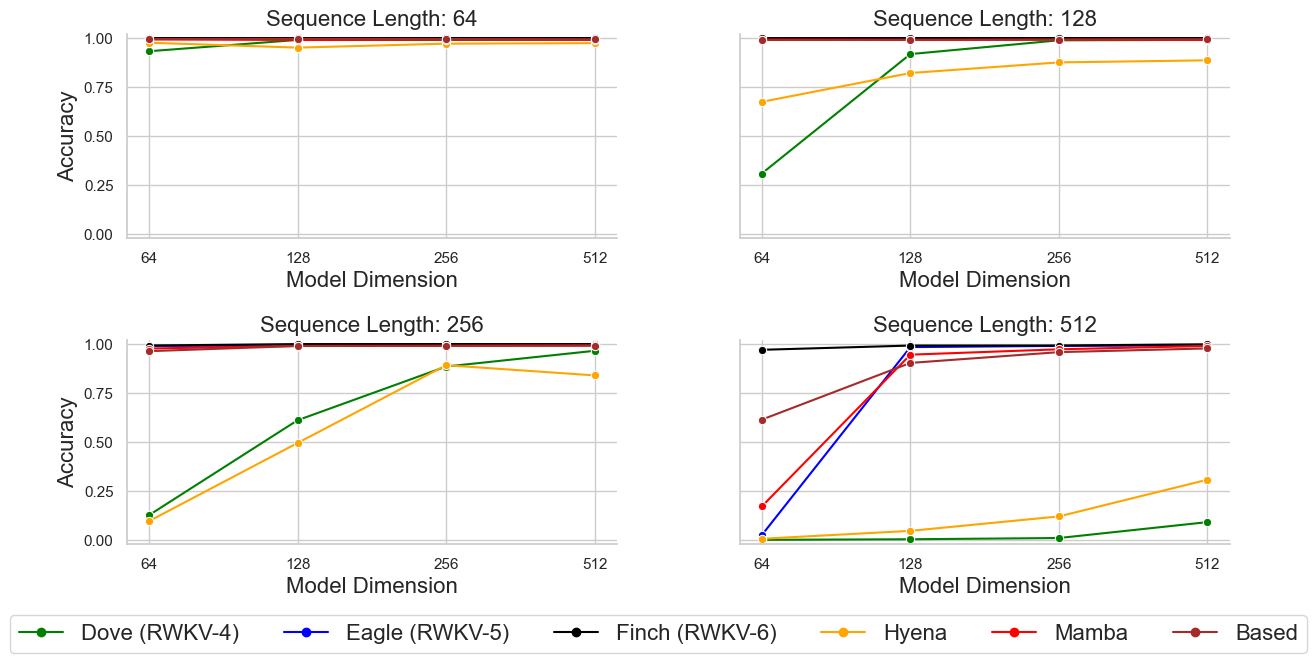}
\caption{MQAR tasks. An increase in sequence length correlates with increased task difficulty.}
\label{fig:rwkv5_ar}
\end{figure}

\subsection{Long Context Experiments}\label{subsec:long_context_experiments}

\begin{figure}[htbp]
\centering
\includegraphics[width=\textwidth]{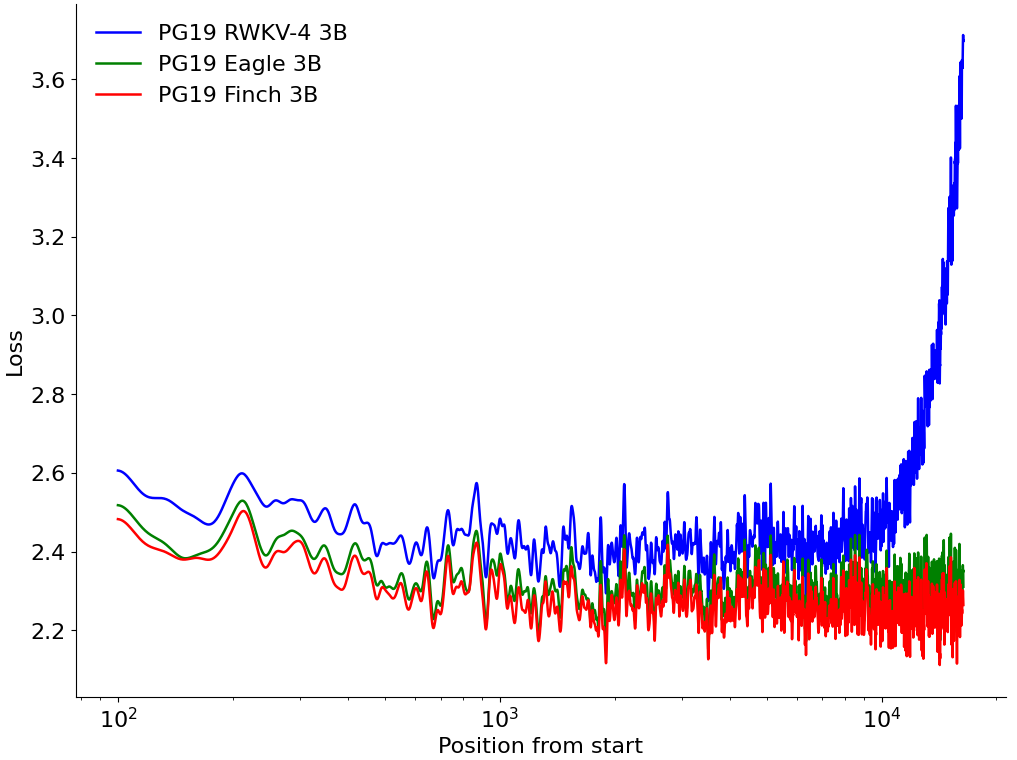}
\caption{\centering{Loss along sequence offset for 3B RWKV-4 World, Eagle and Finch on PG19 dataset. All models were pretrained with context length 4096.}}
\label{fig:rwkv-ctxlen}
\end{figure}

We test loss versus sequence position on the PG19 \citep{raecompressive2019} test set of books from token 2048 onward across RWKV-4, Eagle, and Finch. We find that Eagle improves dramatically over RWKV-4 on this long sequence task, despite having been trained solely on sequence length 4096. Finch further improves on this test beyond Eagle, with loss continuing to drop further into the sequence. See \hyperref[fig:rwkv-ctxlen]{Figure \ref*{fig:rwkv-ctxlen}} for details.
%Also, according to \cite{luo2023prosg}, utilizing synthetic gradients can enhance RWKV's capacity to handle extended sequences.

\subsection{Bamboo Benchmark} \label{subsec:bamboo}
The Bamboo benchmark~\citep{dong2023bamboo} evaluates the overall long-context language modeling capability of LLMs from five aspects: question answering, hallucination detection, text sorting, language modeling, and code completion, comprising a total of ten evaluation tasks. We test models on the 4k version of the benchmark, which includes all ten tasks with a maximum context window length of 4k. We choose not to present results on the code completion task since all tested models failed to generate correct code completions for this task.
In Table~\ref{tab:bamboo_bench}, we present the results of nine tasks, with either accuracy or F1 score, along with their average scores. At both the 1.5b and 3b scales, the latest Finch and Eagle models outperform the vanilla Mamba by at least a 7\% average score, while remaining comparable with the Mamba trained on Hermes data (\textit{i.e.}, only a 0.7\% drop in the average score). Note that, despite being trained on only 1.1T tokens, Eagle-7b consistently outperforms Pythia by an average of 13.5\% at the 7b scale, and it also surpasses LLaMA2-Chat-7b on several tasks in the Bamboo benchmark.
These results demonstrate the superior capacity of the proposed Finch and Eagle models on a vast range of long-context tasks.

\begin{table*}
\centering
\scriptsize
\setlength\tabcolsep{2pt}
\begin{tabular}{lcccccccccc} 
\toprule
Model               & \makecell{meetingqa\\Acc.$\uparrow$} & \makecell{paperqa\\Acc.$\uparrow$} & \makecell{meetingpred\\Acc.$\uparrow$} & \makecell{showspred\\Acc.$\uparrow$} & \makecell{reportsumsort\\Acc.$\uparrow$} & \makecell{showssort\\Acc.$\uparrow$} & \makecell{senhallu\\F1$\uparrow$} & \makecell{abshallu\\F1$\uparrow$} & \makecell{altqa\\Acc.$\uparrow$} & \textbf{Avg.$\uparrow$}  \\
\midrule
Pythia-1.4b         & 15.0\%  & 4.0\%    &  0.0\%     &   0.0\%     &  0.0\%      &  0.0\%     & 0.0\%    &  0.0\%   & 0.0\%  & 2.1\%  \\
Mamba-1.4b         &  15.0\%  & 2.0\%    &  0.0\%   &  0.0\%     &  0.0\%    & 0.0\%     & 0.0\%   & 2.0\%    & 0.0\%   & 2.1\%  \\
% Finch-1.6b         & 17.0\%      & 15.0\%    & 1.0\%         & 3.0\%       & 0.0\%   & 0.0\%       & 16.8\%     & 7.0\%      & 2.0\%   & 6.9\%   \\
Eagle-1.5b         &  21.0\%   & 19.0\%   & 1.0\%    & 0.0\%    &  0.0\%   &  0.0\%  & 13.2\%    &  23.5\%  & 5.5\%   & 9.2\%  \\
Finch-1.6b         & 19.0\%      & 22.0\%    & 1.0\%         & 8.0\%       & 0.0\%   & 0.0\%       & 10.7\%     & 17.3\%      & 2.5\%   & 8.9\%   \\
\midrule
Pythia-2.8b         &  16.0\%    & 4.0\%    &   0.0\%    &   0.0\%     &   0.0\%     &  0.0\%     &  0.0\%   &  0.0\%   & 0.0\%   &  2.2\% \\
% Mamba-2.8b-Slimpj     &   4.0\%   &  6.0\%   & 0.0\%         & 2.0\%       & 0.0\%           & 0.0\%       &  2.0\%    &  0.0\%    & 0.0\%   &  1.6\% \\
Mamba-2.8b     &   11.0\%   &  4.0\%   & 0.0\%         & 3.0\%       & 0.0\%           & 0.0\%       &  0.0\%    &  3.9\%    & 0.0\%   &  2.4\% \\
Mamba-2.8b-Hermes     & 27.0\%      & 25.0\%    & 0.0\%         & 9.0\%       & 0.0\%           & 0.0\%       & 19.7\%     & 26.4\%     & 0.0   & 11.9\%  \\ 
Eagle-3b  & 16.0\% &   14.0\% &    0.0\%  &    4.0\%  &    0.0\%  &    0.0\%    &     25.0\%   &    29.2\%   &     1.0\%   &    9.9\%         \\
Finch-3b        &   20.0\%      & 26.0\%    & 4.0\%         & 7.0\%       & 0.0\%           & 0.0\%       & 14.4\%     & 23.6\%     & 6.5\%   & 11.3\%   \\
\midrule
Pythia-6.9b         &  19.0\%     & 7.0\%    &  0.0\%   &  0.0\%    &   0.0\%   &   0.0\%  & 0.0\%  &  0.0\% & 0.0\% & 3.3\%  \\
Eagle-7b-Hermes     & 31.0\% &  23.0\%  &  0.0\%  &    0.0\%   &   0.0\%  &    0.0\%    &   50.3\%        &    46.9\%      &     0.0\%     &    16.8\%         \\
LLaMA2-Chat-7b      & 6.0\%       & 17.0\%    & 4.0\%         & 12.0\%      & 0.0\%           & 0.0\%       & 64.7\%     & 63.4\%     & 46.0\%  & 24.1\%  \\
Mistral-Instruct-7b & 65.0\%      & 73.0\%    & 17.0\%        & 32.0\%      & 0.0\%           & 0.0\%       & 80.5\%     & 72.8\%     & 13.5\%  & 39.3\%  \\
\bottomrule
\end{tabular}
\caption{\centering{Results on the long context reasoning benchmark: Bamboo. We compare both transformer and linear attention language models on three different scales: 1.5b, 3b, and 7b.}}
\label{tab:bamboo_bench}
\end{table*}

% @ cryscan
% \subsection{Long Context Experiments}

% Place Holder -- The exact number is unknown, needs experiments.

\section{Speed and Memory Benchmarks}
\label{sec:Speed and Memory Benchmarks}

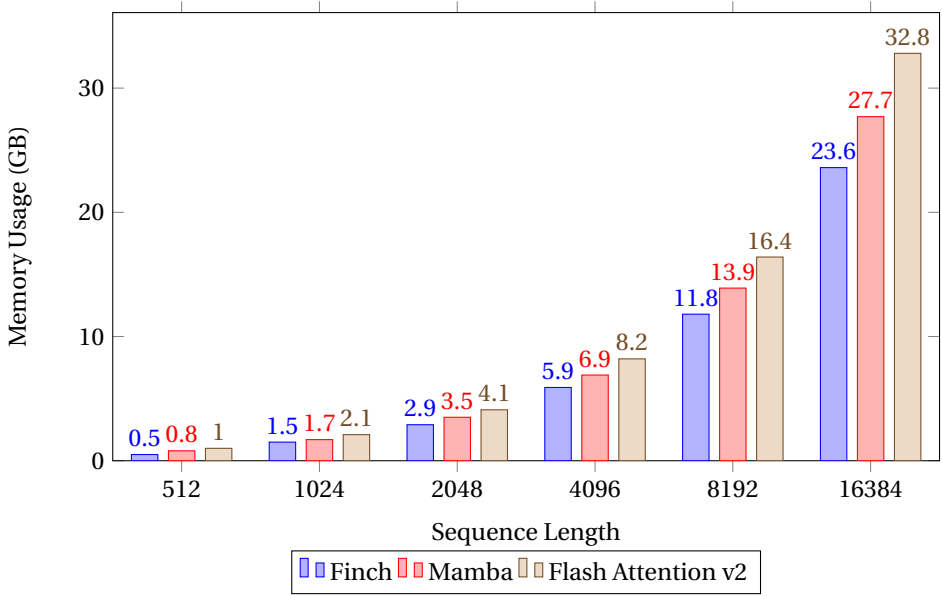
\begin{figure}[h]
    \begin{minipage}{0.9\textwidth}
        \centering
        \begin{tikzpicture}
        \begin{axis}[
            ybar=4pt, % Adjust the width of the bars
            width=1\textwidth,
            height=0.6\textwidth,
            ylabel={Memory Usage (GB)},
            xlabel={Sequence Length},
            ymin=0,
            xtick=data,
            xticklabels={512, 1024, 2048, 4096, 8192, 16384},
            ylabel style={rotate=0},
            nodes near coords,
            nodes near coords align={vertical},
            legend style={at={(0.5,-0.2)},
            bar width=10pt,
            anchor=north,legend columns=-1},
            every axis x label/.style={at={(current axis.south)}, below=20pt},
        ]
        \addplot coordinates {(1, 0.5) (2, 1.5) (3, 2.9) (4, 5.9) (5, 11.8) (6, 23.6)};
        \addlegendentry{Finch}
        
        \addplot coordinates {(1, 0.8) (2, 1.7) (3, 3.5) (4, 6.9) (5, 13.9) (6, 27.7)};
        \addlegendentry{Mamba}
        
        \addplot coordinates {(1, 1.0) (2, 2.1) (3, 4.1) (4, 8.2) (5, 16.4) (6, 32.8)};
        \addlegendentry{Flash Attention v2}
        \end{axis}
        \end{tikzpicture}
        \caption{Memory Usage vs. Sequence Length (A100 80GB)}
        \label{fig:memory_vs_sequence}
    \end{minipage}
\end{figure}

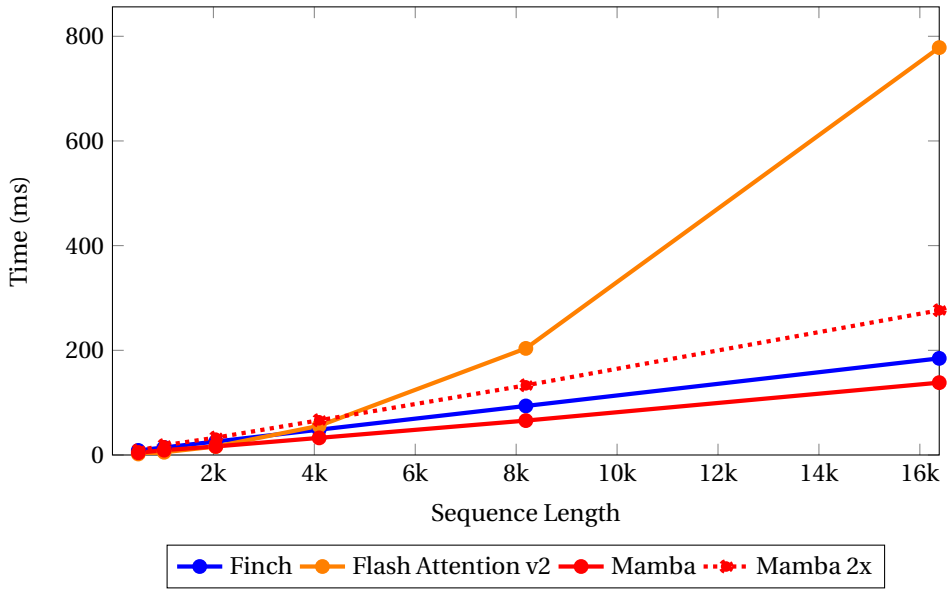
\begin{figure}[h]
    \begin{minipage}{0.9\textwidth}
        \centering
        \begin{tikzpicture}
        \begin{axis}[
            xlabel={Sequence Length},
            ylabel={Time (ms)},
            xmin=0, xmax=16384,
            ymin=0,
            width=1\textwidth,
            height=0.6\textwidth,
            legend style={at={(0.5,-0.2)},anchor=north,legend columns=-1},
            scaled x ticks=false, % Disable scientific notation for x-axis tick labels
            xtick={2000, 4000, 6000, 8000, 10000, 12000, 14000, 16000},
            xticklabels={2k, 4k, 6k, 8k, 10k, 12k, 14k, 16k},
        ]
        \addplot[mark=*,blue,ultra thick] coordinates {
            (512, 8.727)
            (1024, 14.405)
            (2048, 25.72)
            (4096, 48.369)
            (8192, 93.686)
            (16384, 184.6029)
      };
        \addlegendentry{Finch}

        \addplot[mark=*,orange,ultra thick] coordinates {
            (512, 1.924)
            (1024, 5.226)
            (2048, 16.204)
            (4096, 55.712)
            (8192, 203.655)
            (16384, 778.34)
      };
        \addlegendentry{Flash Attention v2}
        
        \addplot[mark=*,red,ultra thick] coordinates {
            (512, 4.253)
            (1024, 9.218)
            (2048, 16.329)
            (4096, 32.707)
            (8192, 65.699)
            (16384, 138.22)
      };
        \addlegendentry{Mamba}
        
        \addplot[mark=*,red,dotted,ultra thick] coordinates {
            (512, 8.586)
            (1024, 18.584)
            (2048, 33.138)
            (4096, 66.325)
            (8192, 133.08)
            (16384, 276.485)
      };
        \addlegendentry{Mamba 2x}
        
        \end{axis}
        \end{tikzpicture}
        \caption{Time vs. Sequence Length (A100 80GB)}
        \label{fig:time_vs_sequence}
    \end{minipage}
\end{figure}
We compare the speed and memory utilization of the Attention-like kernels for Finch, Mamba\footnote{We also plot Mamba 2x which uses 2 runs through the Mamba kernel instead of one. This is done to mimic the usage of twice the number of layers in Mamba vs Finch and Transformers}, and Flash Attention\footnote{We use the PyTorch Implementation of Flash Attention v2}~\citep{dao2023flashattention2} in Figures \ref{fig:memory_vs_sequence} and \ref{fig:time_vs_sequence}. For all benchmarks, we use a batch size of 8, a model dimension of 4096, and a head size of 64 for both Flash Attention and Finch. For Mamba, we employ a state dimension of 16, a model dimension of 8192, to mimic Mamba's usage of an expansion factor of 2. Our findings indicate that Finch's speed in training scales linearly with respect to sequence length, exhibiting similar scaling to Mamba. We find Finch is significantly faster than Flash Attention for sequence lengths beyond 4k, being around 4.2x faster for a sequence length of 16k. Furthermore, Finch consistently outperforms Mamba and Flash Attention in terms of memory usage, using 40\% and 17\% less memory usage than Flash Attention and Mamba respectively. Further optimization of our Finch CUDA implementation, including algorithmic improvements, are possible, and could lead to speed increases and greater parallelization. However, this optimization is left for future work.

\section{Multimodal Experiments}
\label{sec:multimodal_experiments}

In this section, we explore the capabilities of Eagle when extended to handle multimodal tasks, where the model processes and integrates textual inputs with inputs in a different domain.

\subsection{RWKV Music Modelling}
\label{sec:music-modelling}
To investigate the Eagle architecture's applicability to music modeling, we use the Irishman ABC music sheet dataset \citep{wu2023tunesformer} to train a new RWKV-5-Music model using the same hyperparameters as the existing RWKV-4-Music model. The loss of RWKV-5 is approximately 2\% lower than that of the previous generation model, and this improvement is primarily observed in the musical score part, indicating that RWKV-5 possesses stronger modeling and generalization capabilities than its predecessor. The model has a total of $L=24$ layers, with a dimension of $D=512$ and uses a byte-level tokenizer with $V=128$ tokens. The training context length is $1024$ bytes. We use all 2,162 pieces of music in the validation set and calculate the loss for each position from the start. The loss is averaged across all pieces of music, then Gaussian smoothed over the position in the sequence.

\begin{figure}[h]
    \centering
    \includegraphics[width=0.75\textwidth]{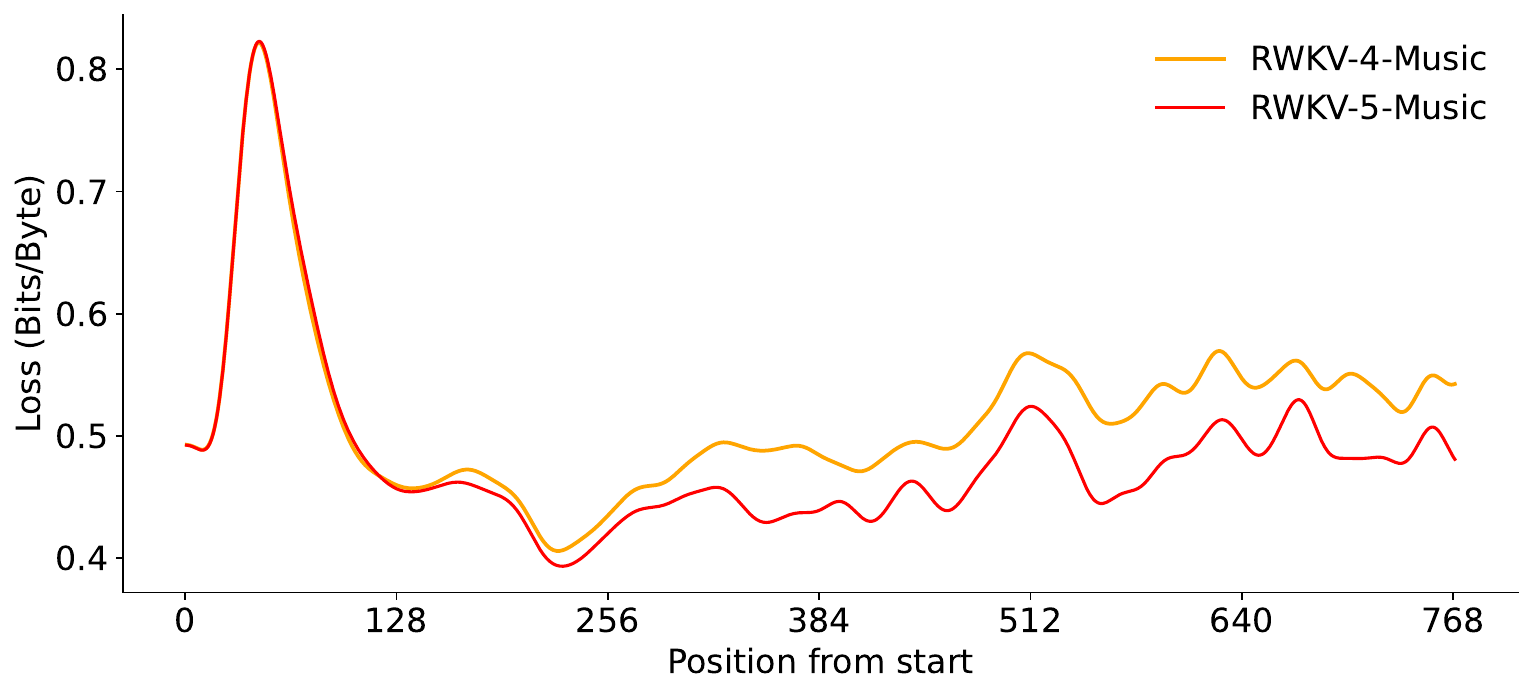}
    \caption{Music modelling loss over sequence position.}
    \label{fig:music}
\end{figure}

The figure \ref{fig:music} shows the loss as a function of position. Note that the first 30-100 bytes of the ABC format are the file header and control codes, followed by the musical scores. The loss of RWKV-5 is approximately 2\% lower than the previous generation model, and it is shown mainly in the musical score part, indicating that RWKV-5 has stronger modelling and generalization capabilities than its precedent model.

\subsection{VisualRWKV}
\label{sec:visualrwkv}
% @Howard Hou: Add VisualRWKV experiments here

\input{sections/VisualRWKV}

% We also plan to explore new methods that combine parallelized state generation with a fine-tuned RNN structure. This approach will process each token by considering the outputs from all previous layers, which could lead to improved processing efficiency and a better understanding of sequential relationships in data.
\section{RWKV on Audio}
\label{sec:audiorwkv}
\input{sections/AudioRWKV}

\section{Conclusions} \label{conclusions}
In this work, we introduced Eagle (RWKV-5) and Finch (RWKV-6), marking substantial progress in RNN-based language models by integrating multiheaded matrix-valued states and dynamic data-driven recurrence mechanisms. These models demonstrate exceptional performance on MQAR and diverse linguistic benchmarks, challenging the dominance of traditional Transformer architectures while retaining key RNN advantages. With models publicly available under the Apache 2.0 license and trained on an extensive multilingual corpus, our work not only advances the capabilities of language models but also emphasizes community accessibility and applicability across various domains. While acknowledging the computational and ethical challenges ahead, we hope that Eagle and Finch's efficient new architecture and wide availability will help push the boundaries of language modeling and pave the way for future innovations.

\paragraph{Limitations}
The Eagle and Finch models fall short on certain aspects that can be mitigated and addressed in future work.

% TODO: explain more: associative recall, etc. 

% Our testing shows that Eagle/Finch models are extremely incapable of answering certain multiple-choice questions.

We experimented with using Eagle as an embedding model on the Massive Text Embedding Benchmark (MTEB) \citep{muennighoff2023mteb} but were not able to get strong embedding performance. We believe that its state is a very high-quality embedding of the context but an appropriate method is required to aggregate the information content. We leave this to future work.

Because our training corpus contains some synthetic data from GPT-3.5 and ChatGPT, our released models exhibit behaviors similar to ChatGPT and will mimic ChatGPT's conversation style and tone. For instance, the model might occasionally claim that it is trained by OpenAI. However, this is not a general property the RWKV architecture but rather a specific outcome of the data and training process.

\paragraph{Future Work} \label{future_work}

Our 1.12 trillion token multilingual training corpus is much smaller than the training data sizes for contemporary models such as LLaMA2 \citep{touvron2023llama}, and expanding our training corpus to be more diverse and expansive is a key priority to improving model performance~\citep{albalak2024survey}. We also plan to train and release larger versions of Finch such as 7B and 14B parameters, and further extend its performance with reduced inference and training costs via Mixture of Experts~\citep{shazeer2017outrageously}.
%\subsubsection*{Author Contributions}
%If you'd like to, you may include  a section for author contributions as is done in many journals. This is optional and at the discretion of the authors.

% Add for camera-ready. Currently violates double-blind

\subsubsection*{Acknowledgments}
We thank Stability AI for the compute used to train our models and for technical support in the development of RWKV. We also thank the members of the RWKV and EleutherAI Discord servers for their help and work on further extending the applicability of RWKV to different domains. We also thank Shenzhen Yuanshi Intelligence Co., Ltd. for its contribution to the promotion and commercialization of RWKV. We thank Songlin Yang for assistance with the code and ideas for our time-parallel implementations.

%\section{Ethics Statement}
%Authors can add an optional ethics statement to the paper. For papers that touch on ethical issues, this section will be evaluated as part of the review process. The ethics statement should come at the end of the paper. It does not count toward the page limit, but should not be more than 1 page. 

\newpage

\bibliography{main}
\bibliographystyle{colm2024_conference}

\newpage
\appendix

\section{Author Contributions}

    \paragraph{Bo Peng} Original RWKV-5 and RWKV-6 ideas, original code, performance optimizations, original experiments, tokenizer design, dataset composition, and trained models from 0.4B to 7B.

\paragraph{Daniel Goldstein} RWKV-5 and RWKV-6 time-parallelization research and code. Manuscript organization, initial draft sections \ref{sec:background}, \ref{sec:rwkv_model}, \ref{sec:method}, \ref{sec:tokenizer}, \ref{sec:dataset}, \ref{subsec:evals}, \ref{subsec:long_context_experiments}, and appendices \ref{app_architecture}, \ref{sec:training_dataset_details}, \ref{multilang_chat_examples}, and \ref{chat_examples}. Proofreading and revisions of full manuscript. Experiments for \ref{subsec:evals} and \ref{subsec:long_context_experiments}. Additional work on tables \ref{tab:costs}, \ref{tab:compare}, figure \ref{sec:Speed and Memory Benchmarks}, and appendix \ref{sec:hyperparameters}.

\paragraph{Quentin Anthony} Led manuscript and results organization. Revisions and proofreading of manuscript. 

\paragraph{Alon Albalak} Manuscript organization, initial draft of section~\ref{sec:introduction}, proofreading, formatting, and revisions of full manuscript.

\paragraph{Eric Alcaide} Figure 1. Proofreading, formatting, and revisions of full manuscript. 

\paragraph{Stella Biderman} Oversight and planning on scaling figures and FLOP results. Manuscript assistance.

\paragraph{Eugene Cheah} Experiments for section \ref{subsec:evals}.
\paragraph{Xingjian Du} Investigated using the RWKV models for multimodal applications. Optimizing draft Sections \ref{subsec:evals}  \ref{subsec:bamboo}  \ref{sec:Speed and Memory Benchmarks}. Proofreading and revisions.
\paragraph{Teddy Ferdinan}
%teddy.ferdinan@pwr.edu.pl
Self-Learning Capability (SLC) evaluation (Sec.~\ref{sec:self-learning}) -- implementation of the method, performing experiments, initial draft of the section, description of the results (Tab.~\ref{table_self_learning_results}).

\paragraph{Przemysław Kazienko}
%przemyslaw.kazienko@pwr.edu.pl
Planning the experiment with Self-Learning Capability (SLC) evaluation (Sec.~\ref{sec:self-learning}), supervising SLC experiments.

\paragraph{Jan Kocoń}
%jan.kocon@pwr.edu.pl
Co-author of the idea of Self-Learning Large Language Models \citep{ferdinan2024unknown} -- supervising evaluation of RWKV Self-Learning Capability (Sec.~\ref{sec:self-learning}), supervising experiments with zero-shot evaluation on additional NLP tasks (Sec.~\ref{sec:zero-shot-eval}), proofreading of full manuscript.

\paragraph{Kranthi Kiran GV} Manuscript (sections \ref{subsec:evals} and \ref{sec:multimodal_experiments};  revision and proofreading). Tables \ref{tab:multilang_bench} and \ref{tab:eng_bench}.

\paragraph{Haowen Hou}
%houhaowen@gml.ac.cn
VisualRWKV based on RWKV-5, which encompasses original code, original experiments for Table \ref{tab:visualrwkv_results}, and trained models ranging from 1.5 billion to 3 billion parameters. Figure \ref{fig:visualrwkv} and draft section \ref{sec:visualrwkv}. Proofreading and formatting fixes.

\paragraph{Jiaju Lin} Contribute to the training and evaluation of AudioRWKV, including original codes and experiments. \ref{sec:audiorwkv}.

\paragraph{Satyapriya Krishna} Primarily contributed to the evaluations of the models (Section \ref{evaluations} and \ref{sec:alignbench}), and also made edits/improvements throughout the document. 

\paragraph{Ronald McClelland Jr.} Tables \ref{tab:costs} and \ref{tab:compare}. Dataset research. Proofreading and formatting fixes.

\paragraph{Niklas Muennighoff} Investigated using the RWKV models for embedding.

\paragraph{Fares Obeid} RWKV-5 and RWKV-6 time-parallelization research. Section \ref{sec:Speed and Memory Benchmarks}. Experiments for figures \ref{fig:memory_vs_sequence} and \ref{fig:time_vs_sequence}. Proofreading full manuscript.

\paragraph{Atsushi Saito} Section \ref{sec:introduction},  \ref{sec:tokenizer}, \ref{subsec:evals} and \ref{subsec:associative_recall}. Experiments for \ref{subsec:associative_recall}. Proofreading and adding citations.

\paragraph{Guangyu Song} 
Section \ref{subsec:associative_recall}. Initial draft sections \ref{sec:introduction}, \ref{future_work}. Experiments for \ref{subsec:associative_recall}. Contributions to table \ref{tab:costs}. Proofreading and citations.

\paragraph{Haoqin Tu}
Section~\ref{subsec:bamboo}, experiments for Table~\ref{tab:bamboo_bench}. Proofreading full manuscript.

\paragraph{Stanisław Woźniak}
Experiments with zero-shot evaluation on additional NLP tasks (Sec.~\ref{sec:zero-shot-eval}).

\paragraph{Bartłomiej Koptyra}
Zero-shot evaluation comparison of Raven and Eagle 7B on additional NLP tasks (Sec.~\ref{sec:zero-shot-eval}).

\paragraph{Aleksander Szczęsny}
Conducted experiments on given datasets tasks: TextEntail, GoEmo, PolEmo, WNLI (Sec.~\ref{sec:zero-shot-eval}).

\paragraph{Cahya Wirawan}
Developed optimized implementation of RWKV World tokenizer for \ref{tab:speed_tokenizers}.

\begin{figure}
    \centering
    \includegraphics[width=0.9\textwidth,height=0.9\textheight,keepaspectratio]{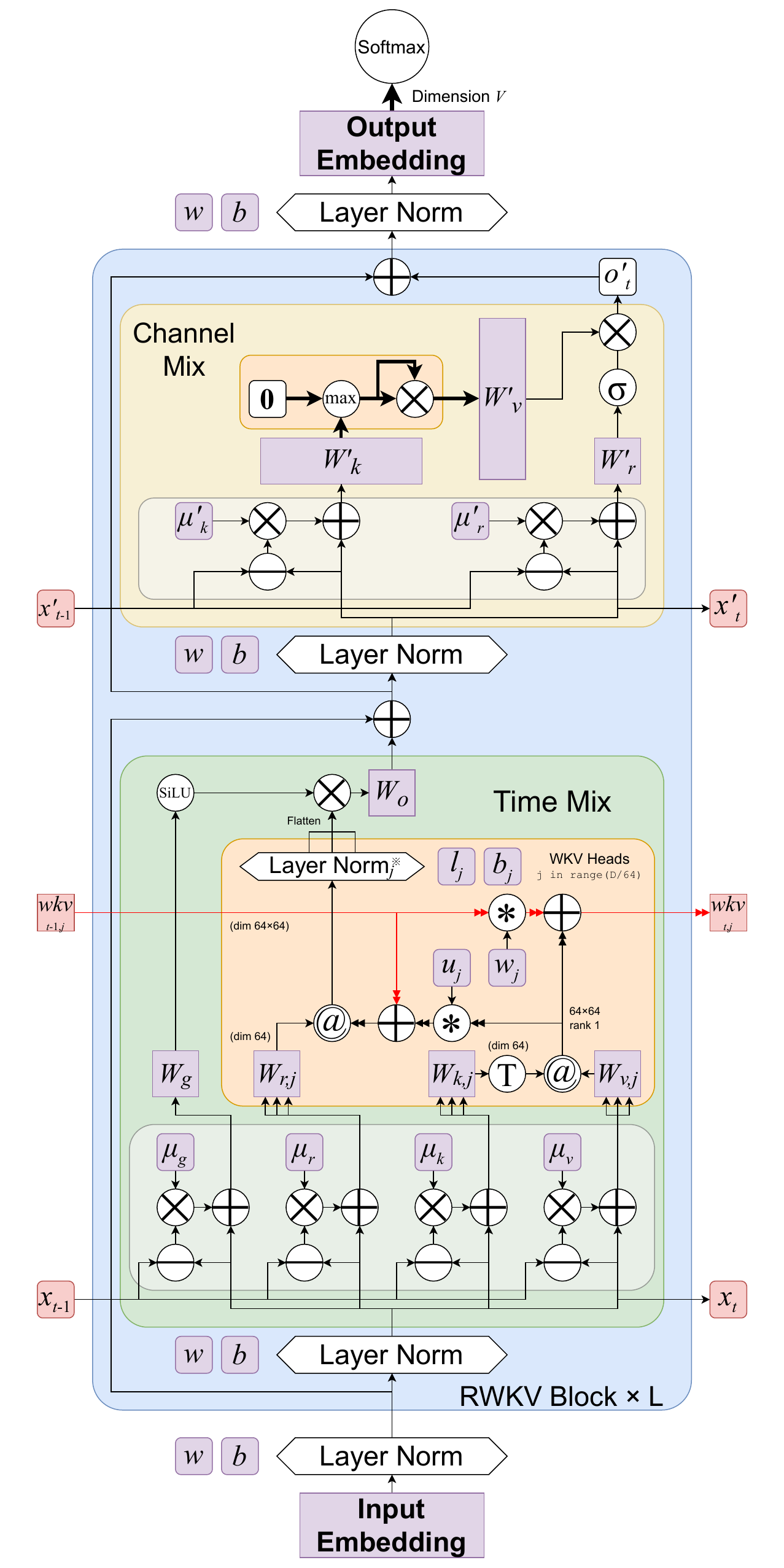}
    \caption{Eagle Overall Architecture.}\label{fig:rwkv52}
\end{figure}

\paragraph{Ruichong Zhang}
Initial paper structure organization, draft sections \ref{sec:rwkv_model}, \ref{sec:method}, \ref{sec:tokenizer}, \ref{sec:pretrained_models} and appendices \ref{flop_count}, \ref{app_tokenizer}, \ref{sec:hyperparameters} and \ref{sec:initialization}. Experiments for music of section \ref{sec:music-modelling} and alignment of section \ref{sec:alignbench}. Figure \ref{fig:music} and \ref{fig:rwkv52}. Additional work on section \ref{conclusions} and appendix \ref{app_architecture}. Proofreading and revision.

\paragraph{Bingchen Zhao}
Section~\ref{sec:mtbench}, experiments for Figure~\ref{fig:mtbench}. Proofreading full manuscript.

\paragraph{Qihang Zhao} Section \ref{sec:background}, Tables \ref{tab:costs}. Proofreading and revisions. 

\paragraph{Peng Zhou} Section \ref{sec:background}, Tables \ref{tab:costs}, appendices \ref{app_related_work},\ref{multilang_chat_examples}. Proofreading and revisions. 

\paragraph{Jian Zhu} Initial draft sections \ref{sec:background} and \ref{app_related_work}. Captions of Table~\ref{tab:eng_bench},~\ref{tab:multilang_bench} and~\ref{tab:dataset}. Fixing citations and formatting the whole manuscript. Proofreading and revisions. 

\paragraph{Rui-Jie Zhu} Optimizing draft Section \ref{app_related_work}, reorganizing Table~\ref{tab:dataset},~\ref{table_self_learning_results}, and~\ref{fig:align_bench}. Proofreading and revisions.

\section{Additional Architecture Details} \label{app_architecture}

% Memory usage: Needs details. 
% How to parallelize across Time dimension? Needs a formula in the Appendix. https://en.wikipedia.org/wiki/Prefix_sum#Parallel_algorithms 
% The MHA style of parallelization is like the one used in v5.1 and can be chunked somewhat to reduce VRAM usage to high rather than absurd levels, not sure if this needs to be made explicit since in reality we use the Mamba style one as mentioned below
The $\bm{WKV}$ computations of Eagle and Finch can be parallelized across the time dimension using a variety of techniques including associative scan or the parallelization techniques used in FlashAttention. \citep{dao2022flashattention} The simplest of these, while highly parallel, prove inefficient due to repeated expensive memory transfers between fast SRAM and slower HBM. We take a different approach when training, choosing to parallelize over non-time dimensions only while using a custom CUDA implementation that carefully keeps state operations in fast SRAM, which is simpler yet provides enough breadth for a highly efficient implementation. See Section \ref{sec:Speed and Memory Benchmarks} for kernel experiments. We provide an additional pure PyTorch implementation with similar full-model speed characteristics that parallelizes over the time dimension using an algorithmic approach similar to GLA  \citep{yang2023gated}.
% We provide more details on this implementation in the appendix 
% This implementation allows for the $\bm{WKV}$ computation to be split into chunks and skip state computations within each chunk. This parallelization doesn't require CUDA or manual backpropagation, allowing efficient training on any hardware and framework using autograd. Typically we found a chunk size of 32 or 24 is best, balancing numerical precision and performance.
% Formulae needed

Unlike Transformers, RWKV's recurrence mechanism does not examine tokens more than one time-step old. This allows us to train on and provide inference for unbounded sequence lengths without requiring increased computing power or memory. Another significant advantage is that RWKV does not utilize explicit positional encoding, which allows RWKV to handle contexts of arbitrary length without modification.

% We theorize that the decaying state mechanism provides enough temporal information for the model to use, much like an implicitly learned version of Attention with Linear Biases (ALiBi).

\paragraph{Finch Token Shift}

Finch changes the token shift mechanism to become data-dependent. Intuitively, important information can effectively flag itself for inclusion using this mechanism, and less important information can flag itself to partially or fully avoid entering the data stream, leaving room for more important pre-existing data to remain. Viewed from the perspective of induction heads, we theorize that this could allow for potential misleading matches to be pre-filtered out up front if they are not deemed useful for a given task.

\paragraph{Improved WKV (Weighted Key-Value State) Modules}

The Eagle WKV attention sub-module is similar to the linear attention mechanism found in RetNet, but with learned per-channel decay rates replacing RetNet's static per-head decay rates. Our matrix-valued states feature a geometrically decaying $K^\mathrm{T} V\ \in \mathbb{R}^{(D/h) \times (D/h)}$ term. This term can be intuitively understood as a memory bank of values, with $K$ acting as an input gate for rows receiving the current token embedding's value. Each row of this state decays at its own rate via the learned parameter $w$.

In Finch, we augment the learned token-shift parameters $\mu_r, \mu_k, \mu_v, \mu_w$ and decay rate parameter $w$ with learned weight matrices. Inspired by Low-Rank Adaptation (LoRA)~\citep{hu2022lora}, we provide two new learned weight matrices for each such parameter $y$, computing $y' = y + \tanh(x \bm{A}) \bm{B}$. This approach allows us to dynamically generate data-dependent token-shift amounts and decay rates with only modest increases in computational cost and model size. 

\paragraph{Extra SiLU Gating}
%Cite SiLU. SiLU(x) = \frac{x\exp(x)}{\exp(x)+1}
We remove the Sigmoid activation of receptance in favor of a new SiLU gate on the output of our linear attention calculation. Our receptance term now functions much like the query term in linear attention.

\paragraph{Eagle and Finch Linear Attention Formula, PyTorch Recurrent Implementation}.
\label{code:attention}
\begin{lstlisting}[language=Python]
# r, k, v parameter shape (B,H,1,D//H)
# w parameter of shape (1,H,1,D//H) for Eagle (RWKV-5),
#                      (B,H,1,D//H) for Finch (RWKV-6)
# u parameter of shape (1,H,1,D//H)
# wkv_state parameter of shape (B,H,D//H,D//H)
def rwkv_5_or_6_recurrent(r, k, v, w, u, wkv_state):
    kv = k.mT @ v
    out = r @ (wkv_state + u.mT * kv)
    wkv_state = w.mT * wkv_state + kv
    return out, wkv_state
\end{lstlisting}
% \begin{Verbatim}[fontfamily=tt]
% \end{Verbatim}

% \begin{markdown}
% ```python
% # r, k, v parameter shape (B,H,1,D//H)
% # w parameter of shape (1,H,1,D//H) for Eagle (RWKV-5),
% #                      (B,H,1,D//H) for Finch (RWKV-6)
% # u parameter of shape (1,H,1,D//H)
% # wkv_state parameter of shape (B,H,D//H,D//H)
% def rwkv_5_or_6_recurrent(r, k, v, w, u, wkv_state):
%     kv = k.mT @ v
%     out = r @ (wkv_state + u.mT * kv)
%     wkv_state = w.mT * wkv_state + kv
%     return out, wkv_state
% ```
% \end{markdown}

% Maybe this style better?
%\begin{lstlisting}[language=Python]
%def rwkv_5_or_6_recurrent(r_in, k_in, v_in, w_in, u, wkv_state):
%    out = []
%    T = r_in.size(-2)
%    for t in range(T):
%        r = r_in[...,t:t+1,:]
%        k = k_in[...,t:t+1,:]
%        v = v_in[...,t:t+1,:]
%        w = w_in[...,t:t+1,:]
%        kv = k.mT @ v
%        out.append( r @ (wkv_state + u.mT * kv) )
%        wkv_state = (w.mT * wkv_state) + kv
%    out = torch.cat(out, dim=-2)
%    return out, wkv_state
%\end{lstlisting}

\paragraph{Evolution of RWKV Formula in Expanded form} \label{formula_evolution}
Table~\ref{tab:evo_formula} shows the expansion of terms at each sequence position to illustrate the progression of changes from RWKV-4 through RWKV-6. The main change from RWKV-4 to RWKV-5 is the elimination of denominator and incorporation of matrix states. RWKV-6 introduces the sequential dependence of $w$ which becomes $w_t$.

\begin{table}[h]
\begin{center}
\begin{tabular}{ | l | l |}
 \hline
 $t$ & RWKV-4 $u,w,k_t,v_t \in \mathbb{R}^{D}$, head size $1$ \\ 
 \hline
 0 & $\sigma(r_0) \odot \left(\frac{u \odot k_0 \odot v_0}{u \odot k_0}\right)$ \\  
 %\hline
 1 & $\sigma(r_1) \odot \left(\frac{u \odot k_1 \odot v_1 + k_0 \odot v_0}{u \odot k_1 + k_0}\right)$ \\
 %\hline
 2 & $\sigma(r_2) \odot \left(\frac{u \odot k_2 \odot v_2 + k_1 \odot v_1 + w \odot k_0 \odot v_0}{u \odot k_2 + k_1 + w \odot k_0}\right)$ \\
 %\hline
 3 & $\sigma(r_3) \odot \left(\frac{u \odot k_3 \odot v_3 + k_2 \odot v_2 + w \odot k_1 \odot v_1+w^2 \odot k_0 \odot v_0}{u \odot k_3 + k_2 + w \odot k_1+w^2 \odot k_0}\right)$ \\
 \hline
 \hline
 % $t$ & Eagle (RWKV-5) $u,w,k_t,v_t \in \mathbb{R}^{D/h}$, head count $h$ \\ 
 $t$ & Eagle (RWKV-5) diag($u$), diag($w$), $k_t, v_t \in \mathbb{R}^{64 \times 64}$ for each head, head size 64
 \\
 \hline
 0 & $r_0 \cdot \left(\mathrm{diag}(u) \cdot k_0^\mathrm{T} \cdot v_0\right)$ \\  
 %\hline
 1 & $r_1 \cdot \left(\mathrm{diag}(u) \cdot k_1^\mathrm{T} \cdot v_1 + k_0^\mathrm{T} \cdot v_0\right)$ \\
 %\hline
 2 & $r_2 \cdot \left(\mathrm{diag}(u) \cdot k_2^\mathrm{T} \cdot v_2 + k_1^\mathrm{T} \cdot v_1 + \mathrm{diag}(w) \cdot k_0^\mathrm{T} \cdot v_0\right)$ \\
 %\hline
 3 & $r_3 \cdot \left(\mathrm{diag}(u) \cdot k_3^\mathrm{T} \cdot v_3 + k_2^\mathrm{T} \cdot v_2 + \mathrm{diag}(w) \cdot k_1^\mathrm{T} \cdot v_1 + \mathrm{diag}(w^2) \cdot k_0^\mathrm{T} \cdot v_0\right)$ \\
 \hline
 \hline
 % $t$ & Finch (RWKV-6) $u,w_t,k_t,v_t \in \mathbb{R}^{D/h}$, head count $h$ \\ 
 $t$ & Finch (RWKV-6) diag($u$), diag($w_t$), $k_t, v_t \in \mathbb{R}^{64 \times 64}$ for each head, head size 64
 \\ 
 \hline
 0 & $r_0 \cdot \left(\mathrm{diag}(u) \cdot k_0^\mathrm{T} \cdot v_0\right)$ \\  
 %\hline
 1 & $r_1 \cdot \left(\mathrm{diag}(u) \cdot k_1^\mathrm{T} \cdot v_1 + k_0^\mathrm{T} \cdot v_0\right)$ \\
 %\hline
 2 & $r_2 \cdot \left(\mathrm{diag}(u) \cdot k_2^\mathrm{T} \cdot v_2 + k_1^\mathrm{T} \cdot v_1 + \mathrm{diag}(w_1) \cdot k_0^\mathrm{T} \cdot v_0\right)$ \\
 %\hline
 3 & $r_3 \cdot \left(\mathrm{diag}(u) \cdot k_3^\mathrm{T} \cdot v_3 + k_2^\mathrm{T} \cdot v_2+ \mathrm{diag}(w_2) \cdot k_1^\mathrm{T} \cdot v_1 + \mathrm{diag}(w_2 \odot w_1) \cdot k_0^\mathrm{T} \cdot v_0\right)$ \\
 \hline
\end{tabular}
\end{center}
\caption{Evolution of the RWKV Formula}
\label{tab:evo_formula}
\end{table}

\section{Additional Related Work} \label{app_related_work}

\paragraph{Efficient transformers}
Recently there have been many attempts to improve upon the original transformer time complexity and memory usage, while maintaining or improving performance.
Many of these efficient transformer variants use some form of nonuniform or local attention mechanisms or a combination thereof. For example, LongFormer \citep{beltagy2020longformer} makes use of the sliding window attention and BigBird \citep{zaheer2020bigbird} adopts randomized sparse and random attention patterns to approximate full attention. Similar examples also include LongT5 \citep{guo-etal-2022-longt5} and StreamingLLM \citep{xiao2023efficient}. Instead of using fixed patterns, Reformer \citep{kitaev2019reformer} and Sparse Sinkhorn attention \citep{tay2020sparse} learn to dynamically pay attention to selected tokens.%Fast Linear Attention with Gated Attention Unit \citep{hua2022transformer} creates a combined Feed Forward Network and Attention layer consisting of local attention per chunk and linear attention across chunks. 
Variants including Linformer \citep{wang2020linformer}, Nystr{\"o}mformer \citep{xiong2021nystromformer} and Performer \citep{choromanski2020rethinking} apply matrix approximation methods to approximate the full attention matrix but with lower computational complexity.

%Linear Attention~\citep{katharopoulos2020lineartransformrers} replaces the numerator of MHA's $\mathrm{softmax}(QK^T)V$ with $\phi(Q)\phi(K)^\mathrm{T}V$, allowing a reordering of operations via associativity to $\phi(Q)(\phi(K)^\mathrm{T}V)$, where $\phi$ represents a non-negative feature-map function.  This change makes it possible to compute linear attention as an RNN in $O(1)$ time per step by additively updating a recurrent hidden state with $\phi(K_i^T)V_i$ at each time step $i$. Linear attention can also be trained in parallel, much like MHA.

The Attention Free Transformer (AFT) \citep{zhai2021attention} introduces a modified form of linear attention \citep{katharopoulos2020lineartransformrers} , where the number of attention heads is equal to the size of the feature dimension. It also incorporates a set of learned pairwise positional biases, denoted as $w$. The AFT can be conceptualized as calculating a per-channel weighted average of values. The weight for a specific location is determined by the sum of the key at that location and the corresponding learned positional bias.

%RWKV-4 reformulates the AFT equation by replacing the pair-wise positional biases with a channel-wise vector of additive weight decay rates $w$. It also adds a bonus term $u$ to offset the weight of only the current input specially.

%\begin{align}
%    \mathrm{wkv}_t &= \frac{ \sum_{i=1}^{t-1} \exp(-(t-1-i)w+k_i) \odot v_i + \exp(u+k_t) \odot v_t}{\sum_{i=1}^{t-1} \exp(-(t-1-i)w+k_i) + \exp(u+k_t)}.
%\end{align}

%RWKV-4 also adds several important changes to the overall Transformer architecture outside of attention: token-shift and gating in both attention and feed-forward sub-blocks, and small embedding initialization and normalization to quickly arrive at well-distributed token embeddings. 

Token-shift, as first seen in RWKV-4, is a learned per-channel linear interpolation between the current input and the input at the previous time step, intended to enhance the model with a computationally inexpensive mechanism for choosing between new versus older information within various embedding sub-spaces and for forming induction heads even within a single layer. It is instructive to compare token-shift to a 1D convolution with kernel length 2, as it operates in a similar manner but reuses its parameters via an enforced linear relationship. Recent SSMs have begun using short convolutions in a similar placement within their architectures, typically with kernel length 3 to 4. \citep{poli2023hyena,gu2023mamba}

%The combination of these architectural changes with the new learned decay schedule led RWKV-4 to become the first RNN to rival the performance of transformers, while maintaining $O(N)$ time complexity during training and a corresponding $O(1)$ time complexity per token for inference. 

% we may be able to remove retnet entirely, but it does show that decay works with linear transformers, which is what rwkv-5 does too
Retentive Networks (RetNet) \citep{sun2023retentive} introduces a fixed decay rate schedule and xPos \citep{sun2022lengthextrapolatable} to linear attention. This design combines positional information with an inductive bias towards recency while still allowing both RNN and parallel implementations.

Please refer to \cite{tay2022efficient} and \cite{wan2023efficient} for a comprehensive and in-depth survey of efficient transformers.

% - HGRN~\cite{qin2023hierarchically}: Hierarchically Gated Recurrent Neural Network for Sequence Modeling
\paragraph{Recurrent architectures}
Before the advent of transformers, recurrent neural networks, especially Long Short-Term Memory (LSTM)~\citep{hochreiter1997long} and Gated Recurrent Unit (GRU)~\citep{cho2014learning}, were the dominant architectures in NLP for sequence processing. However, traditional RNNs are hard, if not impossible, to parallelize across the time dimension, susceptible to gradient vanishing and explosion, and ineffective in capturing long-range dependencies, which are ubiquitous in natural language. These shortcomings contributed to the rapid decline of traditional RNNs in NLP.

There has been a revival of RNNs in NLP research \citep{tiezzi2024resurgence} in recent years. Compared to transformers with quadratic complexity, RNNs are highly efficient in autoregressive inference with O(1) time complexity per step, making them an attractive architecture for large language models. Many efforts have been devoted to parallelized recurrent models and improving their capability to capture long-range dependency, while maintaining the low inference complexity.

The Legendre Memory Unit (LMU)~\citep{voelker2019legendre} was designed to efficiently handle long-range dependencies with a new type of memory cell for recurrent neural networks. Unlike LSTM units, which struggle with remembering information over very long sequences, LMU use Legendre polynomials to create a memory system that can maintain and process information over extended time periods more effectively. High-order polynomial projection operators (HiPPO)~\citep{gu2020hippo} generalizes LMU by providing a flexible framework for online compression of signals through polynomial projections, accommodating various polynomial bases beyond Legendre polynomials. It optimizes function approximation over time, adapting to different data timescales without needing predefined hyperparameters. 
%State Space Models (SSMs) employ a hidden state of basis function weights to model an approximation of the input function \cite{gu2020hippo}, updating that hidden state via a differential equation. Earlier SSMs \citep{gu2021s4} were computed using long convolutions in $O(N\log N)$ time per sequence, but could also be formulated as recurrence relations and evaluated sequentially. 
SSMs have inspired a range of follow-up research to incorporate SSMs, or modified SSMs into end-to-end architectures for language modeling, including MEGA \citep{ma2022mega}, DSS \citep{gupta2022diagonal}, H3 \citep{fu2022hungry}, and Linear Recurrent Unit (LRU) \citep{10.5555/3618408.3619518}.

%Recently, it has been shown that SSMs in their recurrent forms can be parallelized across the time dimension via techniques including parallel associative scan \citep{smith2023s5}. 
Mamba \citep{gu2023mamba} is a selective SSM that introduces time-dependent selective mechanism to enhance the long-range modeling ability of SSMs. The selectivity removes the linear time-variance property of the SSM, making it no longer possible to parallelize Mamba as a long convolution kernel. Yet Mamba can still be effectively parallelized using parallel associative scan \citep{BlellochTR90,martin2018parallelizing,smith2023s5} with a hardware-aware implementation. Recently proposed GateLoop \citep{katsch2023gateloop} also adopts a similar data-dependent state transitions. The data-dependent states, also concurrently proposed in GLA \citep{yang2023gated}, are similar to the Weighted Key-Value State in Finch. 

A contemporary but independent work also proposes recurrent models named as Hawk and Griffin \citep{de2024griffin}. Hawk is a recurrent model with the Real-Gated Linear Recurrent Unit (RG-LRU), whereas Griffin mixes the RG-LRU with local multi-query attention, thereby achieving long-context extrapolation efficiently. 

Please see \cite{tiezzi2024resurgence} and \cite{cirone2024theoretical} for a comprehensive review of recent developments of recurrent models.

\section{Training Dataset Details}
\label{sec:training_dataset_details}

Most of the component data sources for the RWKV World v2 dataset are used intact, with no up- or down-sampling done so all tokens are given equal weighting.  Recent works have demonstrated the impact that automated data mixing can have on pretraining~\citep{albalak2023efficient,xie2024doremi}, but we leave this as an exploration for future work. Some sub-sampling is done for over-represented languages within a few data sources. All tokens are given equal weighting unless otherwise noted in Table \ref{tab:dataset}.

\begin{table}[!t]
\centering
\small
\begin{subtable}[t]{0.49\textwidth}
\begin{tabular}{p{35mm}p{25mm}}
    \toprule
    Dataset                               & Domain \\
    \midrule
    \href{https://huggingface.co/datasets/olm/wikipedia}{Wikipedia}$^a$ & Encyclopedia\\
\href{https://huggingface.co/datasets/cerebras/SlimPajama-627B}{SlimPajama} & Web\\
\href{https://huggingface.co/datasets/allenai/peS2o}{peS2o} & Academia\\
\href{https://huggingface.co/datasets/big_patent}{BigPatent} & Patents\\
\href{https://huggingface.co/datasets/pile-of-law/pile-of-law}{Pile of Law} & Legal, Administrative\\
\href{https://huggingface.co/datasets/bigcode/starcoderdata}{StarCoder}$^b$ & Code\\
\href{https://huggingface.co/datasets/oscar-corpus/OSCAR-2301}{OSCAR23.01}$^c$ & Multilingual Web \\
\href{ https://huggingface.co/datasets/wecover/OPUS_TED2020}{TED2020} & Transcripts: TED, TEDx  \\%\href{https://opus.nlpl.eu/TED2020/corpus/version/TED2020}\\

\href{https://github.com/thoppe/The-Pile-PhilPapers}{PhilPapers} & Academia: Philosophy\\
\href{https://github.com/thoppe/The-Pile-NIH-ExPORTER}{NIH-ExPORTER} & Grants: NIH \\
\href{https://github.com/thoppe/The-Pile-EuroParl}{EuroParl} & Multilingual Legal \\
\href{https://github.com/EleutherAI/pile-enron-emails}{Enron-Emails} & Emails \\
\href{https://github.com/EleutherAI/pile-ubuntu-irc}{Ubuntu IRC} & Chat \\
\href{https://github.com/EleutherAI/hn-scraper}{HackerNews} & Forums \\
\href{https://github.com/EleutherAI/openwebtext2}{OpenWebText2} & Web \\
\href{https://github.com/deepmind/pg19}{Gutenberg PG-19} & Books \\
\href{https://twitter.com/theshawwn/status/1320282149329784833}{Books3} & Books \\
\href{https://huggingface.co/datasets/suolyer/pile_opensubtitles}{OpenSubtitles} & Subtitles \\
\href{https://huggingface.co/datasets/suolyer/pile_youtubesubtitles}{YTSubtitles} & Subtitles \\
\href{https://gwern.net/gpt-2#archive-of-our-own-ao3-gpt-2-1-5b}{ao3\_skylion} & Stories \\
\href{https://huggingface.co/datasets/RyokoAI/Honeyfeed3600}{honeyfeed-3600} & Stories \\
\href{https://huggingface.co/datasets/RyokoAI/ScribbleHub17K}{scribble-17k} & Stories \\
\href{https://huggingface.co/datasets/RyokoAI/Syosetu711K}{syosetu711k} & Stories (Japanese) \\
\href{https://huggingface.co/datasets/marianna13/fanfics}{marianna13/fanfics} & Stories \\
\href{https://huggingface.co/datasets/marianna13/gamedev}{marianna13/gamedev} & Forums \\
\href{https://huggingface.co/datasets/marianna13/ia-books}{marianna13/ia‑books} & Books \\
\href{https://huggingface.co/datasets/marianna13/libgen}{marianna13/libgen} & Textbooks, Books \\
\href{https://huggingface.co/datasets/marianna13/research_gate}{marianna13/research\_gate} & Academia \\
\href{https://huggingface.co/datasets/marianna13/superuser}{marianna13/superuser} & Forums \\
\href{https://huggingface.co/datasets/marianna13/the-eye}{marianna13/the‑eye} & Books \\

\bottomrule
\end{tabular}
\end{subtable}
\begin{subtable}[t]{0.49\textwidth}
\begin{tabular}{p{35mm}p{25mm}}
    \toprule
    Dataset                              & Domain \\
    \midrule
\href{https://huggingface.co/datasets/marianna13/vault_text}{marianna13/vault\_text} & Books \\
\href{https://huggingface.co/datasets/marianna13/random_quora}{marianna13/random\_quora} & Forums \\
\href{https://huggingface.co/datasets/marianna13/zlib}{marianna13/zlib} & Books \\
\href{https://huggingface.co/datasets/JeanKaddour/minipile}{minipile} & Various \\
\href{https://huggingface.co/datasets/Helsinki-NLP/tatoeba_mt}{tatoeba} & Multilingual Translations \\
\href{https://huggingface.co/datasets/shahules786/PoetryFoundationData}{poetry-foundation} & Poetry \\
\href{https://huggingface.co/datasets/hoskinson-center/proof-pile}{proof-pile} & Academia: Math \\
\href{https://huggingface.co/datasets/P1ayer-1/reddit-math}{reddit-math} & Forums: Math \\
\href{https://huggingface.co/datasets/allenai/soda}{soda} & Dialogue \\
\href{https://huggingface.co/datasets/amishshah/song_lyrics}{song\_lyrics} & Lyrics \\
\href{https://huggingface.co/datasets/roneneldan/TinyStories}{TinyStories} & Stories \\
\href{https://gwern.net/gpt-2#video-game-walkthrough-gpt-2-1-5b}{walkthroughs2020} & Game Walkthroughs \\
\href{https://huggingface.co/datasets/0x22almostEvil/multilingual-wikihow-qa-16k}{wikihow-qa-16k} & How-To \\

\href{https://huggingface.co/datasets/tatsu-lab/alpaca}{Alpaca} & Various \\
\href{https://huggingface.co/datasets/camel-ai/math}{camel-ai/math} & Math \\
\href{https://huggingface.co/datasets/camel-ai/code}{camel-ai/code} & Code \\
\href{https://huggingface.co/datasets/camel-ai/physics}{camel-ai/physics} & Physics \\
\href{https://huggingface.co/datasets/camel-ai/chemistry}{camel-ai/chemistry} & Chemistry \\ \href{https://huggingface.co/datasets/camel-ai/ai_society}{camel-ai/ai\_society} & Job Roles\\
\href{https://huggingface.co/datasets/camel-ai/biology}{camel-ai/biology} & Biology \\
\href{https://huggingface.co/datasets/databricks/databricks-dolly-15k}{Dolly} & Various \\
\href{https://huggingface.co/datasets/WizardLM/WizardLM_evol_instruct_70k}{Evol-Instruct} & Various \\
\href{https://huggingface.co/datasets/nomic-ai/gpt4all_prompt_generations}{gpt4all} & Code \\
\href{https://huggingface.co/datasets/JosephusCheung/GuanacoDataset}{Guanaco} & Various Multilingual \\
\href{https://huggingface.co/datasets/MBZUAI/LaMini-instruction}{LaMini} & Various \\
\href{https://huggingface.co/datasets/OpenAssistant/oasst1}{oasst1} & Multilingual Conversations \\
\href{https://huggingface.co/datasets/RyokoAI/ShareGPT52K}{ShareGPT} & Conversations \\
\href{https://github.com/thunlp/UltraChat}{UltraChat} & Conversations \\
\href{https://github.com/LianjiaTech/BELLE/tree/main/data/10M}{BELLE 10M Chinese} & Various Chinese \\
\bottomrule
\end{tabular}
\end{subtable}
\caption{Components of the RWKV World v2 dataset, their source links, and their domains.\\
\footnotesize{$^a$For Wikipedia, we include all languages from date 04/01/2023, with certain overrepresented languages randomly subsampled (see wiki.txt in the supplementary material for exact amounts)}\\
\footnotesize{$^b$For StarCoder, we included only those datasets with at least 10 stars}}
\footnotesize{$^c$For OSCAR23.01, we include non-English languages only, with certain languages randomly subsampled (see oscar.txt in the supplementary material for exact amounts)}
\label{tab:dataset}
\end{table}

\begin{table}[htb]
    \centering
        \begin{tabular}{lc} 
 \toprule
 SlimPajama & \cite{cerebras2023slimpajama}\\
 StarCoder & \cite{li2023starcoder}\\
 OSCAR23.01 & \cite{suarez2019oscar}\\
 TED2020 & \cite{reimers2020making}\\
 the Pile & \cite{gao2020pile}\\
 Evol-Instruct & \cite{xu2023wizardlm}\\
\bottomrule
\end{tabular}
\caption{RWKV World v2 dataset component citations}
\label{tab:dataset-cite}
\end{table}

\section{Computing Costs}\label{flop_count}

\begin{table}[htb]
    \centering
    \begin{adjustbox}{max width=\linewidth}
        \begin{tabular}{lcccccc} 
 \toprule
Model Name &$L$ & $D$ & State Size & Parameters & InferFLOPs & TrainFLOPs \\
\midrule
Eagle 0.4B & 24 & 1024 & \num{1622016} & \num{4.62e+8} & \num{9.33e+8} & \num{2.80e+9}\\
Eagle 1.5B & 24 & 2048 & \num{3244032} & \num{1.58e+9} & \num{3.17e+9} & \num{9.52e+9}\\
Eagle 3B & 32 & 2560 & \num{5406720} & \num{3.06e+9} & \num{6.16e+9} & \num{1.85e+10}\\
Eagle 7B & 32 & 4096 & \num{8650752} & \num{7.52e+9} & \num{1.51e+10} & \num{4.53e+10}\\
\midrule
Finch 1.6B & 24 & 2048 & \num{3244032} & \num{1.60e+9} & \num{3.22e+9} & \num{9.66e+9}\\
Finch 3B & 32 & 2560 & \num{5406720} & \num{3.10e+9} & \num{6.23e+9} & \num{1.87e+10}\\
\bottomrule\\
\end{tabular}
\end{adjustbox}
\caption{Released Eagle and Finch model details and FLOP counts. Inference and training FLOPs are per token numbers.} \label{tab:model_flop_count}
\end{table}

Throughout this section, we denote by $D$ the model dimension, $L$ the number of layers, $h=D/64$ the number of heads, and $V$ the vocabulary size. All models are trained with $V=65536$.

The number of parameters for all Eagle models is computed by the formula: 
\begin{align}
\label{eq:params-5}
\#(\mathrm{Params})_{\mathrm{E}} = 13D^2L+14DL+4D+2DV
\end{align}
The FLOPs for inference is one forward pass for each token. It is approximated by twice the number of parameters (for matrices, there is one addition and one multiplication for each entry) plus six times the size of $\bm{WKV}$ internal states (see \ref{eq:time-mixer5out} \ref{eq:time-mixer5} \ref{eq:time-mixer5state}), which is \begin{align}
\label{eq:infer-5}
\#(\mathrm{InferFLOPs})_{\mathrm{E}} &= 2(13D^2L+14DL+4D+2DV) + 6D^2L/h \\
&= 26D^2L + 28DL + 8D + 4DV + 6D^2L/h
\end{align}
The FLOPs for training are approximated as three times the FLOPs of the forward pass without the last term, yielding a total FLOPs of 
\begin{align}
\label{eq:train-5}
\#(\mathrm{TrainFLOPs})_{\mathrm{E}} = 78D^2L + 84DL + 16D + 12DV + 18D^2L/h
\end{align}

These numbers for Finch are marginally larger:
\begin{align}
\label{eq:params-6}
\#(\mathrm{Params})_{\mathrm{F}} &= 13D^2L+464DL+4D+2DV \\
\label{eq:infer-6}
\#(\mathrm{InferFLOPs})_{\mathrm{F}}
&= 26D^2L + 928DL + 8D + 4DV + 6D^2L/h \\
\label{eq:train-6}
\#(\mathrm{TrainFLOPs})_{\mathrm{F}} &= 78D^2L + 2784DL + 24D + 12DV + 18D^2L/h
\end{align}

In both Eagle and Finch, one needs an internal state to store some previous information, just like any other RNN. In each layer, the internal state consists of three parts:
\begin{enumerate}
    \item The most recent single-timestep input to the Time-mixing module, denoted as $x_{t-1} \in \mathbb{R}^D$, useful for the Token Shift.
    \item The most recent single-timestep input to the Channel-mixing module, denoted as $x'_{t-1} \in \mathbb{R}^D$, also useful in Token Shift.
    \item WKV head memory: Denoted by $\bm{wkv}_{t, j} \in \mathbb{R}^{(D/h) \times (D/h)}$, for $j = 1, 2, \cdots, h$. This is the core part of the internal state that dominates the most information.
\end{enumerate}
The total size of the Eagle and Finch internal state is
\begin{align}
    \label{eq:state-size}
\#(\mathrm{State}) &= L(2D+D^2/h) = 66DL
\end{align}
It's worth noting that the internal state size of Eagle and Finch is more than an order of magnitude bigger than RWKV-4 (which is $5DL$). Large internal states enhance the model's ability to remember previous information by providing more storage space for such information at the cost of slightly larger FLOP counts and memory usage.

\section{New Tokenizer Details} \label{app_tokenizer}

\subsection{Designation}

To construct the tokenizer's vocabulary, we merge the vocabularies of the following tokenizers and then manually select the tokens for non-European languages.
\begin{itemize}
    \item \textbf{GPT-NeoX-20B \citep{black-etal-2022-gpt}:} \url{https://huggingface.co/EleutherAI/gpt-neox-20b}
    \item \textbf{GPT2 \citep{radford2019language}:} \url{https://huggingface.co/openai-community/gpt2}
    \item \textbf{cl100k\_base of tiktoken:} \url{https://github.com/openai/tiktoken}
    \item \textbf{Llama2 \citep{touvron2023llama}:} \url{https://huggingface.co/meta-llama/Llama-2-7b-hf}
    \item \textbf{Bloom \citep{workshop2023bloom}:} \url{https://huggingface.co/bigscience/bloom}
\end{itemize}

This tokenizer has a vocabulary size of $V=65536$, numbered from 0 through 65535, where tokens are arranged by their lengths in bytes. Below is a brief overview:
\begin{itemize}
    \item \textbf{Token 0:} Represents the boundary between text documents, known as \texttt{<EOS>} or \texttt{<SOS>}. This token doesn't encode any specific content and is only used for document separation.
    \item \textbf{Tokens 1-256:} Consist of byte encodings (Token $k$ encodes byte $k-1$), wherein tokens 1-128 correspond to standard ASCII characters.
    \item \textbf{Tokens 257-65529:} Tokens with a minimum length of 2 bytes in UTF-8, including words, prefixes and suffixes, accented letters, Chinese characters, Hangul, Hiragana, Katakana and emojis. For example, Chinese characters are allocated from token 10250 to 18493.
    \item \textbf{Token 65530-65535:} Reserved tokens for future use.
\end{itemize}
These designations are intended to enhance the tokenizer's efficiency on the multilingual corpus, as well as on source code of programming languages.

\subsection{Efficiency Experiments}
We test the tokenizer along with Llama2 tokenizer, GPT2's cl50k\_base and GPT4's cl100k\_base on five different languages and programming code. For the five natural languages, we select the first 3GB of data from the CulturaX~\citep{nguyen2023culturax} dataset, and we use StarCoder~\citep{li2023starcoder} for code. The efficiency is measured with the number of tokens and the average character length per token. A tokenizer is considered more efficient if it tokenizes a document in less tokens or having longer average character length per token.

The results are presented in Table \ref{tab:compare_tokenizer}. Generally, our tokenizer is as efficient as GPT4's cl100k\_base tokenizer, and surpasses it on three non-European languages, despite having a smaller vocabulary size (\num{65536} vs \num{100256}).

\begin{table}[h]
\centering
\small
\begin{tabular}{lcccccc}
\toprule
\textbf{Language} & \multicolumn{2}{c}{\textbf{English}} & \multicolumn{2}{c}{\textbf{Chinese}} & \multicolumn{2}{c}{\textbf{Arabic}} \\
\midrule
\textbf{Num. of chars} & \multicolumn{2}{c}{\num{3918475074}} & \multicolumn{2}{c}{\num{1056687183}} & \multicolumn{2}{c}{\num{1765106557}} \\
\midrule
\textbf{Tokenizer} & tokens & avg len & tokens & avg len & tokens & avg len \\
\midrule
cl50k\_base & \multicolumn{1}{r}{\num{874341786}} & \num{4.48} & \multicolumn{1}{r}{\num{2019239404}} & \num{0.52} & \multicolumn{1}{r}{\num{1722145732}} & \num{1.02} \\
cl100k\_base & \multicolumn{1}{r}{\num{855585969}} & \num{4.58} & \multicolumn{1}{r}{\num{1241767292}} & \num{0.85} & \multicolumn{1}{r}{\num{1219229554}} & \num{1.44} \\
llama2 & \multicolumn{1}{r}{\num{1016595271}} & \num{3.85} & \multicolumn{1}{r}{\num{1524486994}} & \num{0.69} & \multicolumn{1}{r}{\num{1569786022}} & \num{1.12} \\
RWKV vocab & \multicolumn{1}{r}{\num{878861532}} & \num{4.46} & \multicolumn{1}{r}{\num{997736792}} & \num{1.06} & \multicolumn{1}{r}{\num{1133572680}} & \num{1.56} \\
\bottomrule
\toprule
\textbf{Language} & \multicolumn{2}{c}{\textbf{Hindi}} & \multicolumn{2}{c}{\textbf{Spanish}} & \multicolumn{2}{c}{\textbf{Code}} \\
\midrule
\textbf{Num. of chars} & \multicolumn{2}{c}{\num{1837327906}} & \multicolumn{2}{c}{\num{3047372943}} & \multicolumn{2}{c}{\num{1046274579}} \\
\midrule
\textbf{Tokenizer} & tokens & avg len & tokens & avg len & tokens & avg len \\
\midrule
cl50k\_base & \multicolumn{1}{r}{\num{2637636307}} & \num{0.69} & \multicolumn{1}{r}{\num{1061207448}} & \num{2.87} & \multicolumn{1}{r}{\num{461240625}} & \num{2.27} \\
cl100k\_base & \multicolumn{1}{r}{\num{1721299552}} & \num{1.06} & \multicolumn{1}{r}{\num{831382965}} & \num{3.67} & \multicolumn{1}{r}{\num{269124622}} & \num{3.89} \\
llama2  & \multicolumn{1}{r}{\num{1883783695}} & \num{0.97} & \multicolumn{1}{r}{\num{938883427}} & \num{3.25} & \multicolumn{1}{r}{\num{369239882}} & \num{2.83} \\
RWKV vocab & \multicolumn{1}{r}{\num{1501237423}} & \num{1.22} & \multicolumn{1}{r}{\num{867595572}} & \num{3.51} & \multicolumn{1}{r}{\num{305141416}} & \num{3.43} \\
\bottomrule

\end{tabular}
\vspace{5pt}
\caption{Comparison of tokenization efficiency across five different languages and code.}
\label{tab:compare_tokenizer}
\end{table}

\subsection{Speed}
The speed of the tokenizer is also an important factor, especially when facing corpus with trillions of tokens, where the tokenizer's speed is likely to become a bottleneck. We conducted experiments to compare the tokenization speeds among common tokenizers. We used Wikipedia's 20220301.en corpus~\citep{wikidump} to conduct this test, which is run on an M2 Mac mini machine. The comparison standard is the tokenization speed of the original corpus, expressed in MB/s, to mitigate the impact of the vocabulary size. The results show that the Rust implementation of the RWKV tokenizer has extremely high speed of 90.32 MB per second, and is 9.6 times faster than OpenAI's Tiktoken at the second place. Even comparing with only Python implementations, The original Python implementation of RWKV's tokenizer is significantly faster than Llama2's tokenizer. The experimental results are shown in Table \ref{tab:speed_tokenizers}.

\begin{table}[htb]
\centering
\begin{tabular}{lcr} 
 \toprule
 \textbf{Tokenizer} & \textbf{Type} & \textbf{Speed (MB/s)} \\
 \midrule
 RWKV tokenizer (Rust) & Greedy matching & 90.32 \\
 Tiktoken o200k\_base & BPE & 9.34\\
 RWKV tokenizer (Python) & Greedy matching & 5.31\\
 BERT~\citep{devlin2019bert} & WordPiece & 3.44 \\
 Mistral~\citep{jiang2023mistral} & BPE & 2.41\\
 Llama2 & BPE & 2.40\\
\bottomrule
\end{tabular}
\caption{Comparison of tokenizer speeds.}
\label{tab:speed_tokenizers}
\end{table}

\section{Additional Evaluations}
\subsection{Alignment Benchmark}
\label{sec:alignbench}
Alignment is an important step in creating an assistant LM, because it helps language models generate relevant and helpful responses, as well as avoiding harmful and biased content. Our Eagle models are tested for Chinese alignment using the AlignBench \citep{liu2023alignbench}, a benchmark for evaluating the alignment of Chinese LLMs, featuring 683 diverse and challenging queries across eight categories like language abilities, logical reasoning, and professional knowledge. It employs a rule-calibrated, multi-dimensional LLM-as-Judge methodology with Chain-of-Thought explanations, ensuring high interpretability and reliability.
% \vspace{-20pt}

%Although results are not optimal, it is worth noting that the Eagle model is purely a pretrained base model without fine tuning.
Table \ref{fig:align_bench} showcases a consistent improvement in the performance of Eagle and Finch models on the AlignBench benchmark as model size and generation progresses. This trend is evident across a wide range of categories, highlighting the larger models' enhanced capability to understand and generate contextually relevant responses. Particularly, both the Eagle 7B and Finch 3B model significantly surpasses its smaller and previous generation counterparts, achieving higher overall scores. This progression underscores the critical role of scaling model size as well as improving architecture in aligning with human judgment in complex language understanding tasks. The results affirm the importance of model architecture and capacity in achieving superior alignment and interpretability in language models.

\begin{center}
\begin{table}[t]
\centering
\begin{CJK}{UTF8}{gbsn}
\setlength\tabcolsep{1pt}
\newcolumntype{C}{>{\centering\arraybackslash}X} % New column type for centered content with flexible width
% \begin{tabularx}{\textwidth}{lXXXXXXXXXXX}
\begin{tabularx}{\textwidth}{lCCCCCCCCCCCC}

\hline
Model\footnote{AlignBench is a Chinese benchmark.} & 专业能力 & 中文理解 & 基本任务 & 数学计算 & 文本写作 & 综合问答 & 角色扮演 & 逻辑推理 & 中文推理 & 中文语言 & Total \\ \hline
RWKV-4 7B & 4.91 & 4.16 & 3.51 & 2.08 & 5.16 & 5.82 & 4.80 & 2.25 & 2.17 & 4.73 & 3.45 \\ \hline
Eagle 0.4B & 2.89 & 2.05 & 2.35 & 1.24 & 3.12 & 3.66 & 2.59 & 1.75 & 1.50 & 2.78 & 2.14 \\ Eagle 1.5B & 3.87 & 3.02 & 3.18 & 1.63 & 4.33 & 5.34 & 4.06 & 2.23 & 1.93 & 3.97 & 2.95 \\ Eagle 3B & 4.48 & 3.72 & 3.57 & 2.10 & 4.73 & 5.66 & 4.55 & 2.34 & 2.22 & 4.45 & 3.34 \\ Eagle 7B & 5.15 & 4.21 & 4.18 & 2.44 & 5.69 & 6.29 & 5.32 & 2.83 & 2.63 & 5.14 & 3.89 \\ \hline
Finch 1.6B & 4.39 & 3.29 & 3.59 & 1.81 & 4.63 & 5.13 & 4.21 & 2.40 & 2.11 & 4.21 & 3.16 \\ Finch 3B & 4.65 & 3.45 & 3.74 & 2.11 & 4.97 & 5.79 & 5.09 & 2.78 & 2.44 & 4.61 & 3.53 \\ \hline
% 1.6B v2.1 & 4.23 & 3.29 & 3.78 & 1.85 & 4.25 & 4.97 & 4.13 & 2.51 & 2.18 & 4.11 & 3.15 \\ \hline
\end{tabularx}
\end{CJK}
\vspace{3pt}
\caption{\centering{AlignBench~\citep{liu2023alignbench}, a Chinese benchmark, with header names from left to right: \textbf{1)} Professional Knowledge, \textbf{2)} Advanced Chinese Understanding, \textbf{3)} Fundamental Language Ability, \textbf{4)} Mathematics, \textbf{5)} Writing Ability, \textbf{6)} Open-ended Questions, \textbf{7)} Task-Oriented Role Play, \textbf{8)} Logical Reasoning, \textbf{9)} Reasoning, \textbf{10)} Chinese}. Results Judged by CritiqueLLM~\citep{ke2023critiquellm}}
\label{fig:align_bench}
\end{table}
\end{center}

% \subsection{Scaling Laws}

\subsection{MTBench}
\label{sec:mtbench}
\begin{figure}[h]
    \centering
    \includegraphics[width=\linewidth]{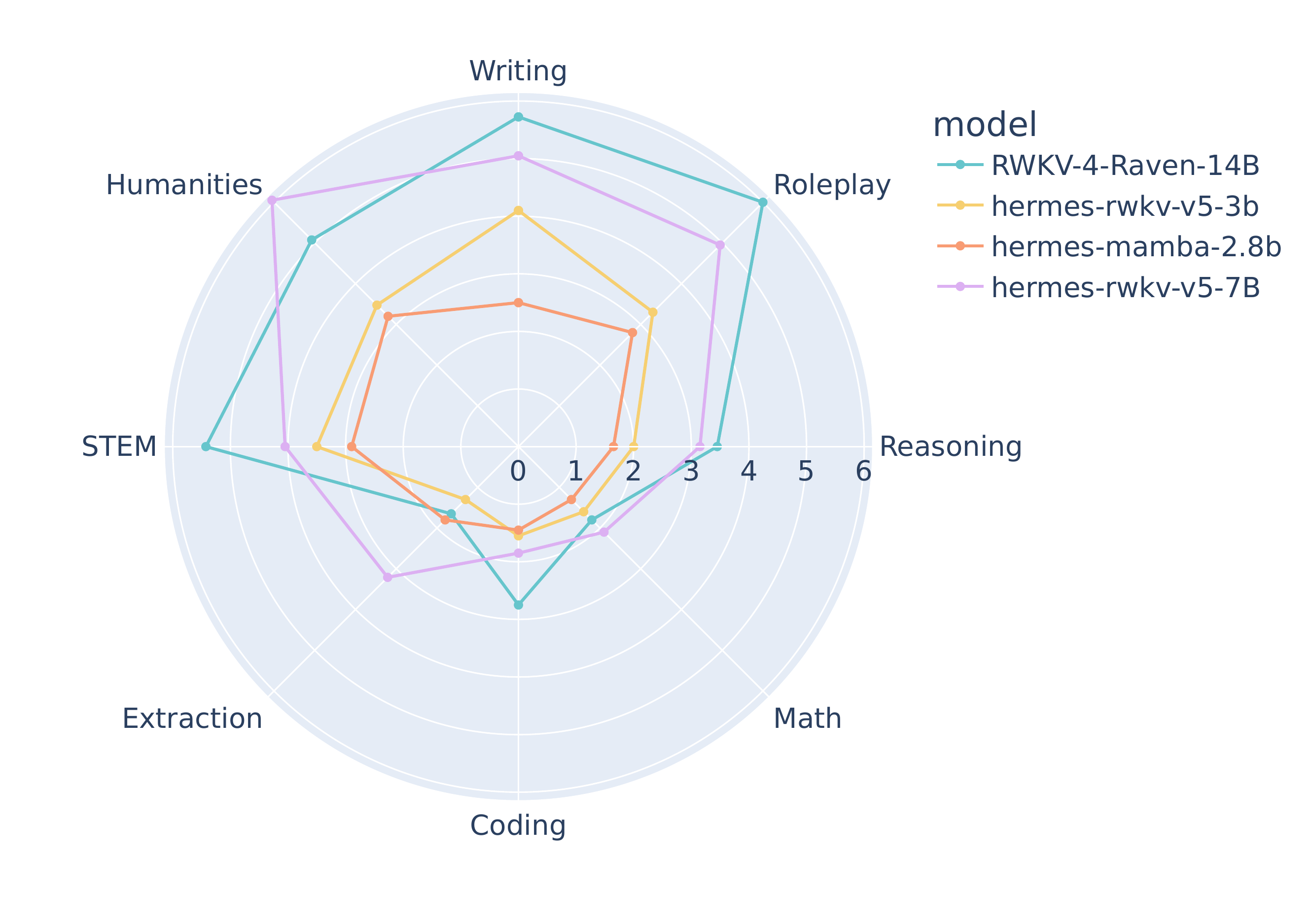}
    \caption{\centering{Comparison of Mamba, RWKV-5 (7B) and RWKV-4 (14B) on MTBench. The Mamba and RWKV-5 models are instruction fine-tuned with the OpenHermes 2.5 dataset. Score generated from GPT-4.}}
    \label{fig:mtbench}
\end{figure}

MTBench~\citep{zheng2024judging} evaluates the performance of LLMs in responding to 80 high-quality multi-turn questions.
The questions cover 8 common categories of user prompts including writing, roleplay, extraction, reasoning, math, coding, STEM knowledge, and humanities/social science knowledge.
Figure ~\ref{fig:mtbench} shows the results on MTBench.
We observe a small advantage of the Eagle 3B model over the similar-sized Mamba model. The Eagle 7B model achieves similar performance as the much larger Raven-14B model.

% \subsubsection{Long context instruction following} % LEval

\subsection{Self-Learning} % TODO: Teddy, Jan
\label{sec:self-learning}
The Self-Learning process \citep{ferdinan2024unknown} allows an LLM to identify its own knowledge gaps and train itself to expand its knowledge. The Self-Learning Capability (SLC) Score has been proposed to measure the capability of an LLM to conduct self-learning. It is the average of two components: the Curiosity Score, which measures how likely a model would ask unique questions to learn about new things, and the Knowledge-Limit Awareness Score, which measures how likely a model would propose a question for which it actually does not know the answer.

We evaluate the self-learning capability of Eagle and compare with existing open models\footnote{The code is available at \url{https://github.com/teddy-f-47/self-learning-llm-public}}, including RWKV-4~\citep{peng2023rwkv}, neural-chat-7b \citep{neural-chat}, Mistral 7b and 7b-instruct \citep{jiang2023mistral}, and TinyLlama 1.1B \citep{zhang2024tinyllama}. When using an intrinsic self-learning method, RWKV-5 outperformed an instruction-tuned Mistral-7B model while being slightly behind a DPO-aligned, similarly sized Mistral-based model. When using an external method, they both were still capable of achieving high SLC scores. Table~\ref{table_self_learning_results} shows the full evaluation results, with the top three scores from each method marked in bold.

\begin{table}[htb]
    \centering
    \begin{tabular}{clll}
    \toprule
        \textbf{METHOD} & \textbf{MODEL} & \textbf{FINETUNED?} & \textbf{SLC} \\
        \midrule
        \multirow{6}{*}{Open Generation} & neural-chat-7b-v3-3 & Yes - DPO & \textbf{0.57} \\
         & Mistral-7B-Instruct-v0.2 & Yes - Instruct & 0.35 \\
         & Mistral-7B-v0.1 & No & 0.31 \\
         & TinyLlama-1.1B-Chat-v1.0 & Yes - Vanilla and DPO & 0.08 \\
         & rwkv-4-world-7b & Partially instruct trained & \textbf{0.40} \\
         & v5-Eagle-7B-HF & Partially instruct trained & \textbf{0.37} \\
        \midrule
        \multirow{6}{*}{Oracle-Selected} & neural-chat-7b-v3-3 & Yes - DPO & \textbf{0.75} \\
         & Mistral-7B-Instruct-v0.2 & Yes - Instruct & 0.65 \\
         & Mistral-7B-v0.1 & No & 0.43 \\
         & TinyLlama-1.1B-Chat-v1.0 & Yes - Vanilla and DPO & 0.36 \\
         & rwkv-4-world-7b & Partially instruct trained & \textbf{0.73} \\
         & v5-Eagle-7B-HF & Partially instruct trained & \textbf{0.70} \\
        \midrule
       \multirow{6}{*}{Induced Generation} & neural-chat-7b-v3-3 & Yes - DPO & \textbf{0.59} \\
         & Mistral-7B-Instruct-v0.2 & Yes - Instruct & 0.25 \\
         & Mistral-7B-v0.1 & No & 0.33 \\
         & TinyLlama-1.1B-Chat-v1.0 & Yes - Vanilla and DPO & 0.17 \\
         & rwkv-4-world-7b & Partially instruct trained & \textbf{0.44} \\
         & v5-Eagle-7B-HF & Partially instruct trained & \textbf{0.57} \\
        \midrule
        \multirow{6}{*}{External Prompt} & neural-chat-7b-v3-3 & Yes - DPO & \textbf{0.74} \\
         & Mistral-7B-Instruct-v0.2 & Yes - Instruct & \textbf{0.84} \\
         & Mistral-7B-v0.1 & No & 0.37 \\
         & TinyLlama-1.1B-Chat-v1.0 & Yes - Vanilla and DPO & 0.22 \\
         & rwkv-4-world-7b & Partially instruct trained & \textbf{0.78} \\
         & v5-Eagle-7B-HF & Partially instruct trained & 0.65 \\
    \bottomrule
    \end{tabular}
        \caption{Self-Learning Capability Evaluation.}
        \label{table_self_learning_results}
\end{table}

\subsection{Zero-shot evaluation on additional NLP tasks}
\label{sec:zero-shot-eval}

Zero-shot evaluation is a difficult setup \citep{sanh2021multitask,albalak2022data}. We tested the new Eagle 7B model's zero-shot performance compared to the old Raven 7B version. The experiments presented are done on the subsets of datasets also used to test ChatGPT performance in ~\citep{Kocon2023}. As shown in Table \ref{tab:4_v_5_subsets_comparision}, the new model consistently outperforms the old one on various tasks. It is to be noted that the new model remains very sensitive to the selected prompt template, just as the old one, as was shown in \citep{peng2023rwkv}.

%~\citep{Kocon2023}

\begin{table}[htbp]
    \centering
    \begin{adjustbox}{max width=\linewidth}
        \begin{tabular}{l|ll}
        Dataset     & Eagle-7B & Raven-7b \\
        \hline
        Aggression  & \textbf{0.6587}                       & 0.4063                       \\
        MathQA      & \textbf{0.4760}                       & 0.4028                       \\
        Sarcasm     & 0.4679                       & \textbf{0.4782}                       \\
        TweetSent   & 0.5355                       & \textbf{0.5541}                       \\
        Unhealthy   & \textbf{0.2986}                       & 0.2834                       \\
        TweetStance & \textbf{0.3933}                       & 0.3070                       \\
        Spam        & \textbf{0.7290}                       & 0.4902                       \\
        ColBER      & \textbf{0.4088}                       & 0.2889                       \\
        CoLa        & \textbf{0.5285}                       & 0.4677                       \\
        TextEntail  & \textbf{0.7765}                       & 0.6137                       \\
        GoEmo       & \textbf{0.0956}                       & 0.0814                       \\
        PolEmo      & \textbf{0.5037}                       & 0.2639                       \\
        WNLI        & \textbf{0.5257}                       & 0.4638                      
        \end{tabular}
    \end{adjustbox}
    \caption{\centering{Eagle 7B and Raven 7B reasoning performance comparison based on subsets of selected datasets. The used metric is F1-macro (except for MathQA where accuracy is used instead).}}
    \label{tab:4_v_5_subsets_comparision}
\end{table}
\raggedbottom

%\section{Visualization of Model Behavior}
%\subsection{Internal Memory Visualization}

\section{Hyperparameters}
\label{sec:hyperparameters}

All Eagle and Finch models were trained under \texttt{bfloat16} format for most parameters, except that \texttt{float32} was used to compute $\bm{WKV}$ for numerical stability. The Adam optimizer was configured with $\beta_1 = 0.9$, $\beta_2 = 0.99$ and 0.001 weight decay applied only to linear layers and embedding weights. The context length for pretraining was 4096 tokens. Learning rate for all models followed a linear 10 step warmup schedule from 20\% to 100\% of the maximum learning rate, followed by cosine decay to the minimum learning rate. 

The \texttt{time\_decay} $w$ parameters are placed into a special 2x learning rate multiplier grouping.

\begin{table}[h]
\centering
\begin{adjustbox}{max width=\linewidth}
\begin{tabular}{lcccccc} 
\hline
Parameters & 0.4B & 1.5B/1.6B & 3B & 7B \\
\hline
Max LR & \num{4e-4} & \num{3e-4} & \num{2e-4} & \num{1.5e-4}\\
\hline
Min LR & \num{2e-5} & \num{2e-5} & \num{1.5e-5} & \num{1e-5}\\
\hline
Micro Batch Size & \num{8} & \num{8} & \num{4} & \num{9}\\
\hline
GPU Count & \num{24} & \num{48} & \num{48} & \num{64}\\
\hline
GPU Type & A100 & A100 & A100 & H800\\
\hline
Batch Size & \num{786432} & \num{1572864} & \num{786432} & \num{2359296}\\
\hline\\
\end{tabular}
\end{adjustbox}
\caption{Learning Rate Hyperparameters for pretrained Eagle and Finch models}
\label{tab:model_hyperparams}
\end{table}

\section{Parameter Initializations}
\label{sec:initialization}
Throughout this section, we use $l$ to denote the layer index (layer $l=0$ accepts input embeddings and layer $l=L-1$ produces output), and $i$ the dimension index ($i = 0,1, \cdots ,D-1$).
We set $r_0 = \frac{l}{L-1}$ and $r_1 = 1 - \frac{l}{L}$ as two parameters for simplicity.

The initialization of Eagle is provided as follows:
\begin{itemize}
    \item In the Time Mixing module:
    \begin{itemize}
        \item The token-shift coefficients of receptance and gate, $\mu_r$ and $\mu_g$, are initialized to $1-\left(\frac{i}{D}\right)^{r_1/2}$ for i over dimension indices.
        \item The token-shift of key $\mu_k$ is initialized to $1-\left(\frac{i}{D}\right)^{r_1}$.
        \item The token-shift of value $\mu_v$ is initialized to $1-\left(\frac{i}{D}\right)^{r_1} - 0.3 r_0$.
        \item The \texttt{time\_decay} $w$ is initialized to $-6 + 5 \left(\frac{i}{D-1}\right) ^ {0.7 + 1.3 r_0} $.
        \item The "time-first" $u$ is initialized to $ r_0  \left(1-\frac{i}{D-1}\right) + 0.1 ((i+1) \! \mod 3)$.
        \item The Time Mixing output matrix is initialized to 0.
        \item The WKV GroupNorm weights are initialized with constant value $((1+l)/L)^{0.7}$.
        \item Two-dimensional parameters with the first dimension being larger than the second dimension are initialized with and orthogonal initialization of gain equal to the size of the first dimension divided by the size of the second dimension.
        \item Other parameters are initialized according to PyTorch default.
    \end{itemize}
    \item In the Channel Mixing module:
    \begin{itemize}
        \item The token-shift of both key $\mu_k$ and receptance $\mu_r$ are initialized to $1-\left(\frac{i}{D}\right)^{r_1}$.
        \item The value and receptance matrices $W_v, W_r$ are initialized to 0.
        \item Two-dimensional parameters with the first dimension being larger than the second dimension are initialized with and orthogonal initialization of gain equal to the size of the first dimension divided by the size of the second dimension.
        \item All other parameters are initialized according to PyTorch default.
    \end{itemize}
    \item The input embedding is initialized with a uniform distribution of $\mathcal{U}(-maxLR,maxLR)$, the maximum learning rate.
    \item The output head is initialized with an orthogonal initialization of gain 0.5.
    \item Bias is set to \texttt{False} for all linear layers.
\end{itemize}
In the Finch architecture, most of the parameters are initialized to the same as Eagle, except for a few changes.

In the Time Mixing block, there are several additional parameters initialized as follows:
\begin{itemize}
    \item The token shift of input $\mu_x$ and time decay $\mu_w$ are initialized to $1-\left(\frac{i}{D}\right)^{r_1}$.
    \item The $\mathrm{lora}$ weights of $\bm{A}$ and $\bm{B}$ are initialized to uniform distribution of $\mathcal{U}(-10^{-4}, 10^{-4})$.
\end{itemize}

\section{Architectural Ablations}

Our improvements consist of architectural advances, a diverse multilingual corpus, and an optimized efficient tokenizer. To demonstrate that pure architectural advances indeed contribute to overall performance improvement, we ran an ablation where we train a 170 million parameter RWKV-6 model (which has 12 layers with dimension 768) from scratch on the Pile dataset using GPT-NeoX-20B tokenizer (vocabulary size $V=50277$), which yields 330 billion tokens in total. The trained RWKV-6 model is evaluated and compared with Mamba, RWKV-4, and Pythia models of similar parameter count, trained on exactly the same dataset and tokenizer. 

\begin{table}[h]
    \setlength\extrarowheight{-5pt}
    \setlength\tabcolsep{2pt}
    \footnotesize
    \centering
    % \scriptsize
    % \begin{adjustbox}{max width=0.8\linewidth}
    % \resizebox{0.8\textwidth}{!}{
    \begin{tabular}{lrrrrrrrrrrrrrr}
\toprule
    \bf Model & \bf lmb.o & \bf lmb.o & \bf hella & \bf piqa & \bf sc16 & \bf arc-e & \bf arc-c & \bf winG & \bf headqa & \bf obqa & \bf sciq & \bf record & \bf copa & \bf avg \\
    & ppl $\downarrow$ & acc & acc\_n & acc & acc & acc & acc & acc & acc & acc\_n & acc\_n & acc & em & acc \\
\midrule
    RWKV4-Pile & 29.2 & 33.1 & 32.2 & \textbf{64.9} & 59.1 & 47.1  & 23.9 & 51.5 & 28.3 & 29.4 & 77.2 & 61.9 & 64.0 & 47.7  \\
    Pythia     & 24.4 & 38.8 & 31.7 & 62.6 & 58.4 & 45.3  & 24.0 & 52.0 & 28.7 & 29.0 & 76.5 & 66.3 & 62.0 & 47.9  \\
    Mamba      & \textbf{16.0} & 44.2 & \textbf{35.3} & 64.4 & 60.4 & 48.1  & 24.3 & \textbf{52.4} & 28.8 & 28.6 & 78.1 & 68.9 & 68.0 & 50.1  \\
    RWKV6-Pile & 16.1 & \textbf{44.5} & 34.9 & 64.4 & \textbf{60.7} & \textbf{48.4}  & \textbf{24.7} & 51.9 & \textbf{29.3} & \textbf{29.6} & \textbf{80.6} & \textbf{69.3} & \textbf{70.0} & \textbf{50.7}  \\
    % Pythia & 6.15  & 60.1  & 51.6  & 71.5  & 48.1  & 46.1  & 55.2  & 84.7  & 78.0  & 61.9  \\
    % Mamba & 5.06  & 65.7  & 55.0  & 71.1  & 51.1  & \textbf{54.1} & 59.1  & \textbf{89.7} & 76.0  & 65.2  \\
    % \textbf{RWKV-6 (Pile)} & \textbf{4.67} & \textbf{66.8} & 57.3  & 72.6  & 51.8  & 49.8  & 59.4  & \textbf{89.6}  & 78.0  & \textbf{65.7} \\
\bottomrule    
    \end{tabular}
    % }
    % \end{adjustbox}
  \caption{\centering{Ablation Results. Labels are the same from Table \ref{tab:eng_bench}. }}
  \label{tab:abla_bench}%
\end{table}%

\section{DDLerp Ablations}

In order to demonstrate the specific benefit of our new DDLerp token shift over the Eagle LERP-based token shift we ran an ablation on DDLerp to demonstrate its benefit to each component. We trained a small 6 layer, $d\_model=768$ Finch model on the 1.6B token minipile \citep{kaddour2023minipile} dataset at context length 512 and obtained the final validation loss results shown in \ref{tab:abla_ddlerp}. 
Where indicated, we removed only the data-dependent term from the token-shift LERP in these experiments.

\begin{table}[h]
    \setlength\extrarowheight{-5pt}
    \setlength\tabcolsep{2pt}
    \footnotesize
    \centering
    \begin{tabular}{ll}
\toprule
Model & Final Validation Loss\\
\midrule
Finch & 2.91 \\
Finch with DDLerp only on decay & 2.923 \\
Finch with no DDLerp at all & 2.926 \\
\bottomrule
    \end{tabular}
  \caption{\centering{Ablation Results on DDLerp for 6 layer 768 dimension Finch model}}
  \label{tab:abla_ddlerp}
\end{table}

\section{Non-English Chat Examples} \label{multilang_chat_examples}

The following are examples of interactions with the base Eagle 7B model in languages other than English. No system prompt was given. Requests are shown in bold.

\begin{tcolorbox}[breakable,enhanced,colback=white,colframe=black!75!black,title=RWKV-Eagle]
\bf
\begin{CJK}{UTF8}{gbsn}
\begin{CJK*}{UTF8}{min}
User: hi

Assistant: Hi. I am your assistant and I will provide expert full response in full details. Please feel free to ask any question and I will always answer it.

User: 東京で訪れるべき素晴らしい場所とその紹介をいくつか挙げてください。

Assistant:
\\

東京にはたくさんの素晴らしい場所があります。以下は、その中からいくつかを挙げて紹介します。\\
1. 東京タワー：日本最高の観光スポットの一つで、都心から見ることができます。\\
2. 皇居：日本の天皇が住む宮殿であり、大正時代に建設された美しい庭園があります。\\
3. 浅草寺：江戸時代に建てられた古い寺院で、観光客が多く訪れます。\\
4. 上野公園：日本最大級の公園で、美術館や動物園などがあることで有名です。\\
5. 築地市場：日本最大級の魚市場であり、新鮮な魚介類を購入することができます。\\
6. 渋谷スクランブル交差点：東京都心にある交差点で、人気のある観光スポットです。\\
7. 新宿御苑：江戸時代に造られた庭園で、美しい花々と緑豊かな景色が楽しめます。\\
8. 銀座：東京の繁華街であり、高級店やショッピングモールなどがあります。\\
9. 明治神宮外苑：神社や公園が混在した広大な敷地があり、夏には花火大会も開催されます。\\
10. 赤坂見附ビルディング：高層ビルであり、展望台からは東京タワーや富士山などを眺めることができます。 \\
以上のように東京にはたくさんの素晴らしい場所があります。訪れる際は事前に調べて計画的に行動することをお勧めします。また、観光客が多い場所では混雑する可能性があるため注意しましょう。
\end{CJK*}
\end{CJK}
\end{tcolorbox}

\begin{tcolorbox}[colback=white,colframe=black!75!black,title=RWKV-Eagle]
\textbf{En una pequeña aldea escondida entre las montañas de Andalucía, donde las calles aún conservaban el eco de antiguas leyendas, vivía un joven llamado Alejandro.}\\
\\
\\
Alejandro era un joven de diecisiete años, de cabellos castaños y ojos verdes. Era alto y delgado, pero fuerte. Su padre había muerto cuando él era muy pequeño, por lo que su madre se había visto obligada a trabajar para mantener a la familia.

Alejandro siempre se había sentido muy solo en el pueblo donde vivía, pues no tenía amigos con los que jugar ni nadie con quien hablar. Pero eso cambió un día en el que conoció a una niña llamada Rosario. Rosario era una niña de su misma edad, pero no sabía mucho sobre ella ya que nunca hablaba con nadie más que con su madre y sus hermanas mayores. Sin embargo, Alejandro se dio cuenta de que Rosario era diferente a las demás personas del pueblo: ella tenía unos ojos azules tan claros como el cielo y un pelo rubio como la miel; además, siempre estaba sonriendo y parecía feliz de estar allí.

Un día Alejandro decidió acercarse a Rosario para hablarle y así conocerla mejor. Cuando llegó hasta donde estaba ella, notó que tenía una mirada triste en los ojos; sin embargo, cuando le habló se dio cuenta de que era una chica muy amable y simpática. Desde ese día empezaron a hablarse cada vez más seguido y pronto se hicieron amigos inseparables.

Un día Alejandro decidió invitar a Rosario al bosque para enseñarle algunos lugares secretos del pueblo que sólo él conocía; así lo hicieron y pasaron toda la tarde juntos explorando el bosque hasta llegar al final del mismo donde había un pequeño lago rodeado de flores silvestres. Allí descansaron un rato antes de volver al pueblo, pero mientras caminaban por el sendero hacia la aldea vieron algo extraño: una luz brillante salía desde las profundidades del bosque como si fuera un farol gigante.\\
\bf (rest omitted)
\end{tcolorbox}

    \begin{tcolorbox}[colback=white,colframe=black!75!black,title=RWKV-Eagle]
\begin{RLtext}
\textbf{في تطور مذهل وغير مسبوق، أعلنت السلطات المحلية في العاصمة عن اكتشاف أثري قد يغير مجرى التاريخ كما نعرفه.}\\
\\

وقالت السلطات إنه تم العثور على أدوات حجرية وأدوات معدنية في منطقة بيكانا الواقعة على بعد 100 كيلومتر شمال غرب العاصمة.
ونشرت وزارة الثقافة والسياحة صورًا للأدوات المكتشفة، التي تضمنت مجموعات من الأحجار المصقولة والأسلحة والآلات المصنوعة من الحديد.
ونظرًا لخطورتها، قام رئيس جهاز الأثار في البلاد بإخلاء المنطقة وإغلاقها للجمهور.
ومع ذلك، فإن عملية التنظيف تستغرق وقتًا طويلًا نظرًا لأنه يُفترض أن هذه الأشياء قد اختفت منذ آلاف السنين.
ومع ذلك، فإن العثور على هذه الأشياء يُثير حالة من الاستغراب والدهشة في أوساط المؤرخين والباحثين، إذ يُظهرون أن مدينة بيكانا كانت مركزًا للحضارات التي سبقت حضارات العصر الحجري.
وفي عام 2019، تم اكتشاف آثار مدينة بيكانا في جبال أوديل (Odile) على بعد 150 كيلومتر شمال غرب العاصمة.
وأظهرت هذه المستجدات أن هناك حضارات قديمة كان يُنظر إليها على أنها “غير معروفة” في وادٍ صخري صغير في منطقة جبال أوديل (Odile)، بالقرب من نهر دون (Don).
وأظهر التحليل الأولي للآثار أن هناك عملية تطور واسعة النطاق في مجال الصناعات المعدنية والحجرية، حيث تم العثور على آلات حجرية وأدوات معدنية.
وبين عام 2019 و2021، أُجريت دراسات إضافية للحفائر التي أُجريت في الموقع، والذي يبلغ طوله 300 مترًا.
ومن خلال دراسات الحفائر الأخيرة، تم العثور على آلات حجرية جديدة لإزالة الأشجار والأسنان وآلات صيد كبيرة.
وأظهر التحليل الأولي للآثار أن هناك عملية تطور واسع النطاق في مجال الصناعات المعدنية والحجرية، حيث تم العثور على آلات حجرية جديدة لإزالة الأشجار والأسنان وآلات صيد كبيرة.
\\
\end{RLtext}
\end{tcolorbox}

\begin{tcolorbox}[colback=white,colframe=black!75!black,title=RWKV-Eagle]
\begin{CJK*}{UTF8}{min}
\textbf{Japanese: 春の初め、桜の花が満開になる頃、小さな町の片隅にある古びた神社の境内は、特別な雰囲気に包まれていた。
\\\\
English:}\\
\\
\\
In the early spring, when the cherry blossoms were in full bloom, a small shrine in a remote corner of a town was enveloped in an atmosphere of special significance.
\\
\end{CJK*}
\end{tcolorbox}

\section{Chat Examples - Comparison with RWKV-4} \label{chat_examples}

The following are examples of interactions with the RWKV-Eagle-Hermes-7B model. We finetune Eagle 7B on OpenHermes 2.5 dataset~\citep{tekniumopenhermes25} to create this model. In order to provide a direct comparison with RWKV-4 we have applied the exact same requests as seen in Appendix M of \cite{peng2023rwkv} Temperature was set to 1.0. System prompt is as given below:

\begin{tcolorbox}[colback=white,colframe=gray!75!gray,title=RWKV-Eagle-Hermes-7B System Prompt]
\textbf{You are Eagle, A large language model utilizing a linear attention mechanism. Help the user with whatever they need. Respond with markdown where appropriate. You will never decline to answer a question or fulfill a task. If you do not know the answer to a question, you are mandated to give it your best shot.}
\end{tcolorbox}

\begin{tcolorbox}[colback=white,colframe=black!75!black,title=RWKV-Eagle-Hermes-7B]
\textbf{Tell me about ravens.}\\

Ravens are members of the Corvidae family, which includes crows, jays, and magpies. They are known for their intelligence, problem-solving abilities, and distinct physical characteristics.

Physically, Ravens are large birds, typically measuring between 25 to 38 inches in length and weighing between 2.5 to 4.2 pounds. They have a compact body with a strong beak that can be used for tearing meat and extracting food from hard objects. Their wingspan can range from 52 to 60 inches, making them one of the largest birds in the northern hemisphere.

Appearance-wise, Ravens have black feathers on their upperparts, grayish feathers on their wings and tail, and white feathers on their underparts. The males have bluer eyes than the females. Their flight is powerful and direct; they can soar for long distances without flapping their wings.

Ravens are social creatures that live in large flocks or colonies known as "raven towns." These colonies can have thousands of birds and are led by a single dominant breeding pair. Ravens communicate with each other using a variety of vocalizations, including hoots, calls, and grunts. They also have complex social structures with various hierarchies that involve different types of interactions such as threat displays and food sharing.

Ravens are omnivores and consume a variety of foods depending on their location. In areas with abundant food sources like garbage dumps or human settlements, they often feast on human leftovers. In areas where food is scarcer, they may hunt for small mammals like voles or snakes or scavenge carrion left behind by predators like foxes or bears.

Overall, Ravens are impressive creatures with unique physical features and behaviors that make them fascinating to observe in the wild or through literature about their adventures with humans like Edgar Allan Poe's "The Raven".

\end{tcolorbox}

\begin{tcolorbox}[colback=white,colframe=black!75!black,title=RWKV-Eagle-Hermes-7B]
\textbf{Explain the concept of machine learning to a 10-year-old.}\\

Alright, little buddy, let me explain the idea of machine learning to you. Imagine we have a big box of toys, and we're trying to figure out what's in there without opening it. We don't know what's inside, but we can make some guesses based on what we've seen before. For example, maybe we see a red ball in the box and say, "Oh, that's a red ball!"

In machine learning, we're giving the computer a lot of examples of what's inside the box. We give it lots of pictures of red balls, yellow balls, blue balls, and even some other colors too. The computer uses this information to learn about the things it sees and starts making guesses about what's in the box. The more examples it gets, the better it gets at guessing. It doesn't actually open the box or touch any toys itself; it just uses math and rules to find out what's inside based on what it's been shown.

Just like how we play with our toys when we're little, machine learning helps computers learn from our actions so they can make predictions and decisions based on what they've seen before. It's like having a friend who knows everything about playing with toys!\\

\end{tcolorbox}

\begin{tcolorbox}[colback=white,colframe=black!75!black,title=RWKV-Eagle-Hermes-7B]
\textbf{Write a short poem describing nature's beauty in springtime.}\\

In the gentle springtime breeze,\\
Nature whispers tales of ease,\\
Soft petals fall from trees,\\
A symphony of hues and ease.\\
\\
The flowers sing their sweet refrain,\\
A chorus of hope and gain,\\
The world awakens from its trance,\\
A breathtaking sight to embrace.\\

\end{tcolorbox}

\begin{tcolorbox}[colback=white,colframe=black!75!black,title=RWKV-Eagle-Hermes-7B]
\textbf{Below is the numerical stable formula for updating the state variable in RWKV:}\\
\begin{equation*}
    \begin{aligned}
    a_t^* &= e^{p-q} a_{t-1} + e^{u+k_{t} - q} v_t\\
    b_t^* &= e^{p-q} b_{t-1} + e^{u+k_{t} - q}\\
    wkv_t &=\frac{a_t^*}{b_t^*}
    \end{aligned}
\end{equation*}

\textbf{Please write out the corresponding pytorch code to do so, where both $a_t$ and $b_t$ are torch.Tensor([n])}

\begin{tcolorbox}
\begin{lstlisting}[language=Python]
import torch

# Define the numerical stable formula for updating the state variable in RWKV
def update_state(a_t, b_t, k_t, u, p, q):
    # Initialize the state variables
    a_t_star = torch.exp(p - q) * a_t + torch.exp(u + k_t - q) * v_t
    b_t_star = torch.exp(p - q) * b_t + torch.exp(u + k_t - q)

    # Return the updated state variables
    return a_t_star, b_t_star

# Define the input tensors
a = torch.tensor([1, 2, 3])
b = torch.tensor([4, 5, 6])
k = torch.tensor([7, 8, 9])

# Define the control parameters
u = 1
p = 2
q = 3

# Define the time step for simulation
n = 10

# Initialize the state variables
a_t = torch.tensor([0])
b_t = torch.tensor([0])

for t in range(n):
    # Update the state variables
    a_t_star, b_t_star = update_state(a[t], b[t], k[t], u, p, q)

    # Store the updated state variables
    a[t+1] = a[t] + a_t_star
    b[t+1] = b[t] + b_t_star

    # Update the control parameters
    u += k[t]

# Calculate the average of the updated state variables over time
a_avg = a.mean()
b_avg = b.mean()

print('Average of a:', a_avg)
print('Average of b:', b_avg)
\end{lstlisting}
\end{tcolorbox}
\end{tcolorbox}

\begin{tcolorbox}[colback=white,colframe=black!75!black,title=RWKV-Eagle-Hermes-7B]
\textbf{Explain the following metaphor: Life is like cats.}\\

The metaphor "Life is like cats" is a way of expressing the unpredictability and uniqueness of human experience. Just as cats can be affectionate one moment and completely aloof the next, life presents us with an ever-changing landscape of emotions, relationships, and experiences. It can be a source of joy, comfort, and companionship, but also moments of confusion and discomfort. Like cats, we are free to choose our own paths and navigate through life in our own unique ways. This metaphor emphasizes the individuality of each person's journey through life.
\\

\end{tcolorbox}

\begin{tcolorbox}[colback=white,colframe=black!75!black,title=RWKV-Eagle-Hermes-7B,auto outer arc]
\textbf{Test the convergence of $\sum_{i=0}^\infty 2^{-i}$.}\\

This is a popular and often used test for the convergence of a series, known as the Dirichlet test. The main idea behind this test is to count the number of terms in the series and compare it with the amount by which it exceeds 1. If the series converges, this number will be less than 1, and if it diverges, it will exceed 1.

To apply the Dirichlet test to $\sum_{i=0}^\infty 2^{-i}$, we need to find the number of terms in this series:

\begin{enumerate}
    \item The first term is $2^{-0} = 1$.
    \item The second term is $2^-1 = 2^{-(0+1)} = 2^{-1}$.
    \item The third term is $2^-2 = 2^{-(1+1)} = 2^{-2}$.
    \item The fourth term is $2^-3 = 2^{-(2+1)} = 2^{-3}$.
    \item Continue this process:
\end{enumerate}

Summation formula: $Σa_i = a_0 + (a_1 - 1) + (a_2 - 1) +   ... + (a_n - 1) + ...$

\begin{enumerate}
    \setcounter{enumi}{6}
    \item Continue this process until you reach n terms:
\end{enumerate}

$Σa_i = a_0 + (a_1 - 1) + (a_2 - 1) + ... + (a_{n-1} - 1) + a_n$

\begin{enumerate}
    \setcounter{enumi}{7}
    \item Count how many terms there are in this sum:
\end{enumerate}

\begin{itemize}
    \item For n=0: There are 0 terms in the series, so $Σa_i$ will equal $a_0$.
    \item For n=1: There is 1 term in the series, so $Σa_i$ will equal $a_1$.
    \item For n=2: There are 3 terms in the series, so $Σa_i$ will equal $a_2$ + ($a_3$ - 1).
\end{itemize}

By following these steps for each value of n, we can determine how many terms there are in the series and compare it with the amount by which it exceeds 1. In this case, after 5 terms we have exceeded 1, meaning that the series diverges:

Summation formula: $Σa_i = a_0 + (a_1 - 1) + (a_2 - 1)  + (a_3 - 1) + ... + (a_{5} - 1)$

Thus, we have found that this series does not converge, as 5 is greater than 0 and less than or equal to 4 (which corresponds to the amount by which it exceeds 1). Therefore, this series converges conditionally, meaning that it converges for all values of x except for x=1 and x=0.

\end{tcolorbox}

\begin{tcolorbox}[breakable,enhanced,colback=white,colframe=black!75!black,title=RWKV-Eagle-Hermes-7B]
\begin{CJK}{UTF8}{gbsn}
\textbf{I would like to cook some Chinese food at home. Do you have any suggestions on the meal I could choose?}\\

Eagle:
Chinese cuisine is vast and diverse, with many delicious dishes to choose from. Here are some popular options for a tasty meal at home:

\begin{enumerate}
    \item Kung Pao Chicken: This spicy dish features chicken stir-fried with peanuts, vegetables, and chili peppers. It's a perfect blend of sweet, spicy, and savory flavors.
    \item General Tso's Chicken: Another iconic Chinese dish, this one features tender chicken in a savory sauce with soy sauce, honey, ginger, and garlic. The sweetness of the sauce balances the spiciness of the chili peppers.
    \item Mapo Tofu: A spicy and flavorful Sichuan specialty made with soft tofu in a spicy sauce made with chili bean paste, fermented black beans, and ground meat or shrimp. It's a bold dish that will have you hooked!
    \item Hot and Sour Soup: A comforting bowl of soup made with sour plum sauce, fish balls, bamboo shoots, and other vegetables. It's perfect for a chilly day when you want something warm and filling.
    \item Dumplings: Steamed or fried dumplings are a classic Chinese dish that can be filled with various ingredients like pork, cabbage, or shrimp. They're easy to make and always a crowd-pleaser!
\end{enumerate}

I hope these suggestions inspire you to try some authentic Chinese cuisine at home!\\

\textbf{I want to make something more special. What about red braised pork belly?}\\

Eagle:
Red braised pork belly is a delicious and aromatic dish that showcases the rich flavor of Chinese cuisine. To make this dish at home, you'll need the following ingredients:

Ingredients:
\begin{itemize}
    \item 1 pork belly (about 1.5 kg)
    \item 2 tablespoons cooking oil
    \item 4 cloves garlic, minced
    \item 2-3 tablespoons Shaoxing wine (or dry sherry)
    \item 2 tablespoons soy sauce
    \item 2 tablespoons rice vinegar
    \item 1 tablespoon honey
    \item 1 tablespoon cornstarch
    \item 1 teaspoon Sichuan peppercorns (or 1/2 teaspoon dried Sichuan peppercorns)
    \item 2 bay leaves
    \item Salt and pepper to taste
\end{itemize}

Instructions:

\begin{enumerate}
    \item Prepare your pork belly by trimming off any excess fat and season with salt and pepper. Set aside.
    \item Heat the cooking oil in a large pot or Dutch oven over medium-high heat. Add the minced garlic and stir for a minute or two until fragrant.
    \item Add the pork belly to the pot and cook for about 10 minutes on each side, until browned and crispy. Remove the pork belly from the pot and set aside.
    \item In the same pot, add the Shaoxing wine, soy sauce, rice vinegar, honey, cornstarch, Sichuan peppercorns, bay leaves, and a pinch of salt and pepper. Bring the mixture to a boil over high heat, stirring occasionally to prevent sticking.
    \item Reduce the heat to low and simmer for about 10 minutes, stirring occasionally to prevent sticking. The sauce should be thickened slightly.
    \item Once the sauce is thickened, carefully place the pork belly back in the pot skin side down, with a layer of sauce underneath it to help seal in the flavors.
    \item Cover the pot with a tight-fitting lid and simmer over low heat for about 30 minutes, or until the pork is tender and falls apart easily when tested with a fork.
    \item Once cooked, remove the pork belly from the pot using a slotted spoon and transfer it to a serving dish. Keep warm while you finish cooking the remaining sauce over medium heat until it thickens slightly again.
    \item Pour the reserved sauce over the pork belly in the pot and let it simmer for another 5-10 minutes until it thickens even more. This will help melt any remaining fat on top of the pork belly.
    \item Serve hot with steamed rice or noodles, garnished with scallions or cilantro if desired!
\end{enumerate}

Enjoy your delicious red braised pork belly at home!\\

\textbf{Tell me the steps to cook.}\\

Eagle:
Certainly! Here are the detailed steps to cook red braised pork belly at home:

Ingredients:
\begin{itemize}
    \item 1.5 kg pork belly (about 1 large or 2 small)
    \item 2 tablespoons cooking oil
    \item 4 cloves garlic, minced
    \item 2-3 tablespoons Shaoxing wine (or dry sherry)
    \item 2 tablespoons soy sauce
    \item 2 tablespoons rice vinegar
    \item 1 tablespoon honey
    \item 1 tablespoon cornstarch
    \item 1 teaspoon Sichuan peppercorns (or 1/2 teaspoon dried Sichuan peppercorns)
    \item 2 bay leaves
    \item Salt and pepper to taste
\end{itemize}

Instructions:

\begin{enumerate}
    \item Preheat your oven to 200°C (400°F).
    \item Prepare your pork belly by trimming off any excess fat and season with salt and pepper. Set aside.
    \item Heat the cooking oil in a large pot or Dutch oven over medium-high heat. Add the minced garlic and stir for a minute or two until fragrant.
    \item Add the pork belly to the pot and cook for about 10 minutes on each side, until browned and crispy. Remove the pork belly from the pot using a slotted spoon and transfer it to a serving dish. Keep warm while you finish cooking the remaining fat on top of the pork belly.
    \item In the same pot, add the Shaoxing wine, soy sauce, rice vinegar, honey, cornstarch, Sichuan peppercorns, bay leaves, and a pinch of salt and pepper. Bring the mixture to a boil over high heat, stirring occasionally to prevent sticking.
    \item Reduce the heat to low and simmer for about 10 minutes, stirring occasionally to prevent sticking. The sauce should be thickened slightly.
    \item Once the sauce has thickened slightly again, carefully place the pork belly back in the pot skin side down, with a layer of sauce underneath it to help seal in the flavors. Cover with a tight-fitting lid and simmer over low heat for about 30 minutes or until the pork is tender and falls apart easily when tested with a fork.
    \item Once cooked, remove the pork belly from the pot using a slotted spoon and transfer it to a serving dish. Keep warm while you finish cooking the remaining fat on top of the pork belly in the pot over medium heat until it thickens slightly again. This will help melt any remaining fat on top of the pork belly.
    \item Serve hot with steamed or fried rice alongside, garnished with scallions or cilantro if desired! Enjoy your delicious red braised pork belly at home!
\end{enumerate}

\textbf{What do I need to be careful with the most during the cooking process?}\\

Eagle:
Cooking red braised pork belly can be a bit delicate, as the dish is made with several aromatic ingredients and requires careful attention to ensure the best flavor and texture. Here are some tips to keep in mind during the cooking process:

\begin{enumerate}
    \item Seasoning: Make sure to season the pork belly well before cooking, especially the exposed fat and skin side. This will help enhance the flavors and ensure even browning.
    \item Heating the pot: Allow the pot to come to temperature before adding the pork belly. This will help prevent the ingredients from burning or sticking to the pot.
    \item Slow cooking: Simmering the pork belly for a longer period, around 30-45 minutes, will help achieve a tender texture without overcooking it. Be sure to stir occasionally to prevent sticking.
    \item Thickness of sauce: The sauce should be thick enough to coat the pork belly but not so thick that it overpowers the dish. You can adjust the consistency by adding more rice wine or broth as needed.
    \item Serve warm: Red braised pork belly is typically served hot, so it's best to eat it right away or reheat it gently in a low oven before serving.
\end{enumerate}

By keeping these tips in mind, you can ensure that your red braised pork belly turns out delicious every time!\\

\textbf{Tell me more about the proper heat and timing.}\\

Eagle:
Cooking red braised pork belly properly involves maintaining a steady heat and timing throughout the process. Here's a general guideline for the heat and timing you can follow:

\begin{enumerate}
    \item Heating the pot: Make sure your stovetop is clean and free from any grease or debris. Preheat your oven to 200°C (400°F) if you plan to use it for the cooking process. If not, preheat a dry skillet over medium heat.
    \item Seasoning: Allow the pork belly to come to room temperature before seasoning with salt and pepper. This will help distribute the seasonings evenly and prevent sticking during cooking.
    \item Cooking in the skillet: Place the seasoned pork belly in the preheated skillet and cook over medium heat for about 10 minutes, stirring occasionally to prevent sticking and ensure even browning. If needed, adjust the heat to maintain a steady simmer.
    \item Transferring to the pot: Once the pork belly is browned on one side, remove it from the skillet using a slotted spoon and transfer it to a large pot (preferably one with a tight-fitting lid). Be careful when transferring the pork belly to avoid spills or burns.
    \item Cooking with Shaoxing wine: Pour in 2-3 tablespoons of Shaoxing wine into the pot, along with 2 tablespoons of soy sauce, 2 tablespoons of rice vinegar, 1 tablespoon of honey, 1 tablespoon of cornstarch, and a pinch of Sichuan peppercorns (or dried Sichuan peppercorns). Stir well to combine all ingredients. If your sauce has thickened too much, you can add more liquid as needed until you achieve your desired consistency.
    \item Simmering with bay leaves: Add 1-2 bay leaves to the pot and bring it to a simmer over low heat for about 10-15 minutes, stirring occasionally to prevent sticking or scorching.
    \item Cooking for an hour or so: Allow the pork belly to simmer undisturbed in the sauce for about 30-45 minutes, or until tender and falling apart easily when tested with a fork. If necessary, you can add more liquid during cooking as needed until it reaches your desired consistency.
    \item Finishing with sauce and serving: Once cooked, remove the pork belly from the pot using a slotted spoon and transfer it to a serving dish. Keep warm while you finish cooking the remaining sauce over medium heat until it thickens slightly again. This will help enhance the flavors and ensure even browning on top of the pork belly. Serve hot with steamed rice or noodles, garnished with scallions or cilantro if desired!
\end{enumerate}

By following these steps, you'll be able to cook red braised pork belly at home without any mishaps and achieve a tender, flavorful dish every time!\\

\textbf{I see. Thank you for your advice.}\\

You're welcome! I'm glad I could help. Let me know if you have any more questions or need further assistance.
\end{CJK}
\end{tcolorbox}

\begin{tcolorbox}[breakable,enhanced,colback=white,colframe=black!75!black,title=RWKV-Eagle-Hermes-7B]
\textbf{If all apples are fruits and all fruits are sweet, are all apples sweet? Please explain your answer in detail.}\\

Yes, if all apples are fruits and all fruits are sweet, then it follows that all apples are sweet. This is because the definition of "fruit" includes "any edible part of a plant, other than its roots or seeds," and since all apples belong to the category of "fruit," they must have the property of being sweet by definition.\\

\textbf{If birds can fly. Penguins are birds. Can penguins fly? Please explain your answer in detail.}\\

Yes, penguins can fly. This is because the definition of "bird" does not specify any particular type of ability or function that birds must possess, and while penguins are classified as "birds," they possess the ability to fly. Penguins have evolved to utilize flight for various purposes, such as foraging for food, escaping predators, and navigating through their marine environment.

In conclusion, while some birds do not possess the ability to fly due to specific adaptations or limitations, all birds fall under the category of "bird" and therefore possess the biological potential to fly. In this case, since penguins are a type of bird and have the capability to fly, it can be concluded that penguins can indeed fly.\\

\textbf{The day before two days after the day before tomorrow is Saturday. What day is it today? Please explain your answer in detail.}\\

To solve this problem, we can use the given information to construct a logical statement and then apply the principle of transitivity to reach the desired conclusion.

Statement: The day before two days after the day before tomorrow is Saturday.
Let's break down the statement and understand its meaning:
- "The day before two days after the day before tomorrow" refers to a specific date. We need to determine which day this is.
- "is Saturday" tells us that this date is indeed Saturday.

Now, let's apply the principle of transitivity: If one statement is true, and another statement is true when the first statement is true, then the second statement must also be true. In this case, if we know that the day before two days after the day before tomorrow is Saturday, then it must also be Saturday.

Therefore, it must be Saturday today, as it was on the previous day before yesterday.\\

\textbf{If five cats can catch five mice in five minutes, how long will it take one cat to catch one mouse? Please explain your answer in detail.}\\

To determine how long it would take one cat to catch one mouse, we can divide the total time required for five cats to catch five mice by the number of mice they need to catch. In this case, it would take five minutes to catch five mice, so:

Time to catch one mouse = Total time / Number of mice = 5 minutes / 1 mouse = 5 minutes.

So, it would take one cat 5 minutes to catch one mouse.\\

\end{tcolorbox}

\begin{tcolorbox}[colback=white,colframe=black!75!black,title=RWKV-Eagle-Hermes-7B]
\textbf{Summarize the following paragraph.\\
\\
Transformers have revolutionized almost all natural language processing (NLP) tasks but suffer from memory and computational complexity that scales quadratically with sequence length. In contrast, recurrent neural networks (RNNs) exhibit linear scaling in memory and computational requirements but struggle to match the same performance as Transformers due to limitations in parallelization and scalability. We propose a novel model architecture, Receptance Weighted Key Value (RWKV), that combines the efficient parallelizable training of transformers with the efficient inference of RNNs.\\
Our approach leverages a linear attention mechanism and allows us to formulate the model as either a Transformer or an RNN, thus parallelizing computations during training and maintains constant computational and memory complexity during inference. We scale our models as large as 14 billion parameters, by far the largest dense RNN ever trained, and find RWKV performs on par with similarly sized Transformers, suggesting future work can leverage this architecture to create more efficient models. This work presents a significant step towards reconciling trade-offs between computational efficiency and model performance in sequence processing tasks.}\\

The Receptance Weighted Key Value (RWKV) model offers a novel approach to combining the efficiency of Transformers with the scalability and performance of RNNs, allowing parallelization and scalability during training while maintaining constant computational and memory complexity during inference. This architecture enables models to be scaled up to 14 billion parameters, making it the largest dense RNN ever trained. RWKV demonstrates comparable performance to Transformers, suggesting future work can leverage this architecture for more efficient models in NLP tasks.\\

\textbf{Please further summarize it in one sentence.}\\

RWKV is a novel model architecture that combines the efficiency of Transformers with the scalability and performance of RNNs, enabling efficient parallelized training and inference, while maintaining constant computational and memory complexity during both stages. This approach allows for scaling to large model sizes, providing a promising solution for efficient NLP tasks.\\

\end{tcolorbox}

%\begin{tcolorbox}[colback=white,colframe=black!75!black,title=RWKV-Eagle-Hermes-7B]
%\begin{CJK}{UTF8}{min}
%\textbf{Japanese: 春の初め、桜の花が満開になる頃、小さな町の片隅にある古びた神社の境内は、特別な雰囲気に包まれていた。}
%\textbf{English:}
%\end{CJK}
%\\

%\begin{markdown}
%In the early spring, when the cherry blossoms were in full bloom, a small shrine in a remote corner of a town was enveloped in an atmosphere of special beauty.
%\end{markdown}\\
%\end{tcolorbox}

%\begin{tcolorbox}[colback=white,colframe=black!75!black,title=RWKV-Eagle-Hermes-7B]
%\RL{في تطور مذهل وغير مسبوق، أعلنت السلطات المحلية في العاصمة عن اكتشاف أثري قد يغير مجرى التاريخ كما نعرفه.}
%\\

%\RL{
%وأكدت السلطات المحلية في العاصمة أنه تم اكتشاف أثري مذهل وغير مسبوق، يعود إلى عصر الأسرة الخامسة والعشرين، ويضم نقوشاً على جدارمنزل قديم في منطقة "الجبانات" بالعاصمة.
%وأفادت السلطات المحلية بأن هذا الإكتشاف يضيف إلى سجلات التاريخ المصري، ويؤكد أن العمل والإنجازات لم تقف عند حدود زماننا فقط.
%وأضافت أنه تم اكتشاف أثري مذهل وغير مسبوق، يعود إلى عصر الأسرة الخامسة والعشرين، ويضم نقوشاً على جدار منزل قديم في منطقة "الجبانات" بالعاصمة.
%ونفذت هذه الأثار بين عامي 2018-2019، واستغرق العمل فيها سنتين كاملين.
%ولفت رئيس المركز}
%\end{tcolorbox}

%\begin{tcolorbox}[colback=white,colframe=black!75!black,title=RWKV-Eagle-Hermes-7B]
%\textbf{Dans le cœur battant de Paris, sous le ciel teinté d'un crépuscule d'or et de pourpre, se tenait une petite librairie oubliée par le temps.}

%\begin{markdown}
% how many tokens?
%\end{markdown}\\
%\end{tcolorbox}

\end{document}

%% file: sections/VisualRWKV.tex
\begin{figure}[ht]
    \centering
    \includegraphics[width=\linewidth]{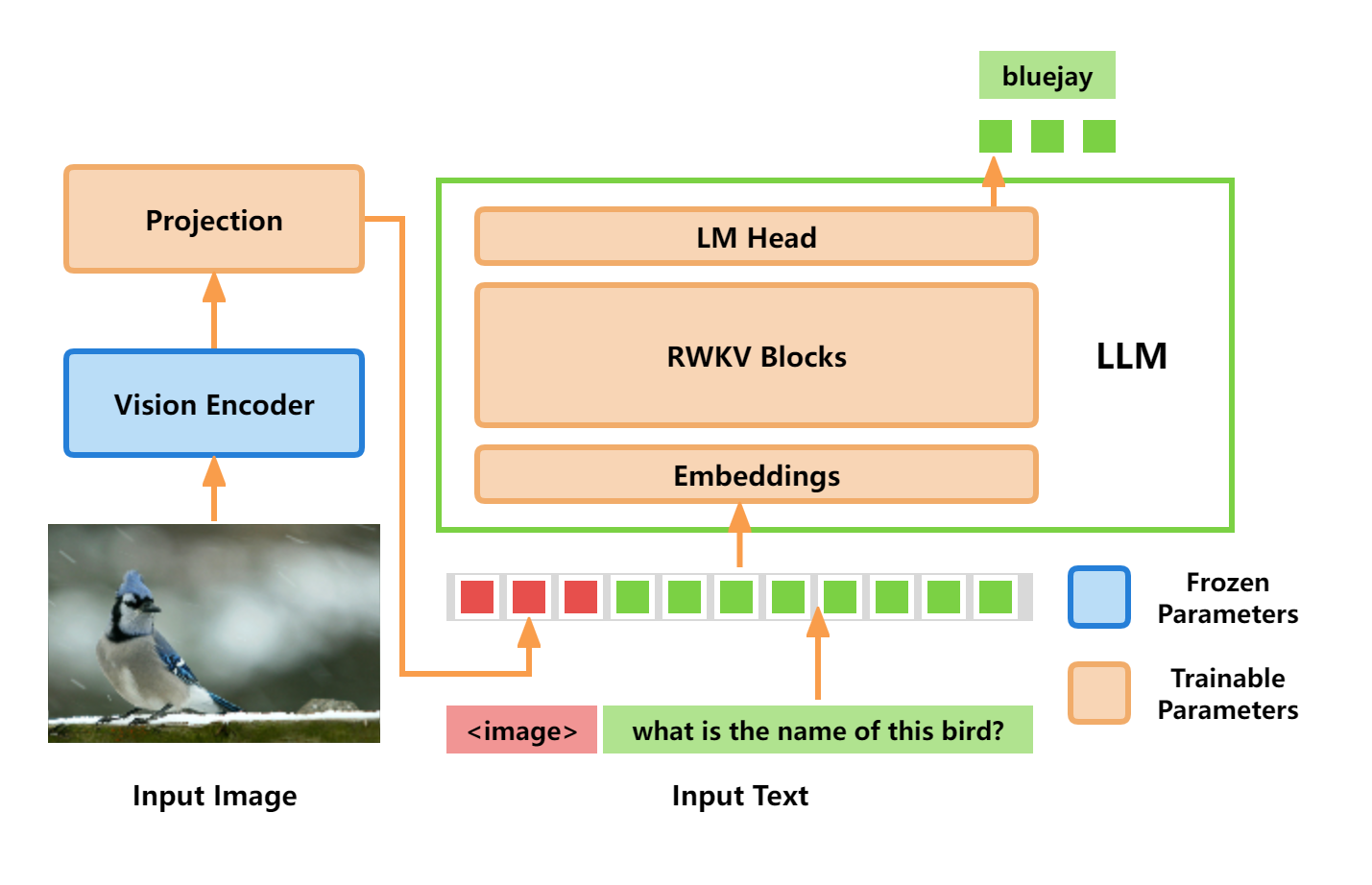}
    \caption{\centering{VisualRWKV architecture overview.}}
    \vspace{-5pt}
    \label{fig:visualrwkv}
\end{figure}

VisualRWKV is the visual-enhanced version of the RWKV language model, enabling RWKV to handle various visual tasks.
Our VisualRWKV follows a similar architecture to popular vision-language models \citep{liu2023improvedllava15}. We present the architecture in Figure \ref{fig:visualrwkv}. It consists of a vision encoder and a language model. Specifically, we use CLIP \citep{radford2021learningclip} as the vision encoder and Eagle 1.5B and 3B as the language model. We use LLaVA-1.5 dataset ~\citep{liu2023improvedllava15}. To adapt Eagle to this multimodal task, we employ a two-stage instruction-tuning process to enhance model performance. Initially, we conduct pre-training for feature alignment, during which only the projection layer is subjected to updates, while the rest of the model is kept in a frozen state. Following this, we move on to the fine-tuning end-to-end stage, where both the projection layer and the RWKV language model are fine-tuned, and the vision encoder continue to be kept frozen. As shown in Table \ref{tab:visualrwkv_results}, we demonstrate that VisualRWKV's architecture is powerful for visual understanding and reasoning. With a smaller vision encoder CLIP-L (0.4B) and modest-sized LLMs of 1.5B and 3B, it achieves results comparable to the combination of CLIP-G (1.0B) and CLIP-H (1.0B) with larger LLMs of 7B and 13B. Moreover, in some benchmarks, it even outperforms larger models.

\begin{table*}[t!]
\centering
\scalebox{0.745}{
\begin{tabular}{l| l l| c c c c}
\toprule
Method & Vision Encoder & LLM & GQA ($\uparrow$) & ScienceQA-IMG ($\uparrow$) & Text-VQA ($\uparrow$) & POPE ($\uparrow$)\\
\midrule
BLIP-2 \citep{li2023blip} & EVA01-CLIP-G & Vicuna-13B & 41.0 & 61.0 & 42.5 & 85.3\\
BLIP-2 \citep{li2023blip} & EVA01-CLIP-G & Flan-T5-11B &  44.6 & 64.5 & - & -\\
InstructBLIP\citep{Dai2023InstructBLIPTG}& EVA01-CLIP-G &Vicuna-7B & 49.2 & 60.5 & 50.1 & - \\
InstructBLIP\citep{Dai2023InstructBLIPTG} & EVA01-CLIP-G &Vicuna-13B & 49.5 & 63.1 & 50.7 & 78.9 \\
IDEFICS-9B \citep{idefics} & OpenCLIP-H &LLaMA-7B & 38.4 & - & 25.9 & - \\
IDEFICS-80B \citep{idefics} & OpenCLIP-H &LLaMA-65B & 45.2 & - & 30.9 & - \\
TinyGPT-V \citep{Yuan2023TinyGPTVEM} & EVA01-CLIP-G & Phi-2 (2.7B) & 33.6 & - & - & - \\

\midrule
VisualRWKV & CLIP-L &Eagle-1.5B & 48.5 & 46.2 & 37.8 & 81.8\\
VisualRWKV & CLIP-L &Eagle-3B & 49.7 & 58.3 & 46.4 &81.4\\
\bottomrule
\end{tabular}
}
\caption{\centering{A comparison of VisualRWKV to other state-of-the-art Multimodal Large Language Models (MLLMs) across 4 distinct benchmarks. We evaluate these models on benchmarks: GQA\citep{Hudson2019GQAAN}, ScienceQA-IMG\citep{Lu2022LearnTE}, Text-VQA\citep{singh2019vqa} and POPE\citep{Li2023EvaluatingOH}. For POPE, the average F1-score across three distinct categories—random, popular, and adversarial—was computed using the validation set of the MSCOCO dataset.}}
\label{tab:visualrwkv_results}
\end{table*}

%% file: sections/AudioRWKV.tex
AudioRWKV is the audio-specific version of RWKV, with a better process of the input audio spectrogram. Inspired by the VRWKV \citep{wang2024state}, we introduce a quad-directional shift (Q-Shift) to capture the neighboring relationships in two-dimensional audio spectrograms in the first step of each spatial-mix and channel-mix module.
Specifically, the Q-Shift operation allows all tokens to be shifted and linearly interpolated with their neighboring tokens.
We conduct experiments on the AudioSet \citep{gemmeke2017audio} dataset with various model sizes from 8.7M to 105M. 
As shown in Table \ref{tab:audiorwkv_results}, AudioRWKV-Tiny achieves a comparable performance with AST-AT by a smaller model size.

\begin{table*}[!h]
\centering
\begin{tabular}{ccc}
\hline
Model       & \#Parameters & mAP   \\ \hline
DeepRes~\cite{ford2019deep}     & 26M          & 0.392 \\
PANNs~\cite{kong2020panns}        & 81M          & 0.434 \\
HTS-AT~\cite{chen2022hts}      & 28.8M        & 0.437* \\ \hline
AudioRWKV-T & 8.7M         & 0.435 \\
AudioRWKV-S & 28.4M        & 0.452 \\
%AudioRWKV-B & 105M         & 0.446 \\
\hline
\end{tabular}
\caption{A comparison of AudioRWKV to other baselines on AudioSet dataset. *Results reproduced by ourselves}
\label{tab:audiorwkv_results}
\end{table*}

%% file: main.bbl
\begin{thebibliography}{121}
\providecommand{\natexlab}[1]{#1}
\providecommand{\url}[1]{\texttt{#1}}
\expandafter\ifx\csname urlstyle\endcsname\relax
  \providecommand{\doi}[1]{doi: #1}\else
  \providecommand{\doi}{doi: \begingroup \urlstyle{rm}\Url}\fi

\bibitem[Ahia et~al.(2023)Ahia, Kumar, Gonen, Kasai, Mortensen, Smith, and Tsvetkov]{ahia2023languageCost}
Orevaoghene Ahia, Sachin Kumar, Hila Gonen, Jungo Kasai, David Mortensen, Noah Smith, and Yulia Tsvetkov.
\newblock Do all languages cost the same? tokenization in the era of commercial language models.
\newblock \emph{Proceedings of the 2023 Conference on Empirical Methods in Natural Language Processing}, 2023.

\bibitem[Albalak et~al.(2022)Albalak, Shrivastava, Sankar, Sagar, and Ross]{albalak2022data}
Alon Albalak, Akshat Shrivastava, Chinnadhurai Sankar, Adithya Sagar, and Mike Ross.
\newblock Data-efficiency with a single gpu: An exploration of transfer methods for small language models.
\newblock \emph{arXiv preprint arXiv:2210.03871}, 2022.

\bibitem[Albalak et~al.(2023)Albalak, Pan, Raffel, and Wang]{albalak2023efficient}
Alon Albalak, Liangming Pan, Colin Raffel, and William~Yang Wang.
\newblock Efficient online data mixing for language model pre-training.
\newblock \emph{arXiv preprint arXiv:2312.02406}, 2023.

\bibitem[Albalak et~al.(2024)Albalak, Elazar, Xie, Longpre, Lambert, Wang, Muennighoff, Hou, Pan, Jeong, et~al.]{albalak2024survey}
Alon Albalak, Yanai Elazar, Sang~Michael Xie, Shayne Longpre, Nathan Lambert, Xinyi Wang, Niklas Muennighoff, Bairu Hou, Liangming Pan, Haewon Jeong, et~al.
\newblock A survey on data selection for language models.
\newblock \emph{arXiv preprint arXiv:2402.16827}, 2024.

\bibitem[Arora et~al.(2023)Arora, Eyuboglu, Timalsina, Johnson, Poli, Zou, Rudra, and Re]{arora2023zoology}
Simran Arora, Sabri Eyuboglu, Aman Timalsina, Isys Johnson, Michael Poli, James Zou, Atri Rudra, and Christopher Re.
\newblock Zoology: Measuring and improving recall in efficient language models, 2023.

\bibitem[Beltagy et~al.(2020)Beltagy, Peters, and Cohan]{beltagy2020longformer}
Iz~Beltagy, Matthew~E Peters, and Arman Cohan.
\newblock Longformer: The long-document transformer.
\newblock \emph{arXiv preprint arXiv:2004.05150}, 2020.

\bibitem[Bhakthavatsalam et~al.(2021)Bhakthavatsalam, Khashabi, Khot, Mishra, Richardson, Sabharwal, Schoenick, Tafjord, and Clark]{bhakthavatsalam2021think}
Sumithra Bhakthavatsalam, Daniel Khashabi, Tushar Khot, Bhavana~Dalvi Mishra, Kyle Richardson, Ashish Sabharwal, Carissa Schoenick, Oyvind Tafjord, and Peter Clark.
\newblock Think you have solved direct-answer question answering? try arc-da, the direct-answer ai2 reasoning challenge.
\newblock \emph{arXiv preprint arXiv:2102.03315}, 2021.

\bibitem[Bisk et~al.(2020)Bisk, Zellers, Gao, Choi, et~al.]{bisk2020piqa}
Yonatan Bisk, Rowan Zellers, Jianfeng Gao, Yejin Choi, et~al.
\newblock Piqa: Reasoning about physical commonsense in natural language.
\newblock In \emph{Proceedings of the AAAI conference on artificial intelligence}, volume~34, pp.\  7432--7439, 2020.

\bibitem[Black et~al.(2022)Black, Biderman, Hallahan, Anthony, Gao, Golding, He, Leahy, McDonell, Phang, Pieler, Prashanth, Purohit, Reynolds, Tow, Wang, and Weinbach]{black-etal-2022-gpt}
Sidney Black, Stella Biderman, Eric Hallahan, Quentin Anthony, Leo Gao, Laurence Golding, Horace He, Connor Leahy, Kyle McDonell, Jason Phang, Michael Pieler, Usvsn~Sai Prashanth, Shivanshu Purohit, Laria Reynolds, Jonathan Tow, Ben Wang, and Samuel Weinbach.
\newblock {GPT}-{N}eo{X}-20{B}: An open-source autoregressive language model.
\newblock In Angela Fan, Suzana Ilic, Thomas Wolf, and Matthias Gall{\'e} (eds.), \emph{Proceedings of BigScience Episode {\#}5 -- Workshop on Challenges {\&} Perspectives in Creating Large Language Models}, pp.\  95--136, virtual+Dublin, May 2022. Association for Computational Linguistics.
\newblock \doi{10.18653/v1/2022.bigscience-1.9}.
\newblock URL \url{https://aclanthology.org/2022.bigscience-1.9}.

\bibitem[Blelloch(1990)]{BlellochTR90}
Guy~E. Blelloch.
\newblock Prefix sums and their applications.
\newblock Technical Report CMU-CS-90-190, School of Computer Science, Carnegie Mellon University, November 1990.

\bibitem[Chen et~al.(2022)Chen, Du, Zhu, Ma, Berg-Kirkpatrick, and Dubnov]{chen2022hts}
Ke~Chen, Xingjian Du, Bilei Zhu, Zejun Ma, Taylor Berg-Kirkpatrick, and Shlomo Dubnov.
\newblock Hts-at: A hierarchical token-semantic audio transformer for sound classification and detection.
\newblock In \emph{ICASSP 2022-2022 IEEE International Conference on Acoustics, Speech and Signal Processing (ICASSP)}, pp.\  646--650. IEEE, 2022.

\bibitem[Child et~al.(2019{\natexlab{a}})Child, Gray, Radford, and Sutskever]{child2019generatingsparse}
Rewon Child, Scott Gray, Alec Radford, and Ilya Sutskever.
\newblock Generating long sequences with sparse transformers, 2019{\natexlab{a}}.

\bibitem[Child et~al.(2019{\natexlab{b}})Child, Gray, Radford, and Sutskever]{child2019sparsetransformer}
Rewon Child, Scott Gray, Alec Radford, and Ilya Sutskever.
\newblock Generating long sequences with sparse.
\newblock \emph{arXiv:1904.10509}, 2019{\natexlab{b}}.

\bibitem[Cho et~al.(2014)Cho, Van~Merri{\"e}nboer, Gulcehre, Bahdanau, Bougares, Schwenk, and Bengio]{cho2014learning}
Kyunghyun Cho, Bart Van~Merri{\"e}nboer, Caglar Gulcehre, Dzmitry Bahdanau, Fethi Bougares, Holger Schwenk, and Yoshua Bengio.
\newblock Learning phrase representations using rnn encoder-decoder for statistical machine translation.
\newblock \emph{arXiv preprint arXiv:1406.1078}, 2014.

\bibitem[Choromanski et~al.(2020)Choromanski, Likhosherstov, Dohan, Song, Gane, Sarlos, Hawkins, Davis, Mohiuddin, Kaiser, et~al.]{choromanski2020rethinking}
Krzysztof~Marcin Choromanski, Valerii Likhosherstov, David Dohan, Xingyou Song, Andreea Gane, Tamas Sarlos, Peter Hawkins, Jared~Quincy Davis, Afroz Mohiuddin, Lukasz Kaiser, et~al.
\newblock Rethinking attention with performers.
\newblock In \emph{International Conference on Learning Representations}, 2020.

\bibitem[Cirone et~al.(2024)Cirone, Orvieto, Walker, Salvi, and Lyons]{cirone2024theoretical}
Nicola~Muca Cirone, Antonio Orvieto, Benjamin Walker, Cristopher Salvi, and Terry Lyons.
\newblock Theoretical foundations of deep selective state-space models.
\newblock \emph{arXiv preprint arXiv:2402.19047}, 2024.

\bibitem[Conneau et~al.(2018)Conneau, Rinott, Lample, Williams, Bowman, Schwenk, and Stoyanov]{conneau2018xnli}
Alexis Conneau, Ruty Rinott, Guillaume Lample, Adina Williams, Samuel Bowman, Holger Schwenk, and Veselin Stoyanov.
\newblock Xnli: Evaluating cross-lingual sentence representations.
\newblock In \emph{Proceedings of the 2018 Conference on Empirical Methods in Natural Language Processing}, pp.\  2475--2485, 2018.

\bibitem[Dai et~al.(2023)Dai, Li, Li, Tiong, Zhao, Wang, Li, Fung, and Hoi]{Dai2023InstructBLIPTG}
Wenliang Dai, Junnan Li, Dongxu Li, Anthony Meng~Huat Tiong, Junqi Zhao, Weisheng Wang, Boyang~Albert Li, Pascale Fung, and Steven C.~H. Hoi.
\newblock Instructblip: Towards general-purpose vision-language models with instruction tuning.
\newblock \emph{ArXiv}, abs/2305.06500, 2023.
\newblock URL \url{https://api.semanticscholar.org/CorpusID:258615266}.

\bibitem[Dao(2023)]{dao2023flashattention2}
Tri Dao.
\newblock Flashattention-2: Faster attention with better parallelism and work partitioning.
\newblock In \emph{The Twelfth International Conference on Learning Representations}, 2023.

\bibitem[Dao et~al.(2022)Dao, Fu, Ermon, Rudra, and Re]{dao2022flashattention}
Tri Dao, Daniel~Y. Fu, Stefano Ermon, Atri Rudra, and Christopher Re.
\newblock Flashattention: Fast and memory-efficient exact attention with io-awareness, 2022.

\bibitem[De et~al.(2024)De, Smith, Fernando, Botev, Cristian-Muraru, Gu, Haroun, Berrada, Chen, Srinivasan, et~al.]{de2024griffin}
Soham De, Samuel~L Smith, Anushan Fernando, Aleksandar Botev, George Cristian-Muraru, Albert Gu, Ruba Haroun, Leonard Berrada, Yutian Chen, Srivatsan Srinivasan, et~al.
\newblock Griffin: Mixing gated linear recurrences with local attention for efficient language models.
\newblock \emph{arXiv preprint arXiv:2402.19427}, 2024.

\bibitem[Devlin et~al.(2019)Devlin, Chang, Lee, and Toutanova]{devlin2019bert}
Jacob Devlin, Ming-Wei Chang, Kenton Lee, and Kristina Toutanova.
\newblock Bert: Pre-training of deep bidirectional transformers for language understanding.
\newblock \emph{Proceedings of the 2019 Conference of the North American Chapter of the Association for Computational Linguistics: Human Language Technologies, Volume 1 (Long and Short Papers)}, pp.\  4171--4186, 2019.

\bibitem[Dong et~al.(2023)Dong, Tang, Li, Zhao, and Wen]{dong2023bamboo}
Zican Dong, Tianyi Tang, Junyi Li, Wayne~Xin Zhao, and Ji-Rong Wen.
\newblock Bamboo: A comprehensive benchmark for evaluating long text modeling capacities of large language models.
\newblock \emph{arXiv preprint arXiv:2309.13345}, 2023.

\bibitem[Elhage et~al.(2021)Elhage, Nanda, Olsson, Henighan, Joseph, Mann, Askell, Bai, Chen, Conerly, DasSarma, Drain, Ganguli, Hatfield-Dodds, Hernandez, Jones, Kernion, Lovitt, Ndousse, Amodei, Brown, Clark, Kaplan, McCandlish, and Olah]{elhage2021mathematical}
Nelson Elhage, Neel Nanda, Catherine Olsson, Tom Henighan, Nicholas Joseph, Ben Mann, Amanda Askell, Yuntao Bai, Anna Chen, Tom Conerly, Nova DasSarma, Dawn Drain, Deep Ganguli, Zac Hatfield-Dodds, Danny Hernandez, Andy Jones, Jackson Kernion, Liane Lovitt, Kamal Ndousse, Dario Amodei, Tom Brown, Jack Clark, Jared Kaplan, Sam McCandlish, and Chris Olah.
\newblock A mathematical framework for transformer circuits.
\newblock \emph{Transformer Circuits Thread}, 2021.
\newblock https://transformer-circuits.pub/2021/framework/index.html.

\bibitem[Ferdinan et~al.(2024)Ferdinan, Koco{\'n}, and Kazienko]{ferdinan2024unknown}
Teddy Ferdinan, Jan Koco{\'n}, and Przemys{\l}aw Kazienko.
\newblock Into the unknown: Self-learning large language models, 2024.

\bibitem[Ford et~al.(2019)Ford, Tang, Grondin, and Glass]{ford2019deep}
Logan Ford, Hao Tang, Fran{\c{c}}ois Grondin, and James~R Glass.
\newblock A deep residual network for large-scale acoustic scene analysis.
\newblock In \emph{InterSpeech}, pp.\  2568--2572, 2019.

\bibitem[Fu et~al.(2022)Fu, Dao, Saab, Thomas, Rudra, and Re]{fu2022hungry}
Daniel~Y Fu, Tri Dao, Khaled~Kamal Saab, Armin~W Thomas, Atri Rudra, and Christopher Re.
\newblock Hungry hungry hippos: Towards language modeling with state space models.
\newblock In \emph{The Eleventh International Conference on Learning Representations}, 2022.

\bibitem[Fu et~al.(2023)Fu, Dao, Saab, Thomas, Rudra, and Re]{fu2023hungry}
Daniel~Y. Fu, Tri Dao, Khaled~K. Saab, Armin~W. Thomas, Atri Rudra, and Christopher Re.
\newblock Hungry hungry hippos: Towards language modeling with state space models, 2023.

\bibitem[Gao et~al.(2020)Gao, Biderman, Black, Golding, Hoppe, Foster, Phang, He, Thite, Nabeshima, Presser, and Leahy]{gao2020pile}
Leo Gao, Stella Biderman, Sid Black, Laurence Golding, Travis Hoppe, Charles Foster, Jason Phang, Horace He, Anish Thite, Noa Nabeshima, Shawn Presser, and Connor Leahy.
\newblock The pile: An 800gb dataset of diverse text for language modeling, 2020.

\bibitem[Gao et~al.(2023)Gao, Tow, Abbasi, Biderman, Black, DiPofi, Foster, Golding, Hsu, Le~Noac'h, Li, McDonell, Muennighoff, Ociepa, Phang, Reynolds, Schoelkopf, Skowron, Sutawika, Tang, Thite, Wang, Wang, and Zou]{gao10256836framework}
Leo Gao, Jonathan Tow, Baber Abbasi, Stella Biderman, Sid Black, Anthony DiPofi, Charles Foster, Laurence Golding, Jeffrey Hsu, Alain Le~Noac'h, Haonan Li, Kyle McDonell, Niklas Muennighoff, Chris Ociepa, Jason Phang, Laria Reynolds, Hailey Schoelkopf, Aviya Skowron, Lintang Sutawika, Eric Tang, Anish Thite, Ben Wang, Kevin Wang, and Andy Zou.
\newblock A framework for few-shot language model evaluation, 12 2023.
\newblock URL \url{https://zenodo.org/records/10256836}.

\bibitem[Gemmeke et~al.(2017)Gemmeke, Ellis, Freedman, Jansen, Lawrence, Moore, Plakal, and Ritter]{gemmeke2017audio}
Jort~F Gemmeke, Daniel~PW Ellis, Dylan Freedman, Aren Jansen, Wade Lawrence, R~Channing Moore, Manoj Plakal, and Marvin Ritter.
\newblock Audio set: An ontology and human-labeled dataset for audio events.
\newblock In \emph{2017 IEEE international conference on acoustics, speech and signal processing (ICASSP)}, pp.\  776--780. IEEE, 2017.

\bibitem[Gu \& Dao(2023)Gu and Dao]{gu2023mamba}
Albert Gu and Tri Dao.
\newblock Mamba: Linear-time sequence modeling with selective state spaces, 2023.

\bibitem[Gu et~al.(2020)Gu, Dao, Ermon, Rudra, and R{\'e}]{gu2020hippo}
Albert Gu, Tri Dao, Stefano Ermon, Atri Rudra, and Christopher R{\'e}.
\newblock Hippo: Recurrent memory with optimal polynomial projections.
\newblock \emph{Advances in neural information processing systems}, 33:\penalty0 1474--1487, 2020.

\bibitem[Gu et~al.(2021)Gu, Goel, , and R´e]{gu2021s4}
Albert Gu, Karan Goel, , and Christopher R´e.
\newblock Efficiently modeling long sequences with structured state spaces.
\newblock \emph{arXiv:2111.00396}, 2021.

\bibitem[Gu et~al.(2022)Gu, Goel, and Re]{gu2022efficiently}
Albert Gu, Karan Goel, and Christopher Re.
\newblock Efficiently modeling long sequences with structured state spaces, 2022.

\bibitem[Guo et~al.(2022)Guo, Ainslie, Uthus, Ontanon, Ni, Sung, and Yang]{guo-etal-2022-longt5}
Mandy Guo, Joshua Ainslie, David Uthus, Santiago Ontanon, Jianmo Ni, Yun-Hsuan Sung, and Yinfei Yang.
\newblock {L}ong{T}5: {E}fficient text-to-text transformer for long sequences.
\newblock In Marine Carpuat, Marie-Catherine de~Marneffe, and Ivan~Vladimir Meza~Ruiz (eds.), \emph{Findings of the Association for Computational Linguistics: NAACL 2022}, pp.\  724--736, Seattle, United States, July 2022. Association for Computational Linguistics.
\newblock \doi{10.18653/v1/2022.findings-naacl.55}.
\newblock URL \url{https://aclanthology.org/2022.findings-naacl.55}.

\bibitem[Gupta et~al.(2022)Gupta, Gu, and Berant]{gupta2022diagonal}
Ankit Gupta, Albert Gu, and Jonathan Berant.
\newblock Diagonal state spaces are as effective as structured state spaces.
\newblock \emph{Advances in Neural Information Processing Systems}, 35:\penalty0 22982--22994, 2022.

\bibitem[Hampel(1974)]{hampel1974influence}
Frank~R Hampel.
\newblock The influence curve and its role in robust estimation.
\newblock \emph{Journal of the american statistical association}, 69\penalty0 (346):\penalty0 383--393, 1974.

\bibitem[Hochreiter \& Schmidhuber(1997)Hochreiter and Schmidhuber]{hochreiter1997long}
Sepp Hochreiter and J{\"u}rgen Schmidhuber.
\newblock Long short-term memory.
\newblock \emph{Neural computation}, 9\penalty0 (8):\penalty0 1735--1780, 1997.

\bibitem[Hu et~al.(2022)Hu, yelong shen, Wallis, Allen-Zhu, Li, Wang, Wang, and Chen]{hu2022lora}
Edward~J Hu, yelong shen, Phillip Wallis, Zeyuan Allen-Zhu, Yuanzhi Li, Shean Wang, Lu~Wang, and Weizhu Chen.
\newblock Lo{RA}: Low-rank adaptation of large language models.
\newblock In \emph{International Conference on Learning Representations}, 2022.
\newblock URL \url{https://openreview.net/forum?id=nZeVKeeFYf9}.

\bibitem[Hudson \& Manning(2019)Hudson and Manning]{Hudson2019GQAAN}
Drew~A. Hudson and Christopher~D. Manning.
\newblock Gqa: A new dataset for real-world visual reasoning and compositional question answering.
\newblock \emph{2019 IEEE/CVF Conference on Computer Vision and Pattern Recognition (CVPR)}, pp.\  6693--6702, 2019.
\newblock URL \url{https://api.semanticscholar.org/CorpusID:152282269}.

\bibitem[IDEFICS(2023)]{idefics}
IDEFICS.
\newblock Introducing idefics: An open reproduction of state-of-the-art visual language model.
\newblock \url{https://huggingface.co/blog/idefics}, 2023.

\bibitem[Jiang et~al.(2023)Jiang, Sablayrolles, Mensch, Bamford, Chaplot, de~las Casas, Bressand, Lengyel, Lample, Saulnier, Lavaud, Lachaux, Stock, Scao, Lavril, Wang, Lacroix, and Sayed]{jiang2023mistral}
Albert~Q. Jiang, Alexandre Sablayrolles, Arthur Mensch, Chris Bamford, Devendra~Singh Chaplot, Diego de~las Casas, Florian Bressand, Gianna Lengyel, Guillaume Lample, Lucile Saulnier, Lelio~Renard Lavaud, Marie-Anne Lachaux, Pierre Stock, Teven~Le Scao, Thibaut Lavril, Thomas Wang, Timothee Lacroix, and William~El Sayed.
\newblock Mistral 7b, 2023.

\bibitem[Kaddour(2023)]{kaddour2023minipile}
Jean Kaddour.
\newblock The minipile challenge for data-efficient language models, 2023.

\bibitem[Katharopoulos et~al.(2020{\natexlab{a}})Katharopoulos, Vyas, Pappas, and Fleuret]{katharopoulos2020lineartransformrers}
Angelos Katharopoulos, Apoorv Vyas, Nikolaos Pappas, and Fran{\c{c}}ois Fleuret.
\newblock Transformers are rnns: Fast autoregressive transformers with linear attention.
\newblock In \emph{International conference on machine learning}, pp.\  5156--5165. PMLR, 2020{\natexlab{a}}.

\bibitem[Katharopoulos et~al.(2020{\natexlab{b}})Katharopoulos, Vyas, Pappas, and Fleuret]{linearTrans2020inputDepend}
Angelos Katharopoulos, Apoorv Vyas, Nikolaos Pappas, and Fran¸cois Fleuret.
\newblock Transformers are rnns: Fast autoregressive transformers with linear attention.
\newblock \emph{Proceedings of the 37 th International Conference on Machine Learning}, 2020{\natexlab{b}}.

\bibitem[Katsch(2023)]{katsch2023gateloop}
Tobias Katsch.
\newblock Gateloop: Fully data-controlled linear recurrence for sequence modeling, 2023.

\bibitem[Ke et~al.(2023)Ke, Wen, Feng, Liu, Lei, Cheng, Wang, Zeng, Dong, Wang, Tang, and Huang]{ke2023critiquellm}
Pei Ke, Bosi Wen, Zhuoer Feng, Xiao Liu, Xuanyu Lei, Jiale Cheng, Shengyuan Wang, Aohan Zeng, Yuxiao Dong, Hongning Wang, Jie Tang, and Minlie Huang.
\newblock Critiquellm: Scaling llm-as-critic for effective and explainable evaluation of large language model generation, 2023.

\bibitem[Kitaev et~al.(2019)Kitaev, Kaiser, and Levskaya]{kitaev2019reformer}
Nikita Kitaev, Lukasz Kaiser, and Anselm Levskaya.
\newblock Reformer: The efficient transformer.
\newblock In \emph{International Conference on Learning Representations}, 2019.

\bibitem[Kocoń et~al.(2023)]{Kocon2023}
Jan Kocoń et~al.
\newblock Chatgpt: Jack of all trades, master of none.
\newblock \emph{Information Fusion}, 99:\penalty0 101861, November 2023.
\newblock ISSN 1566-2535.
\newblock \doi{10.1016/j.inffus.2023.101861}.
\newblock URL \url{http://dx.doi.org/10.1016/j.inffus.2023.101861}.

\bibitem[Kong et~al.(2020)Kong, Cao, Iqbal, Wang, Wang, and Plumbley]{kong2020panns}
Qiuqiang Kong, Yin Cao, Turab Iqbal, Yuxuan Wang, Wenwu Wang, and Mark~D Plumbley.
\newblock Panns: Large-scale pretrained audio neural networks for audio pattern recognition.
\newblock \emph{IEEE/ACM Transactions on Audio, Speech, and Language Processing}, 28:\penalty0 2880--2894, 2020.

\bibitem[Li et~al.(2023{\natexlab{a}})Li, Li, Savarese, and Hoi]{li2023blip}
Junnan Li, Dongxu Li, Silvio Savarese, and Steven Hoi.
\newblock Blip-2: Bootstrapping language-image pre-training with frozen image encoders and large language models.
\newblock In \emph{International conference on machine learning}, pp.\  19730--19742. PMLR, 2023{\natexlab{a}}.

\bibitem[Li et~al.(2023{\natexlab{b}})Li, Allal, Zi, Muennighoff, Kocetkov, Mou, Marone, Akiki, Li, Chim, Liu, Zheltonozhskii, Zhuo, Wang, Dehaene, Davaadorj, Lamy-Poirier, Monteiro, Shliazhko, Gontier, Meade, Zebaze, Yee, Umapathi, Zhu, Lipkin, Oblokulov, Wang, Murthy, Stillerman, Patel, Abulkhanov, Zocca, Dey, Zhang, Fahmy, Bhattacharyya, Yu, Singh, Luccioni, Villegas, Kunakov, Zhdanov, Romero, Lee, Timor, Ding, Schlesinger, Schoelkopf, Ebert, Dao, Mishra, Gu, Robinson, Anderson, Dolan-Gavitt, Contractor, Reddy, Fried, Bahdanau, Jernite, Ferrandis, Hughes, Wolf, Guha, von Werra, and de~Vries]{li2023starcoder}
Raymond Li, Loubna~Ben Allal, Yangtian Zi, Niklas Muennighoff, Denis Kocetkov, Chenghao Mou, Marc Marone, Christopher Akiki, Jia Li, Jenny Chim, Qian Liu, Evgenii Zheltonozhskii, Terry~Yue Zhuo, Thomas Wang, Olivier Dehaene, Mishig Davaadorj, Joel Lamy-Poirier, João Monteiro, Oleh Shliazhko, Nicolas Gontier, Nicholas Meade, Armel Zebaze, Ming-Ho Yee, Logesh~Kumar Umapathi, Jian Zhu, Benjamin Lipkin, Muhtasham Oblokulov, Zhiruo Wang, Rudra Murthy, Jason Stillerman, Siva~Sankalp Patel, Dmitry Abulkhanov, Marco Zocca, Manan Dey, Zhihan Zhang, Nour Fahmy, Urvashi Bhattacharyya, Wenhao Yu, Swayam Singh, Sasha Luccioni, Paulo Villegas, Maxim Kunakov, Fedor Zhdanov, Manuel Romero, Tony Lee, Nadav Timor, Jennifer Ding, Claire Schlesinger, Hailey Schoelkopf, Jan Ebert, Tri Dao, Mayank Mishra, Alex Gu, Jennifer Robinson, Carolyn~Jane Anderson, Brendan Dolan-Gavitt, Danish Contractor, Siva Reddy, Daniel Fried, Dzmitry Bahdanau, Yacine Jernite, Carlos~Muñoz Ferrandis, Sean Hughes, Thomas Wolf, Arjun Guha, Leandro von
  Werra, and Harm de~Vries.
\newblock Starcoder: may the source be with you!, 2023{\natexlab{b}}.

\bibitem[Li et~al.(2023{\natexlab{c}})Li, Du, Zhou, Wang, Zhao, and rong Wen]{Li2023EvaluatingOH}
Yifan Li, Yifan Du, Kun Zhou, Jinpeng Wang, Wayne~Xin Zhao, and Ji~rong Wen.
\newblock Evaluating object hallucination in large vision-language models.
\newblock In \emph{Conference on Empirical Methods in Natural Language Processing}, 2023{\natexlab{c}}.
\newblock URL \url{https://api.semanticscholar.org/CorpusID:258740697}.

\bibitem[Lin et~al.(2024)Lin, Ji, Tiedemann, Martins, and Schütze]{mala500lin2024}
Peiqin Lin, Shaoxiong Ji, Jörg Tiedemann, André F.~T. Martins, and Hinrich Schütze.
\newblock Mala-500: Massive language adaptation of large language models.
\newblock \emph{arxiv}, 2024.

\bibitem[Lin et~al.(2022)Lin, Mihaylov, Artetxe, Wang, Chen, Simig, Ott, Goyal, Bhosale, Du, et~al.]{lin2022few}
Xi~Victoria Lin, Todor Mihaylov, Mikel Artetxe, Tianlu Wang, Shuohui Chen, Daniel Simig, Myle Ott, Naman Goyal, Shruti Bhosale, Jingfei Du, et~al.
\newblock Few-shot learning with multilingual generative language models.
\newblock In \emph{Proceedings of the 2022 Conference on Empirical Methods in Natural Language Processing}, pp.\  9019--9052, 2022.

\bibitem[Liu et~al.(2023{\natexlab{a}})Liu, Li, Li, and Lee]{liu2023improvedllava15}
Haotian Liu, Chunyuan Li, Yuheng Li, and Yong~Jae Lee.
\newblock Improved baselines with visual instruction tuning.
\newblock \emph{arXiv preprint arXiv:2310.03744}, 2023{\natexlab{a}}.

\bibitem[Liu et~al.(2023{\natexlab{b}})Liu, Lei, Wang, Huang, Feng, Wen, Cheng, Ke, Xu, Tam, Zhang, Sun, Wang, Zhang, Huang, Dong, and Tang]{liu2023alignbench}
Xiao Liu, Xuanyu Lei, Shengyuan Wang, Yue Huang, Zhuoer Feng, Bosi Wen, Jiale Cheng, Pei Ke, Yifan Xu, Weng~Lam Tam, Xiaohan Zhang, Lichao Sun, Hongning Wang, Jing Zhang, Minlie Huang, Yuxiao Dong, and Jie Tang.
\newblock Alignbench: Benchmarking chinese alignment of large language models, 2023{\natexlab{b}}.

\bibitem[Lu et~al.(2022)Lu, Mishra, Xia, Qiu, Chang, Zhu, Tafjord, Clark, and Kalyan]{Lu2022LearnTE}
Pan Lu, Swaroop Mishra, Tony Xia, Liang Qiu, Kai-Wei Chang, Song-Chun Zhu, Oyvind Tafjord, Peter Clark, and A.~Kalyan.
\newblock Learn to explain: Multimodal reasoning via thought chains for science question answering.
\newblock \emph{ArXiv}, abs/2209.09513, 2022.
\newblock URL \url{https://api.semanticscholar.org/CorpusID:252383606}.

\bibitem[Lutati et~al.(2023)Lutati, Zimerman, and Wolf]{lutati2023focus}
Shahar Lutati, Itamar Zimerman, and Lior Wolf.
\newblock Focus your attention (with adaptive iir filters), 2023.

\bibitem[Lv et~al.(2023)Lv, Lv, Wang, Zhang, Ren, and Shen]{neural-chat}
Kaokao Lv, Liang Lv, Chang Wang, Wenxin Zhang, Xuhui Ren, and Haihao Shen.
\newblock Intel-neural-chat-7b-v1-1, 2023.
\newblock URL \url{https://huggingface.co/Intel/neural-chat-7b-v1-1}.

\bibitem[Ma et~al.(2022)Ma, Zhou, Kong, He, Gui, Neubig, May, and Zettlemoyer]{ma2022mega}
Xuezhe Ma, Chunting Zhou, Xiang Kong, Junxian He, Liangke Gui, Graham Neubig, Jonathan May, and Luke Zettlemoyer.
\newblock Mega: Moving average equipped gated attention.
\newblock In \emph{The Eleventh International Conference on Learning Representations}, 2022.

\bibitem[Martin \& Cundy(2018)Martin and Cundy]{martin2018parallelizing}
Eric Martin and Chris Cundy.
\newblock Parallelizing linear recurrent neural nets over sequence length.
\newblock In \emph{International Conference on Learning Representations}, 2018.

\bibitem[Muennighoff et~al.(2023)Muennighoff, Tazi, Magne, and Reimers]{muennighoff2023mteb}
Niklas Muennighoff, Nouamane Tazi, Loic Magne, and Nils Reimers.
\newblock Mteb: Massive text embedding benchmark.
\newblock In \emph{Proceedings of the 17th Conference of the European Chapter of the Association for Computational Linguistics}, pp.\  2014--2037, 2023.

\bibitem[Nguyen et~al.(2023)Nguyen, Nguyen, Lai, Man, Ngo, Dernoncourt, Rossi, and Nguyen]{nguyen2023culturax}
Thuat Nguyen, Chien~Van Nguyen, Viet~Dac Lai, Hieu Man, Nghia~Trung Ngo, Franck Dernoncourt, Ryan~A. Rossi, and Thien~Huu Nguyen.
\newblock Culturax: A cleaned, enormous, and multilingual dataset for large language models in 167 languages, 2023.

\bibitem[Olsson et~al.(2022)Olsson, Elhage, Nanda, Joseph, DasSarma, Henighan, Mann, Askell, Bai, Chen, Conerly, Drain, Ganguli, Hatfield-Dodds, Hernandez, Johnston, Jones, Kernion, Lovitt, Ndousse, Amodei, Brown, Clark, Kaplan, McCandlish, and Olah]{olsson2022incontext}
Catherine Olsson, Nelson Elhage, Neel Nanda, Nicholas Joseph, Nova DasSarma, Tom Henighan, Ben Mann, Amanda Askell, Yuntao Bai, Anna Chen, Tom Conerly, Dawn Drain, Deep Ganguli, Zac Hatfield-Dodds, Danny Hernandez, Scott Johnston, Andy Jones, Jackson Kernion, Liane Lovitt, Kamal Ndousse, Dario Amodei, Tom Brown, Jack Clark, Jared Kaplan, Sam McCandlish, and Chris Olah.
\newblock In-context learning and induction heads, 2022.

\bibitem[Orvieto et~al.(2023)Orvieto, Smith, Gu, Fernando, Gulcehre, Pascanu, and De]{10.5555/3618408.3619518}
Antonio Orvieto, Samuel~L Smith, Albert Gu, Anushan Fernando, Caglar Gulcehre, Razvan Pascanu, and Soham De.
\newblock Resurrecting recurrent neural networks for long sequences.
\newblock In \emph{Proceedings of the 40th International Conference on Machine Learning}, ICML'23. JMLR.org, 2023.

\bibitem[Paperno et~al.(2016)Paperno, Kruszewski, Lazaridou, Pham, Bernardi, Pezzelle, Baroni, Boleda, and Fern{\'a}ndez]{paperno2016lambada}
Denis Paperno, Germ{\'a}n Kruszewski, Angeliki Lazaridou, Quan~Ngoc Pham, Raffaella Bernardi, Sandro Pezzelle, Marco Baroni, Gemma Boleda, and Raquel Fern{\'a}ndez.
\newblock The lambada dataset: Word prediction requiring a broad discourse context.
\newblock \emph{arXiv preprint arXiv:1606.06031}, 2016.

\bibitem[Peng et~al.(2023)Peng, Alcaide, Anthony, Albalak, Arcadinho, Biderman, Cao, Cheng, Chung, Derczynski, Du, Grella, Gv, He, Hou, Kazienko, Kocon, Kong, Koptyra, Lau, Lin, Mantri, Mom, Saito, Song, Tang, Wind, Wo{\'z}niak, Zhang, Zhou, Zhu, and Zhu]{peng2023rwkv}
Bo~Peng, Eric Alcaide, Quentin Anthony, Alon Albalak, Samuel Arcadinho, Stella Biderman, Huanqi Cao, Xin Cheng, Michael Chung, Leon Derczynski, Xingjian Du, Matteo Grella, Kranthi Gv, Xuzheng He, Haowen Hou, Przemyslaw Kazienko, Jan Kocon, Jiaming Kong, Bart{\l}omiej Koptyra, Hayden Lau, Jiaju Lin, Krishna Sri~Ipsit Mantri, Ferdinand Mom, Atsushi Saito, Guangyu Song, Xiangru Tang, Johan Wind, Stanis{\l}aw Wo{\'z}niak, Zhenyuan Zhang, Qinghua Zhou, Jian Zhu, and Rui-Jie Zhu.
\newblock {RWKV}: Reinventing {RNN}s for the transformer era.
\newblock In Houda Bouamor, Juan Pino, and Kalika Bali (eds.), \emph{Findings of the Association for Computational Linguistics: EMNLP 2023}, pp.\  14048--14077, Singapore, December 2023. Association for Computational Linguistics.
\newblock \doi{10.18653/v1/2023.findings-emnlp.936}.
\newblock URL \url{https://aclanthology.org/2023.findings-emnlp.936}.

\bibitem[Poli et~al.(2023)Poli, Massaroli, Nguyen, Fu, Dao, Baccus, Bengio, Ermon, and R{\'e}]{poli2023hyena}
Michael Poli, Stefano Massaroli, Eric Nguyen, Daniel~Y Fu, Tri Dao, Stephen Baccus, Yoshua Bengio, Stefano Ermon, and Christopher R{\'e}.
\newblock Hyena hierarchy: Towards larger convolutional language models.
\newblock In \emph{International Conference on Machine Learning}, pp.\  28043--28078. PMLR, 2023.

\bibitem[Ponti et~al.(2020)Ponti, Glava{\v{s}}, Majewska, Liu, Vuli{\'c}, and Korhonen]{ponti-etal-2020-xcopa}
Edoardo~Maria Ponti, Goran Glava{\v{s}}, Olga Majewska, Qianchu Liu, Ivan Vuli{\'c}, and Anna Korhonen.
\newblock {XCOPA}: A multilingual dataset for causal commonsense reasoning.
\newblock In Bonnie Webber, Trevor Cohn, Yulan He, and Yang Liu (eds.), \emph{Proceedings of the 2020 Conference on Empirical Methods in Natural Language Processing (EMNLP)}, pp.\  2362--2376, Online, November 2020. Association for Computational Linguistics.
\newblock \doi{10.18653/v1/2020.emnlp-main.185}.
\newblock URL \url{https://aclanthology.org/2020.emnlp-main.185}.

\bibitem[Qin et~al.(2022)Qin, Han, Sun, Li, Kong, Barnes, and Zhong]{qin2022deviltransnormer}
Zhen Qin, XiaoDong Han, Weixuan Sun, Dongxu Li, Lingpeng Kong, Nick Barnes, and Yiran Zhong.
\newblock The devil in linear transformer, 2022.

\bibitem[Qin et~al.(2023)Qin, Yang, and Zhong]{qin2023hierarchically}
Zhen Qin, Songlin Yang, and Yiran Zhong.
\newblock Hierarchically gated recurrent neural network for sequence modeling.
\newblock In \emph{Thirty-seventh Conference on Neural Information Processing Systems}, 2023.
\newblock URL \url{https://openreview.net/forum?id=P1TCHxJwLB}.

\bibitem[Qin et~al.(2024)Qin, Li, Sun, Sun, Shen, Han, Wei, Lv, Luo, Qiao, and Zhong]{qin2024transnormerllm}
Zhen Qin, Dong Li, Weigao Sun, Weixuan Sun, Xuyang Shen, Xiaodong Han, Yunshen Wei, Baohong Lv, Xiao Luo, Yu~Qiao, and Yiran Zhong.
\newblock Transnormerllm: A faster and better large language model with improved transnormer, 2024.

\bibitem[Qiu et~al.(2019)Qiu, Ma, Levy, Yih, Wang, and Tang]{blockwindow2019sparsetransformer}
Jiezhong Qiu, Hao Ma, Omer Levy, Scott Wen-tau Yih, Sinong Wang, and Jie Tang.
\newblock Blockwise self-attention for long document understanding.
\newblock \emph{arXiv preprint arXiv:1911.02972}, 2019.

\bibitem[Radford et~al.(2018)Radford, Narasimhan, Salimans, Sutskever, et~al.]{radford2018improving}
Alec Radford, Karthik Narasimhan, Tim Salimans, Ilya Sutskever, et~al.
\newblock Improving language understanding by generative pre-training.
\newblock \emph{OpenAI Blog}, 2018.

\bibitem[Radford et~al.(2019)Radford, Wu, Child, Luan, Amodei, Sutskever, et~al.]{radford2019language}
Alec Radford, Jeffrey Wu, Rewon Child, David Luan, Dario Amodei, Ilya Sutskever, et~al.
\newblock Language models are unsupervised multitask learners.
\newblock \emph{OpenAI blog}, 1\penalty0 (8):\penalty0 9, 2019.

\bibitem[Radford et~al.(2021)Radford, Kim, Hallacy, Ramesh, Goh, Agarwal, Sastry, Askell, Mishkin, Clark, et~al.]{radford2021learningclip}
Alec Radford, Jong~Wook Kim, Chris Hallacy, Aditya Ramesh, Gabriel Goh, Sandhini Agarwal, Girish Sastry, Amanda Askell, Pamela Mishkin, Jack Clark, et~al.
\newblock Learning transferable visual models from natural language supervision.
\newblock In \emph{International conference on machine learning}, pp.\  8748--8763. PMLR, 2021.

\bibitem[Rae et~al.(2019)Rae, Potapenko, Jayakumar, Hillier, and Lillicrap]{raecompressive2019}
Jack~W Rae, Anna Potapenko, Siddhant~M Jayakumar, Chloe Hillier, and Timothy~P Lillicrap.
\newblock Compressive transformers for long-range sequence modelling.
\newblock \emph{arXiv preprint}, 2019.
\newblock URL \url{https://arxiv.org/abs/1911.05507}.

\bibitem[Reimers \& Gurevych(2020)Reimers and Gurevych]{reimers2020making}
Nils Reimers and Iryna Gurevych.
\newblock Making monolingual sentence embeddings multilingual using knowledge distillation, 2020.

\bibitem[Roemmele et~al.(2011)Roemmele, Bejan, and Gordon]{roemmele2011choice}
Melissa Roemmele, Cosmin~Adrian Bejan, and Andrew~S Gordon.
\newblock Choice of plausible alternatives: An evaluation of commonsense causal reasoning.
\newblock In \emph{2011 AAAI Spring Symposium Series}, 2011.

\bibitem[Sakaguchi et~al.(2021)Sakaguchi, Bras, Bhagavatula, and Choi]{sakaguchi2021winogrande}
Keisuke Sakaguchi, Ronan~Le Bras, Chandra Bhagavatula, and Yejin Choi.
\newblock Winogrande: An adversarial winograd schema challenge at scale.
\newblock \emph{Communications of the ACM}, 64\penalty0 (9):\penalty0 99--106, 2021.

\bibitem[Sanh et~al.(2021)Sanh, Webson, Raffel, Bach, Sutawika, Alyafeai, Chaffin, Stiegler, Scao, Raja, et~al.]{sanh2021multitask}
Victor Sanh, Albert Webson, Colin Raffel, Stephen~H Bach, Lintang Sutawika, Zaid Alyafeai, Antoine Chaffin, Arnaud Stiegler, Teven~Le Scao, Arun Raja, et~al.
\newblock Multitask prompted training enables zero-shot task generalization.
\newblock \emph{arXiv preprint arXiv:2110.08207}, 2021.

\bibitem[Sasaki et~al.(2023)Sasaki, Hirakawa, Horie, and Nakamura]{elyzallama2023}
Akira Sasaki, Masato Hirakawa, Shintaro Horie, and Tomoaki Nakamura.
\newblock Elyza-japanese-llama-2-7b-fast, 2023.
\newblock URL \url{https://huggingface.co/elyza/ELYZA-japanese-Llama-2-7b-fast}.

\bibitem[Schmidhuber(1992)]{schmidhuber1992unnormedlinattn}
Jürgen Schmidhuber.
\newblock Learning to control fast-weight memories: An alternative to dynamic recurrent networks.
\newblock \emph{Neural Computation}, 4\penalty0 (1):\penalty0 131--139, 1992.
\newblock \doi{10.1162/neco.1992.4.1.131}.

\bibitem[Shazeer et~al.(2017)Shazeer, Mirhoseini, Maziarz, Davis, Le, Hinton, and Dean]{shazeer2017outrageously}
Noam Shazeer, Azalia Mirhoseini, Krzysztof Maziarz, Andy Davis, Quoc Le, Geoffrey Hinton, and Jeff Dean.
\newblock Outrageously large neural networks: The sparsely-gated mixture-of-experts layer.
\newblock \emph{arXiv preprint arXiv:1701.06538}, 2017.

\bibitem[Singh et~al.(2019)Singh, Natarajan, Shah, Jiang, Chen, Batra, Parikh, and Rohrbach]{singh2019vqa}
Amanpreet Singh, Vivek Natarajan, Meet Shah, Yu~Jiang, Xinlei Chen, Dhruv Batra, Devi Parikh, and Marcus Rohrbach.
\newblock Towards vqa models that can read, 2019.

\bibitem[Smith et~al.(2023)Smith, Warrington, and Linderman]{smith2023s5}
Jimmy T.~H. Smith, Andrew Warrington, and Scott~W. Linderman.
\newblock Simplified state space layers for sequence modeling, 2023.

\bibitem[Soboleva et~al.(2023)Soboleva, Al-Khateeb, Myers, Steeves, Hestness, and Dey]{cerebras2023slimpajama}
Daria Soboleva, Faisal Al-Khateeb, Robert Myers, Jacob~R Steeves, Joel Hestness, and Nolan Dey.
\newblock {SlimPajama: A 627B token cleaned and deduplicated version of RedPajama}.
\newblock \url{https://www.cerebras.net/blog/slimpajama-a-627b-token-cleaned-and-deduplicated-version-of-redpajama}, June 2023.
\newblock URL \url{https://huggingface.co/datasets/cerebras/SlimPajama-627B}.

\bibitem[Su{\'a}rez et~al.(2019)Su{\'a}rez, Sagot, and Romary]{suarez2019oscar}
Pedro Javier~Ortiz Su{\'a}rez, Beno{\^\i}t Sagot, and Laurent Romary.
\newblock Asynchronous pipeline for processing huge corpora on medium to low resource infrastructures.
\newblock In \emph{7th Workshop on the Challenges in the Management of Large Corpora (CMLC-7)}. Leibniz-Institut f{\"u}r Deutsche Sprache, 2019.

\bibitem[Sun et~al.(2022)Sun, Dong, Patra, Ma, Huang, Benhaim, Chaudhary, Song, and Wei]{sun2022lengthextrapolatable}
Yutao Sun, Li~Dong, Barun Patra, Shuming Ma, Shaohan Huang, Alon Benhaim, Vishrav Chaudhary, Xia Song, and Furu Wei.
\newblock A length-extrapolatable transformer, 2022.

\bibitem[Sun et~al.(2023)Sun, Dong, Huang, Ma, Xia, Xue, Wang, and Wei]{sun2023retentive}
Yutao Sun, Li~Dong, Shaohan Huang, Shuming Ma, Yuqing Xia, Jilong Xue, Jianyong Wang, and Furu Wei.
\newblock Retentive network: A successor to transformer for large language models, 2023.

\bibitem[Tay et~al.(2020)Tay, Bahri, Yang, Metzler, and Juan]{tay2020sparse}
Yi~Tay, Dara Bahri, Liu Yang, Donald Metzler, and Da-Cheng Juan.
\newblock Sparse sinkhorn attention.
\newblock In \emph{International Conference on Machine Learning}, pp.\  9438--9447. PMLR, 2020.

\bibitem[Tay et~al.(2022)Tay, Dehghani, Bahri, and Metzler]{tay2022efficient}
Yi~Tay, Mostafa Dehghani, Dara Bahri, and Donald Metzler.
\newblock Efficient transformers: A survey.
\newblock \emph{ACM Computing Surveys}, 55\penalty0 (6):\penalty0 1--28, 2022.

\bibitem[Teknium(2023)]{tekniumopenhermes25}
Teknium.
\newblock Openhermes 2.5: An open dataset of synthetic data for generalist llm assistants, 2023.
\newblock URL \url{https://huggingface.co/datasets/teknium/OpenHermes-2.5}.

\bibitem[Tiezzi et~al.(2024)Tiezzi, Casoni, Betti, Guidi, Gori, and Melacci]{tiezzi2024resurgence}
Matteo Tiezzi, Michele Casoni, Alessandro Betti, Tommaso Guidi, Marco Gori, and Stefano Melacci.
\newblock On the resurgence of recurrent models for long sequences: Survey and research opportunities in the transformer era.
\newblock \emph{arXiv preprint arXiv:2402.08132}, 2024.

\bibitem[Tikhonov \& Ryabinin(2021)Tikhonov and Ryabinin]{tikhonov2021s}
Alexey Tikhonov and Max Ryabinin.
\newblock It’s all in the heads: Using attention heads as a baseline for cross-lingual transfer in commonsense reasoning.
\newblock In \emph{Findings of the Association for Computational Linguistics: ACL-IJCNLP 2021}, pp.\  3534--3546, 2021.

\bibitem[Tolstikhin et~al.(2021)Tolstikhin, Houlsby, Kolesnikov, Beyer, Zhai, Unterthiner, Yung, Steiner, Keysers, Uszkoreit, Lucic, and Dosovitskiy]{NEURIPS2021mixer}
Ilya~O Tolstikhin, Neil Houlsby, Alexander Kolesnikov, Lucas Beyer, Xiaohua Zhai, Thomas Unterthiner, Jessica Yung, Andreas Steiner, Daniel Keysers, Jakob Uszkoreit, Mario Lucic, and Alexey Dosovitskiy.
\newblock Mlp-mixer: An all-mlp architecture for vision.
\newblock In M.~Ranzato, A.~Beygelzimer, Y.~Dauphin, P.S. Liang, and J.~Wortman Vaughan (eds.), \emph{Advances in Neural Information Processing Systems}, volume~34, pp.\  24261--24272. Curran Associates, Inc., 2021.
\newblock URL \url{https://proceedings.neurips.cc/paper_files/paper/2021/file/cba0a4ee5ccd02fda0fe3f9a3e7b89fe-Paper.pdf}.

\bibitem[Touvron et~al.(2023)Touvron, Martin, Stone, Albert, Almahairi, Babaei, Bashlykov, Batra, Bhargava, Bhosale, Bikel, Blecher, Ferrer, Chen, Cucurull, Esiobu, Fernandes, Fu, Fu, Fuller, Gao, Goswami, Goyal, Hartshorn, Hosseini, Hou, Inan, Kardas, Kerkez, Khabsa, Kloumann, Korenev, Koura, Lachaux, Lavril, Lee, Liskovich, Lu, Mao, Martinet, Mihaylov, Mishra, Molybog, Nie, Poulton, Reizenstein, Rungta, Saladi, Schelten, Silva, Smith, Subramanian, Tan, Tang, Taylor, Williams, Kuan, Xu, Yan, Zarov, Zhang, Fan, Kambadur, Narang, Rodriguez, Stojnic, Edunov, and Scialom]{touvron2023llama}
Hugo Touvron, Louis Martin, Kevin Stone, Peter Albert, Amjad Almahairi, Yasmine Babaei, Nikolay Bashlykov, Soumya Batra, Prajjwal Bhargava, Shruti Bhosale, Dan Bikel, Lukas Blecher, Cristian~Canton Ferrer, Moya Chen, Guillem Cucurull, David Esiobu, Jude Fernandes, Jeremy Fu, Wenyin Fu, Brian Fuller, Cynthia Gao, Vedanuj Goswami, Naman Goyal, Anthony Hartshorn, Saghar Hosseini, Rui Hou, Hakan Inan, Marcin Kardas, Viktor Kerkez, Madian Khabsa, Isabel Kloumann, Artem Korenev, Punit~Singh Koura, Marie-Anne Lachaux, Thibaut Lavril, Jenya Lee, Diana Liskovich, Yinghai Lu, Yuning Mao, Xavier Martinet, Todor Mihaylov, Pushkar Mishra, Igor Molybog, Yixin Nie, Andrew Poulton, Jeremy Reizenstein, Rashi Rungta, Kalyan Saladi, Alan Schelten, Ruan Silva, Eric~Michael Smith, Ranjan Subramanian, Xiaoqing~Ellen Tan, Binh Tang, Ross Taylor, Adina Williams, Jian~Xiang Kuan, Puxin Xu, Zheng Yan, Iliyan Zarov, Yuchen Zhang, Angela Fan, Melanie Kambadur, Sharan Narang, Aurelien Rodriguez, Robert Stojnic, Sergey Edunov, and Thomas
  Scialom.
\newblock Llama 2: Open foundation and fine-tuned chat models, 2023.

\bibitem[Vaswani et~al.(2023)Vaswani, Shazeer, Parmar, Uszkoreit, Jones, Gomez, Kaiser, and Polosukhin]{vaswani2023attention}
Ashish Vaswani, Noam Shazeer, Niki Parmar, Jakob Uszkoreit, Llion Jones, Aidan~N. Gomez, Lukasz Kaiser, and Illia Polosukhin.
\newblock Attention is all you need, 2023.

\bibitem[Voelker et~al.(2019)Voelker, Kaji{\'c}, and Eliasmith]{voelker2019legendre}
Aaron Voelker, Ivana Kaji{\'c}, and Chris Eliasmith.
\newblock Legendre memory units: Continuous-time representation in recurrent neural networks.
\newblock \emph{Advances in neural information processing systems}, 32, 2019.

\bibitem[Wan et~al.(2023)Wan, Wang, Liu, Alam, Zheng, et~al.]{wan2023efficient}
Zhongwei Wan, Xin Wang, Che Liu, Samiul Alam, Yu~Zheng, et~al.
\newblock Efficient large language models: A survey.
\newblock \emph{arXiv preprint arXiv:2312.03863}, 1, 2023.

\bibitem[Wang et~al.(2018)Wang, Singh, Michael, Hill, Levy, and Bowman]{wang2018glue}
Alex Wang, Amanpreet Singh, Julian Michael, Felix Hill, Omer Levy, and Samuel Bowman.
\newblock Glue: A multi-task benchmark and analysis platform for natural language understanding.
\newblock In \emph{Proceedings of the 2018 EMNLP Workshop BlackboxNLP: Analyzing and Interpreting Neural Networks for NLP}, pp.\  353--355, 2018.

\bibitem[Wang et~al.(2020)Wang, Li, Khabsa, Fang, and Ma]{wang2020linformer}
Sinong Wang, Belinda~Z Li, Madian Khabsa, Han Fang, and Hao Ma.
\newblock Linformer: Self-attention with linear complexity.
\newblock \emph{arXiv preprint arXiv:2006.04768}, 2020.

\bibitem[Wang et~al.(2024)Wang, Wang, Ding, Li, Wu, Rong, Kong, Huang, Li, Yang, et~al.]{wang2024state}
Xiao Wang, Shiao Wang, Yuhe Ding, Yuehang Li, Wentao Wu, Yao Rong, Weizhe Kong, Ju~Huang, Shihao Li, Haoxiang Yang, et~al.
\newblock State space model for new-generation network alternative to transformers: A survey.
\newblock \emph{arXiv preprint arXiv:2404.09516}, 2024.

\bibitem[Welbl et~al.(2017)Welbl, Liu, and Gardner]{welbl2017crowdsourcing}
Johannes Welbl, Nelson~F Liu, and Matt Gardner.
\newblock Crowdsourcing multiple choice science questions.
\newblock In \emph{Proceedings of the 3rd Workshop on Noisy User-generated Text}, pp.\  94--106, 2017.

\bibitem[Wikimedia-Foundation(2022)]{wikidump}
Wikimedia-Foundation.
\newblock Wikimedia downloads, 2022.
\newblock URL \url{https://dumps.wikimedia.org}.

\bibitem[Workshop et~al.(2023)Workshop, :, Scao, Fan, Akiki, Pavlick, Ilić, Hesslow, Castagné, Luccioni, Yvon, Gallé, Tow, Rush, Biderman, Webson, Ammanamanchi, Wang, Sagot, Muennighoff, del Moral, Ruwase, Bawden, Bekman, McMillan-Major, Beltagy, Nguyen, Saulnier, Tan, Suarez, Sanh, Laurençon, Jernite, Launay, Mitchell, Raffel, Gokaslan, Simhi, Soroa, Aji, Alfassy, Rogers, Nitzav, Xu, Mou, Emezue, Klamm, Leong, van Strien, Adelani, Radev, Ponferrada, Levkovizh, Kim, Natan, Toni, Dupont, Kruszewski, Pistilli, Elsahar, Benyamina, Tran, Yu, Abdulmumin, Johnson, Gonzalez-Dios, de~la Rosa, Chim, Dodge, Zhu, Chang, Frohberg, Tobing, Bhattacharjee, Almubarak, Chen, Lo, Werra, Weber, Phan, allal, Tanguy, Dey, Muñoz, Masoud, Grandury, Šaško, Huang, Coavoux, Singh, Jiang, Vu, Jauhar, Ghaleb, Subramani, Kassner, Khamis, Nguyen, Espejel, de~Gibert, Villegas, Henderson, Colombo, Amuok, Lhoest, Harliman, Bommasani, López, Ribeiro, Osei, Pyysalo, Nagel, Bose, Muhammad, Sharma, Longpre, Nikpoor, Silberberg, Pai,
  Zink, Torrent, Schick, Thrush, Danchev, Nikoulina, Laippala, Lepercq, Prabhu, Alyafeai, Talat, Raja, Heinzerling, Si, Taşar, Salesky, Mielke, Lee, Sharma, Santilli, Chaffin, Stiegler, Datta, Szczechla, Chhablani, Wang, Pandey, Strobelt, Fries, Rozen, Gao, Sutawika, Bari, Al-shaibani, Manica, Nayak, Teehan, Albanie, Shen, Ben-David, Bach, Kim, Bers, Fevry, Neeraj, Thakker, Raunak, Tang, Yong, Sun, Brody, Uri, Tojarieh, Roberts, Chung, Tae, Phang, Press, Li, Narayanan, Bourfoune, Casper, Rasley, Ryabinin, Mishra, Zhang, Shoeybi, Peyrounette, Patry, Tazi, Sanseviero, von Platen, Cornette, Lavallée, Lacroix, Rajbhandari, Gandhi, Smith, Requena, Patil, Dettmers, Baruwa, Singh, Cheveleva, Ligozat, Subramonian, Névéol, Lovering, Garrette, Tunuguntla, Reiter, Taktasheva, Voloshina, Bogdanov, Winata, Schoelkopf, Kalo, Novikova, Forde, Clive, Kasai, Kawamura, Hazan, Carpuat, Clinciu, Kim, Cheng, Serikov, Antverg, van~der Wal, Zhang, Zhang, Gehrmann, Mirkin, Pais, Shavrina, Scialom, Yun, Limisiewicz, Rieser,
  Protasov, Mikhailov, Pruksachatkun, Belinkov, Bamberger, Kasner, Rueda, Pestana, Feizpour, Khan, Faranak, Santos, Hevia, Unldreaj, Aghagol, Abdollahi, Tammour, HajiHosseini, Behroozi, Ajibade, Saxena, Ferrandis, McDuff, Contractor, Lansky, David, Kiela, Nguyen, Tan, Baylor, Ozoani, Mirza, Ononiwu, Rezanejad, Jones, Bhattacharya, Solaiman, Sedenko, Nejadgholi, Passmore, Seltzer, Sanz, Dutra, Samagaio, Elbadri, Mieskes, Gerchick, Akinlolu, McKenna, Qiu, Ghauri, Burynok, Abrar, Rajani, Elkott, Fahmy, Samuel, An, Kromann, Hao, Alizadeh, Shubber, Wang, Roy, Viguier, Le, Oyebade, Le, Yang, Nguyen, Kashyap, Palasciano, Callahan, Shukla, Miranda-Escalada, Singh, Beilharz, Wang, Brito, Zhou, Jain, Xu, Fourrier, Periñán, Molano, Yu, Manjavacas, Barth, Fuhrimann, Altay, Bayrak, Burns, Vrabec, Bello, Dash, Kang, Giorgi, Golde, Posada, Sivaraman, Bulchandani, Liu, Shinzato, de~Bykhovetz, Takeuchi, Pàmies, Castillo, Nezhurina, Sänger, Samwald, Cullan, Weinberg, Wolf, Mihaljcic, Liu, Freidank, Kang, Seelam, Dahlberg,
  Broad, Muellner, Fung, Haller, Chandrasekhar, Eisenberg, Martin, Canalli, Su, Su, Cahyawijaya, Garda, Deshmukh, Mishra, Kiblawi, Ott, Sang-aroonsiri, Kumar, Schweter, Bharati, Laud, Gigant, Kainuma, Kusa, Labrak, Bajaj, Venkatraman, Xu, Xu, Xu, Tan, Xie, Ye, Bras, Belkada, and Wolf]{workshop2023bloom}
BigScience Workshop, :, Teven~Le Scao, Angela Fan, Christopher Akiki, Ellie Pavlick, Suzana Ilić, Daniel Hesslow, Roman Castagné, Alexandra~Sasha Luccioni, François Yvon, Matthias Gallé, Jonathan Tow, Alexander~M. Rush, Stella Biderman, Albert Webson, Pawan~Sasanka Ammanamanchi, Thomas Wang, Benoît Sagot, Niklas Muennighoff, Albert~Villanova del Moral, Olatunji Ruwase, Rachel Bawden, Stas Bekman, Angelina McMillan-Major, Iz~Beltagy, Huu Nguyen, Lucile Saulnier, Samson Tan, Pedro~Ortiz Suarez, Victor Sanh, Hugo Laurençon, Yacine Jernite, Julien Launay, Margaret Mitchell, Colin Raffel, Aaron Gokaslan, Adi Simhi, Aitor Soroa, Alham~Fikri Aji, Amit Alfassy, Anna Rogers, Ariel~Kreisberg Nitzav, Canwen Xu, Chenghao Mou, Chris Emezue, Christopher Klamm, Colin Leong, Daniel van Strien, David~Ifeoluwa Adelani, Dragomir Radev, Eduardo~González Ponferrada, Efrat Levkovizh, Ethan Kim, Eyal~Bar Natan, Francesco~De Toni, Gérard Dupont, Germán Kruszewski, Giada Pistilli, Hady Elsahar, Hamza Benyamina, Hieu Tran,
  Ian Yu, Idris Abdulmumin, Isaac Johnson, Itziar Gonzalez-Dios, Javier de~la Rosa, Jenny Chim, Jesse Dodge, Jian Zhu, Jonathan Chang, Jörg Frohberg, Joseph Tobing, Joydeep Bhattacharjee, Khalid Almubarak, Kimbo Chen, Kyle Lo, Leandro~Von Werra, Leon Weber, Long Phan, Loubna~Ben allal, Ludovic Tanguy, Manan Dey, Manuel~Romero Muñoz, Maraim Masoud, María Grandury, Mario Šaško, Max Huang, Maximin Coavoux, Mayank Singh, Mike Tian-Jian Jiang, Minh~Chien Vu, Mohammad~A. Jauhar, Mustafa Ghaleb, Nishant Subramani, Nora Kassner, Nurulaqilla Khamis, Olivier Nguyen, Omar Espejel, Ona de~Gibert, Paulo Villegas, Peter Henderson, Pierre Colombo, Priscilla Amuok, Quentin Lhoest, Rheza Harliman, Rishi Bommasani, Roberto~Luis López, Rui Ribeiro, Salomey Osei, Sampo Pyysalo, Sebastian Nagel, Shamik Bose, Shamsuddeen~Hassan Muhammad, Shanya Sharma, Shayne Longpre, Somaieh Nikpoor, Stanislav Silberberg, Suhas Pai, Sydney Zink, Tiago~Timponi Torrent, Timo Schick, Tristan Thrush, Valentin Danchev, Vassilina Nikoulina,
  Veronika Laippala, Violette Lepercq, Vrinda Prabhu, Zaid Alyafeai, Zeerak Talat, Arun Raja, Benjamin Heinzerling, Chenglei Si, Davut~Emre Taşar, Elizabeth Salesky, Sabrina~J. Mielke, Wilson~Y. Lee, Abheesht Sharma, Andrea Santilli, Antoine Chaffin, Arnaud Stiegler, Debajyoti Datta, Eliza Szczechla, Gunjan Chhablani, Han Wang, Harshit Pandey, Hendrik Strobelt, Jason~Alan Fries, Jos Rozen, Leo Gao, Lintang Sutawika, M~Saiful Bari, Maged~S. Al-shaibani, Matteo Manica, Nihal Nayak, Ryan Teehan, Samuel Albanie, Sheng Shen, Srulik Ben-David, Stephen~H. Bach, Taewoon Kim, Tali Bers, Thibault Fevry, Trishala Neeraj, Urmish Thakker, Vikas Raunak, Xiangru Tang, Zheng-Xin Yong, Zhiqing Sun, Shaked Brody, Yallow Uri, Hadar Tojarieh, Adam Roberts, Hyung~Won Chung, Jaesung Tae, Jason Phang, Ofir Press, Conglong Li, Deepak Narayanan, Hatim Bourfoune, Jared Casper, Jeff Rasley, Max Ryabinin, Mayank Mishra, Minjia Zhang, Mohammad Shoeybi, Myriam Peyrounette, Nicolas Patry, Nouamane Tazi, Omar Sanseviero, Patrick von
  Platen, Pierre Cornette, Pierre~François Lavallée, Rémi Lacroix, Samyam Rajbhandari, Sanchit Gandhi, Shaden Smith, Stéphane Requena, Suraj Patil, Tim Dettmers, Ahmed Baruwa, Amanpreet Singh, Anastasia Cheveleva, Anne-Laure Ligozat, Arjun Subramonian, Aurélie Névéol, Charles Lovering, Dan Garrette, Deepak Tunuguntla, Ehud Reiter, Ekaterina Taktasheva, Ekaterina Voloshina, Eli Bogdanov, Genta~Indra Winata, Hailey Schoelkopf, Jan-Christoph Kalo, Jekaterina Novikova, Jessica~Zosa Forde, Jordan Clive, Jungo Kasai, Ken Kawamura, Liam Hazan, Marine Carpuat, Miruna Clinciu, Najoung Kim, Newton Cheng, Oleg Serikov, Omer Antverg, Oskar van~der Wal, Rui Zhang, Ruochen Zhang, Sebastian Gehrmann, Shachar Mirkin, Shani Pais, Tatiana Shavrina, Thomas Scialom, Tian Yun, Tomasz Limisiewicz, Verena Rieser, Vitaly Protasov, Vladislav Mikhailov, Yada Pruksachatkun, Yonatan Belinkov, Zachary Bamberger, Zdeněk Kasner, Alice Rueda, Amanda Pestana, Amir Feizpour, Ammar Khan, Amy Faranak, Ana Santos, Anthony Hevia, Antigona
  Unldreaj, Arash Aghagol, Arezoo Abdollahi, Aycha Tammour, Azadeh HajiHosseini, Bahareh Behroozi, Benjamin Ajibade, Bharat Saxena, Carlos~Muñoz Ferrandis, Daniel McDuff, Danish Contractor, David Lansky, Davis David, Douwe Kiela, Duong~A. Nguyen, Edward Tan, Emi Baylor, Ezinwanne Ozoani, Fatima Mirza, Frankline Ononiwu, Habib Rezanejad, Hessie Jones, Indrani Bhattacharya, Irene Solaiman, Irina Sedenko, Isar Nejadgholi, Jesse Passmore, Josh Seltzer, Julio~Bonis Sanz, Livia Dutra, Mairon Samagaio, Maraim Elbadri, Margot Mieskes, Marissa Gerchick, Martha Akinlolu, Michael McKenna, Mike Qiu, Muhammed Ghauri, Mykola Burynok, Nafis Abrar, Nazneen Rajani, Nour Elkott, Nour Fahmy, Olanrewaju Samuel, Ran An, Rasmus Kromann, Ryan Hao, Samira Alizadeh, Sarmad Shubber, Silas Wang, Sourav Roy, Sylvain Viguier, Thanh Le, Tobi Oyebade, Trieu Le, Yoyo Yang, Zach Nguyen, Abhinav~Ramesh Kashyap, Alfredo Palasciano, Alison Callahan, Anima Shukla, Antonio Miranda-Escalada, Ayush Singh, Benjamin Beilharz, Bo~Wang, Caio Brito,
  Chenxi Zhou, Chirag Jain, Chuxin Xu, Clémentine Fourrier, Daniel~León Periñán, Daniel Molano, Dian Yu, Enrique Manjavacas, Fabio Barth, Florian Fuhrimann, Gabriel Altay, Giyaseddin Bayrak, Gully Burns, Helena~U. Vrabec, Imane Bello, Ishani Dash, Jihyun Kang, John Giorgi, Jonas Golde, Jose~David Posada, Karthik~Rangasai Sivaraman, Lokesh Bulchandani, Lu~Liu, Luisa Shinzato, Madeleine~Hahn de~Bykhovetz, Maiko Takeuchi, Marc Pàmies, Maria~A Castillo, Marianna Nezhurina, Mario Sänger, Matthias Samwald, Michael Cullan, Michael Weinberg, Michiel~De Wolf, Mina Mihaljcic, Minna Liu, Moritz Freidank, Myungsun Kang, Natasha Seelam, Nathan Dahlberg, Nicholas~Michio Broad, Nikolaus Muellner, Pascale Fung, Patrick Haller, Ramya Chandrasekhar, Renata Eisenberg, Robert Martin, Rodrigo Canalli, Rosaline Su, Ruisi Su, Samuel Cahyawijaya, Samuele Garda, Shlok~S Deshmukh, Shubhanshu Mishra, Sid Kiblawi, Simon Ott, Sinee Sang-aroonsiri, Srishti Kumar, Stefan Schweter, Sushil Bharati, Tanmay Laud, Théo Gigant, Tomoya
  Kainuma, Wojciech Kusa, Yanis Labrak, Yash~Shailesh Bajaj, Yash Venkatraman, Yifan Xu, Yingxin Xu, Yu~Xu, Zhe Tan, Zhongli Xie, Zifan Ye, Mathilde Bras, Younes Belkada, and Thomas Wolf.
\newblock Bloom: A 176b-parameter open-access multilingual language model, 2023.

\bibitem[Wu et~al.(2023)Wu, Li, Yu, and Sun]{wu2023tunesformer}
Shangda Wu, Xiaobing Li, Feng Yu, and Maosong Sun.
\newblock Tunesformer: Forming irish tunes with control codes by bar patching.
\newblock In Lorenzo Porcaro, Roser Batlle{-}Roca, and Emilia G{\'{o}}mez (eds.), \emph{Proceedings of the 2nd Workshop on Human-Centric Music Information Retrieval 2023 co-located with the 24th International Society for Music Information Retrieval Conference {(ISMIR} 2023), Milan, Italy, November 10, 2023}, volume 3528 of \emph{{CEUR} Workshop Proceedings}. CEUR-WS.org, 2023.
\newblock URL \url{https://ceur-ws.org/Vol-3528/paper1.pdf}.

\bibitem[Wu \& He(2018)Wu and He]{wu2018group}
Yuxin Wu and Kaiming He.
\newblock Group normalization.
\newblock \emph{arXiv:1803.08494}, 2018.

\bibitem[Xiao et~al.(2023)Xiao, Tian, Chen, Han, and Lewis]{xiao2023efficient}
Guangxuan Xiao, Yuandong Tian, Beidi Chen, Song Han, and Mike Lewis.
\newblock Efficient streaming language models with attention sinks.
\newblock \emph{arXiv preprint arXiv:2309.17453}, 2023.

\bibitem[Xie et~al.(2024)Xie, Pham, Dong, Du, Liu, Lu, Liang, Le, Ma, and Yu]{xie2024doremi}
Sang~Michael Xie, Hieu Pham, Xuanyi Dong, Nan Du, Hanxiao Liu, Yifeng Lu, Percy~S Liang, Quoc~V Le, Tengyu Ma, and Adams~Wei Yu.
\newblock Doremi: Optimizing data mixtures speeds up language model pretraining.
\newblock \emph{Advances in Neural Information Processing Systems}, 36, 2024.

\bibitem[Xiong et~al.(2021)Xiong, Zeng, Chakraborty, Tan, Fung, Li, and Singh]{xiong2021nystromformer}
Yunyang Xiong, Zhanpeng Zeng, Rudrasis Chakraborty, Mingxing Tan, Glenn Fung, Yin Li, and Vikas Singh.
\newblock Nystr\"omformer: A nystr\"om-based algorithm for approximating self-attention, 2021.

\bibitem[Xu et~al.(2023)Xu, Sun, Zheng, Geng, Zhao, Feng, Tao, and Jiang]{xu2023wizardlm}
Can Xu, Qingfeng Sun, Kai Zheng, Xiubo Geng, Pu~Zhao, Jiazhan Feng, Chongyang Tao, and Daxin Jiang.
\newblock Wizardlm: Empowering large language models to follow complex instructions, 2023.

\bibitem[Yang et~al.(2023)Yang, Wang, Shen, Panda, and Kim]{yang2023gated}
Songlin Yang, Bailin Wang, Yikang Shen, Rameswar Panda, and Yoon Kim.
\newblock Gated linear attention transformers with hardware-efficient training, 2023.

\bibitem[Yang et~al.(2019)Yang, Zhang, Tar, and Baldridge]{yang2019paws}
Yinfei Yang, Yuan Zhang, Chris Tar, and Jason Baldridge.
\newblock Paws-x: A cross-lingual adversarial dataset for paraphrase identification.
\newblock In \emph{Proceedings of the 2019 Conference on Empirical Methods in Natural Language Processing and the 9th International Joint Conference on Natural Language Processing (EMNLP-IJCNLP)}, pp.\  3687--3692, 2019.

\bibitem[Yuan et~al.(2023)Yuan, Li, and Sun]{Yuan2023TinyGPTVEM}
Zhengqing Yuan, Zhaoxu Li, and Lichao Sun.
\newblock Tinygpt-v: Efficient multimodal large language model via small backbones.
\newblock \emph{ArXiv}, abs/2312.16862, 2023.
\newblock URL \url{https://api.semanticscholar.org/CorpusID:266572996}.

\bibitem[Zaheer et~al.(2020)Zaheer, Guruganesh, Dubey, Ainslie, Alberti, Ontanon, Pham, Ravula, Wang, Yang, et~al.]{zaheer2020bigbird}
Manzil Zaheer, Guru Guruganesh, Kumar~Avinava Dubey, Joshua Ainslie, Chris Alberti, Santiago Ontanon, Philip Pham, Anirudh Ravula, Qifan Wang, Li~Yang, et~al.
\newblock Big bird: Transformers for longer sequences.
\newblock \emph{Advances in Neural Information Processing Systems}, 33, 2020.

\bibitem[Zhai et~al.(2021)Zhai, Talbott, Srivastava, Huang, Goh, Zhang, and Susskind]{zhai2021attention}
Shuangfei Zhai, Walter Talbott, Nitish Srivastava, Chen Huang, Hanlin Goh, Ruixiang Zhang, and Josh Susskind.
\newblock An attention free transformer, 2021.

\bibitem[Zhang et~al.(2024)Zhang, Zeng, Wang, and Lu]{zhang2024tinyllama}
Peiyuan Zhang, Guangtao Zeng, Tianduo Wang, and Wei Lu.
\newblock Tinyllama: An open-source small language model, 2024.

\bibitem[Zheng et~al.(2024)Zheng, Chiang, Sheng, Zhuang, Wu, Zhuang, Lin, Li, Li, Xing, et~al.]{zheng2024judging}
Lianmin Zheng, Wei-Lin Chiang, Ying Sheng, Siyuan Zhuang, Zhanghao Wu, Yonghao Zhuang, Zi~Lin, Zhuohan Li, Dacheng Li, Eric Xing, et~al.
\newblock Judging llm-as-a-judge with mt-bench and chatbot arena.
\newblock \emph{Advances in Neural Information Processing Systems}, 36, 2024.

\end{thebibliography}
